\documentclass{article}
\pdfoutput=1
\usepackage[american]{babel}
\usepackage{amssymb}
\usepackage[all]{xy}
\usepackage{xspace}
\usepackage{amsfonts}
\usepackage{amsthm}
\usepackage{placeins}
\usepackage{color}
\usepackage{ifthen}
\usepackage{url}

\usepackage{graphicx}

\newboolean{final}
\setboolean{final}{true}

  \usepackage[final]{showlabels}
  \usepackage[hide]{ed}

\newif\ifpdf\ifx\pdfoutput\undefined\pdffalse\else\pdfoutput=1\pdftrue\fi

\ifpdf
\fi

\newcommand{\rel}[1]{\ensuremath{\;\textnormal{\rm #1}\;}}
\newcommand{\relnsp}[1]{\ensuremath{\textnormal{\rm #1}}\xspace}

\newcommand{\map}[2]{\colon#1\!\longrightarrow\!#2}
\newcommand{\dipole}{\mathbb{D}}
\newcommand{\opra}{\mathbb{OP}}
\newcommand{\vect}[2]{\overrightarrow{#1~ #2}}

\def\inv#1{#1^{\smallsmile}}

\newenvironment{pf}{\par\addvspace{1em}\noindent{\bfseries Proof. }\enspace}{\qed\par\addvspace{1em}}
\newcommand{\R}[0]{\ensuremath{\mathbb{R}}}

\theoremstyle{plain}
\newtheorem{thm}{Theorem}
\newtheorem{cor}[thm]{Corollary}
\newtheorem{lem}[thm]{Lemma}

\newtheorem{prop}[thm]{Proposition}

\theoremstyle{definition}
\newtheorem{defn}[thm]{Definition}
\newtheorem{exmp}[thm]{Example}

\newtheorem{notation}[thm]{Notation}

\newcommand{\LR}{\mathcal{LR}}
\newcommand{\DRAf}{\mathcal{DRA}_{f}}
\newcommand{\DRAc}{\mathcal{DRA}_{c}}
\newcommand{\DRAfp}{\mathcal{DRA}_{\mathit{fp}}}
\newcommand{\DRAop}{\mathcal{DRA}_{\mathit{op}}}
\newcommand{\DRAopp}{\mathcal{DRA}_{\mathit{opp}}}
\newcommand{\DRAlr}{\mathcal{DRA}_{\mathit{lr}}}
\newcommand{\OPRA}{\mathcal{OPRA}}

\newcommand{\rf}{\varphi_{\mathit{f}}}
\newcommand{\rfp}{\varphi_{\mathit{fp}}}
\newcommand{\rop}{\varphi_{\mathit{op}}}
\newcommand{\ropp}{\varphi_{\mathit{opp}}}

\newcommand{\D}{{\cal U}}
\newcommand{\V}{{\cal V}}
\newcommand{\Dsim}{\D\!/\!\!\sim}
\newcommand{\Asim}{A\!/\!\!\sim_A}
\newcommand{\varphisim}{\varphi\!/\!\!\sim}
\newcommand{\hD}{i}

\begin{document}


\title{Oriented Straight Line Segment Algebra:
  Qualitative Spatial Reasoning\\
  about Oriented Objects}
\author{$\mbox{Reinhard Moratz}^1$ and  $\mbox{Dominik L{\"u}cke}^2$ and $\mbox{Till Mossakowski}^2$}
\maketitle
\begin{center}
{$^1\mbox{University of Maine,}$\\
National Center for Geographic Information and Analysis,\\
Department of Spatial Information Science and Engineering,\\
348 Boardman Hall, Orono, 04469 Maine, USA.\\
{\tt moratz@spatial.maine.edu}\\[0.3cm]
$^2\mbox{University of Bremen,}$\\
Collaborative Research Center on Spatial Cognition (SFB/TR 8),\\
Department of Mathematics and Informatics,\\
Bibliothekstr. 1, 28359 Bremen, Germany.\\
{\tt till$\;|\,$luecke@informatik.uni-bremen.de}
}
\end{center}

\begin{abstract}
\begin{sloppy}
Nearly 15 years ago, a set of qualitative spatial relations between
oriented straight line segments (dipoles) was suggested by Schlieder. This work
received substantial interest amongst the qualitative spatial reasoning
community. However, it turned out to be difficult to establish a sound
constraint calculus based on these relations.
In this paper, we present the results of a new investigation into dipole
constraint calculi which uses algebraic methods to derive
sound results on the composition of relations and other properties
of dipole calculi.  Our results are based on a condensed semantics of the dipole relations.

In contrast to the points that are normally used, dipoles are extended
and have an intrinsic direction. Both features
are important properties of natural objects.
This allows for a straightforward representation of prototypical reasoning tasks for
spatial agents.
As an example, we show how to generate survey knowledge from local observations in a
street network.  The example illustrates the fast constraint-based reasoning capabilities
of the dipole calculus.  We integrate our results into two reasoning
tools which are publicly available.
\end{sloppy}
\end{abstract}



\section{Introduction}
Qualitative Reasoning about space abstracts from the physical world and
enables computers to make predictions about spatial relations,
even when precise quantitative information is not
available \cite{cohn97}.
A {\it qualitative} representation provides mechanisms
which characterize the essential properties of objects or
configurations.
In contrast, a {\it quantitative} representation establishes a
measure in relation to
a unit of measurement which must be generally available
\cite{freksa_cosit0}.
The constant and general availability of common measures is now self-evident.
However, one needs only recall the history of length measurement technologies
to see that the more local relative measures,
which are represented qualitatively\footnote{Compare for example the
qualitative expression ''one piece of material is longer than another''
with the quantitative expression
''this thing is two meters long''},
can be managed by biological/epigenetic cognitive systems
much more easily than absolute quantitative representations.

\begin{figure}[htb]
\begin{center}
\includegraphics[keepaspectratio, scale=0.85]{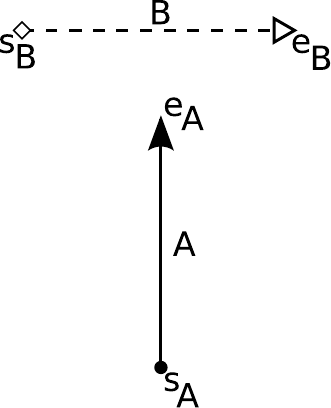}
\caption{\label{left} Orientation between two dipoles}
\end{center}
\end{figure}

Qualitative spatial calculi usually deal with elementary objects (e.g.
positions, directions, regions) and qualitative relations between them
(e.g. ''adjacent'', ''to the left of'', ''included in'').
This is the reason why qualitative descriptions are quite natural for people.
The two main trends in Qualitative Spatial
Reasoning (QSR) are topological reasoning about regions
\cite{Randell89_RCCa,Randell92_RCCb,egenhofer9intersection,renz99_RCC_Complexity,worboysClementini01}
and positional (e.g. direction and distance)
reasoning about point configurations
\cite{freksa92b,frank91,Ligozat98_CardDir,hernandez_aij,appliedAI_zimmermann,isli_moratz_memo,moratz05_QSR_RPP}.
Positions can refer to a
global reference system, e.g. cardinal directions, or
just to local reference systems, e.g. egocentric views.
Positional calculi can be related to the results of Psycholinguistic
research \cite{MoratzTenbrink2006} in the field of reference systems.
In Psycholinguistics, local reference systems are divided into two modalities:
intrinsic reference systems and extrinsic reference systems.
Then, the three resulting options for giving a linguistic description of the spatial arrangements of objects
are: {\it intrinsic}, {\it extrinsic}, and {\it absolute} (i.e. global) reference systems
\cite{levinson96}\footnote{In \cite{levinson96}, extrinsic references are called relative references.}.
Corresponding QSR calculi can be found in Psycholinguistics for all three types of reference systems.
An intrinsic reference system employs an oriented physical object as the origin of a reference system
(relatum).
The orientation of the physical object then serves as a reference direction for the reference system.
For instance, an intrinsic reference system is used in the calculus
of oriented line segments (see Fig.~\ref{left}) which is the main topic of this paper.
Another calculus corresponding to intrinsic reference systems is
the $\OPRA$ calculus \cite{Moratz06_ECAI}.
In the $\OPRA$ calculus, oriented points are the basic entities (see Fig.~\ref{Regions}).

Extrinsic reference systems are closely related to intrinsic reference systems.
Both reference system options share the feature of focusing on the local context.
The difference is that the extrinsic reference system superimposes the view direction
from an external observer as reference direction
to the relatum of the reference system.
A typical example for a QSR calculus corresponding to an extrinsic reference system
is Freksa's double cross calculus \cite{cosyZIM96}.
In the double cross calculus, two points span a reference system to localize a third point.
The first point then projects a view towards the second point which
generates the reference direction.

Since intrinsic and extrinsic references are
closely related in the rest of the paper, we sometimes refer to QSR calculi which use either intrinsic
or extrinsic reference systems as relative position QSR calculi.
Then, the two terms {\it local reference systems} and {\it relative reference systems}
refer to the same concept.
An interesting special case refers to the representation of a relative orientation
without the concept of distance.
These relative orientations can be viewed as decoupled from anchor points. Then there is no means
for distinguishing between different point locations.
The great advantage is that much more efficient reasoning mechanisms become available.
The work by Isli and Cohn \cite{Isli00_2DOriOrdering} consists of a ternary calculus for reasoning about such
pure orientations.
In contrast to relative position calculi, their algebra has a tractable
subset containing the base relations.

Absolute (or global) directions
can relate directional knowledge from distant places to each other.
Cardinal directions as an example can be registered with a compass and compared over a large distance.
And for that reason Frank's cardinal direction calculus
corresponds to such an absolute reference system
\cite{frank91}, \cite{ligozat98}.
There is a variant of a cardinal direction calculus, which has a flexible granularity,
the Star Calculus \cite{Renz04_QDCArbGranu}.

\sloppypar
In the previous paragraphs, we discussed the {\it representation} of spatial knowledge.
Another important aspect are the {\it reasoning} mechanisms which are employed to make
use of the represented initial knowledge to infer indirect knowledge.
In Qualitative Spatial Reasoning
two main reasoning modes are used: Conceptual neighbourhood-based reasoning, and
constraint-based reasoning about (static) spatial configurations.
Conceptual neighborhood-based reasoning describes whether two spatial configurations
of objects can be transformed into each other by small changes
\cite{cosyFRE91}.
The conceptual neighborhood of a qualitative spatial relation which
holds for a spatial arrangement is the set of relations into which
a relation can be changed with minimal transformations, e.g. by continuous
deformation.
Such a transformation can be a movement
of one object in the configuration in a short period of time.
At the discrete level of concepts, the neighborhood corresponds
to continuity on the geometric or physical level of description:
continuous processes map onto identical or neighboring classes of
descriptions \cite{cosyFreksa04}. Spatial conceptual neighborhoods are very natural
perceptual and cognitive entities and other neighborhood structures can be
derived from spatial neighborhoods, e.g. temporal neighborhoods.
The movement of an agent can then be modeled qualitatively as a sequence of neighboring
spatial relations which hold for adjacent time intervals\footnote{This was the reasoning
used in the first investigation of dipole relations by Schlieder \cite{Schlieder95}}. Based on this
qualitative representation of trajectories, neighborhood-based spatial
reasoning can for example be used as a simple, abstract model of the navigation
of a spatial agent\footnote{for an application of neighbourhood based
reasoning of spatial agents, we refer the reader to the simulation model
SAILAWAY \cite{cosy:R3-R4-sailaway-aisb}}.

In constraint-based reasoning about spatial configurations, typically a
partial initial knowledge of a scene is represented in terms of qualitative
constraints between spatial objects.  Implicit knowledge about spatial
relations is then derived by constraint propagation\footnote{For an
application of constraint-based reasoning for spatial agents, we refer the reader
to the AIBO robot example in \cite{moratz05_QSR_RPP}}.  Previous
research has found that the mathematical notion of a \emph{relation
  algebra} and related notions are well-suited for this kind of
reasoning.  In many cases, relation algebra-based reasoning only
provides approximate results \cite{RenzLigozat} and the constraint
consistency problem for relative position calculi is NP-hard
\cite{LeeWolterAIJ}.  Hence we use constraint reasoning with
polynomial time algorithms as an approximation of an intractable
problem. The technical details of constraint reasoning are explained
in Section~\ref{ConstraintPropagationEtc}.

In point-based reasoning, all objects are mapped onto the plane.
The centers of projected objects can be used as point-like representation of the objects.
By contrast,
Schlieder's line segment calculus \cite{Schlieder95}
uses more complex
basic entities. Thus, it is
based on extended objects which are represented as oriented straight
line segments (see Fig.~\ref{left}).
These more complex basic entities capture important
features of natural objects:
\begin{itemize}
\item
Natural Objects are extended.
\item
Natural Objects often have an intrinsic direction.
\end{itemize}
Oriented straight line segments
(which were called {\em dipoles} by
Moratz et al. \cite{Moratz00_QSRwithLineSegs})
are the simplest geometric objects presenting these features.
Dipoles may be specified by their start and end points.

Using dipoles as basic blocks, more complex objects can
be constructed (e.g. polylines, polygons) in a straightforward manner.
Therefore, dipoles can be used as the basic units in numerous
applications. To give an example,
line segments are central to edge-based image segmentation and grouping in computer vision. In addition, GIS systems often have
line segments as basic entities \cite{hoel91}.
Polylines are particularly interesting for
representing paths in cognitive robotics \cite{musto_ijcai99} and can serve as the geometric basis of a mobile robot when autonomously mapping its working environment \cite{Wolter04_ShapeMatching}.

The next sections of this paper present a detailed and technical description of dipole calculi.
In Section~\ref{sec:RepresentationAndRelationAlgebras} we introduce the relations of the dipole
calculi and revisit the theory of relation algebras and non-associate algebras underlying
qualitative spatial reasoning. Furthermore, we investigate quotient of calculi on a general level as well as for the
dipole calculi.
Section~\ref{condensedsemantics} provides a condensed semantics for the dipole calculus.
A condensed semantics, as we name it, provides spatial domain knowledge to the calculus in the form of
an abstract symbolic model of a specific fragment of the spatial domain.
In this model, possible configurations of very few of the basic spatial entities of a calculus are enumerated.
In our case, we
use orbits in the affine group $\mathbf{GA}(\R^2)$. This provides a useful
abstraction for reasoning about qualitatively different configurations in Euclidean space.
We use affine geometry at a rather elementary level and appeal to pictures instead
of complete analytic arguments, whenever it is easy to fill in the details -- however,
at key points in the argument, careful analytic treatments are provided. Further, we
calculate the composition tables for the dipole calculi using the condensed semantics
and we investigate properties of the composition. In Section~\ref{reasoning}
we answer the question whether the standard constraint resoning method algebraic closure decides consistency for the dipole
calculi.
After the presentation of the technical details of dipole calculi and some of their properties,
a sample application of dipole calculi using a spatial reasoning toolbox
is presented in Section~\ref{application}.
The example uses the reasoning capabilities of a dipole calculus based on constraint reasoning.
Our paper ends with a summary and conclusion and pointers to future work.

\section{Representation of Dipole Relations and Relation Algebras}
\label{sec:RepresentationAndRelationAlgebras}
In this section, we first present a set of spatial relations between dipoles, then
variants of this set of spatial relations. The final subsection shows mathematical
structures for constraint reasoning about dipole relations.

\subsection{Basic Representation of Dipole Relations\label{basic}}

The basic entities we use are dipoles, i.e.\ oriented line segments
formed by a pair of two points, a start point and an end
point.
Dipoles are denoted by $A, B, C,\ldots$, start points by ${\bf s}_A$ and end points by ${\bf e}_A$,
respectively (see Fig.~\ref{left}).
These dipoles are used for representing spatial objects with an
intrinsic orientation. Given a set of dipoles, it is possible to
specify many different relations of different arity, e.g. depending
on the length of dipoles, on the angle between different dipoles, or on the
dimension and nature of the underlying space.
When examining different relations, the goal is to obtain a set of
jointly exhaustive and pairwise disjoint {\em atomic} or {\em base} relations, such that exactly one relation
holds between any two dipoles. The elements of the powerset of the {\em base} relations are
called \emph{general} relations. These are used to express uncertainty about the relative
position of dipoles. If these relations
form an algebra which fulfills certain requirements,
it is possible to apply standard constraint-based reasoning
mechanisms that were originally developed for temporal reasoning
\cite{allen83} and
that have also proved valuable for spatial reasoning.

\begin{figure}[thb]
\begin{center}
\includegraphics[keepaspectratio, scale=0.85]{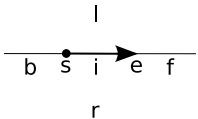}
\caption{\label{fig:finerelations} Dipole-point relations (= $\LR$ relations)}
\end{center}
\end{figure}

So as to enable efficient reasoning, an attempt should be made to keep the
number of different base relations relatively small.
For this reason, we will restrict ourselves to using two-dimensional
continuous space for now, in particular ${\mathbb{R}}^2$, and distinguish the location and
orientation of different dipoles only according to
a small set of seven different dipole-point relations.
We distinguish between whether a point
lies to the left, to the right, or at one of five qualitatively different
locations on the straight line that passes through the
corresponding dipole
\footnote{In his introduction of a set of qualitative spatial relations between oriented line
segments, Schlieder \cite{Schlieder95} mainly focused on configurations in which
no more than two end or start points were on the same straight line (e.g. all points
were in general position).
However, in many domains, we may wish
to represent spatial arrangements
in which more than two start or end points of dipoles are on a
straight line.}.
The corresponding regions are shown on Fig.~\ref{fig:finerelations}.
A corresponding set of relations between three points was proposed by Ligozat \cite{Ligozat1993} under the name flip-flop calculus and
later extended to the $\LR$ calculus \cite{ScivosN-a:04-finest}\footnote{The $\LR$ calculus also features the relations
$\relnsp{dou}$ and $\relnsp{tri}$ for both reference points or all points being equal, respectively. These cases are not
possible for dipoles, since the start and end points cannot coincide by definition.}.

Then these dipole-point relations
describe cases when
the point is: to the left of the dipole ($\rm l$);
to the right of the dipole ($\rm r$);
straight behind the dipole ($\rm b$);
at the start point of the dipole ($\rm s$);
inside the dipole ($\rm i$);
at the end of the dipole ($\rm e$);
or straight in front of the dipole ($\rm f$).
For example, in Fig.~\ref{left},
${\bf s}_B$ lies to the
left of $A$, expressed as $A \;{\rm l}\; {\bf s}_B$.
Using these seven possible relations between a dipole and a point,
the relations between two dipoles may be specified according to the
following four relationships:
\begin{displaymath}
  A \;{\rm R_1}\; {\bf s}_B \wedge A \;{\rm R_2}\; {\bf
e}_B \wedge B \;{\rm R_3}\; {\bf s}_A \wedge B \;{\rm R_4}\; {\bf e}_A,
\end{displaymath}
where ${\rm R_i} \in \left\{\rm l,r,b,s,i,e,f\right\}$ with $1 \le i \le 4$.
Theoretically, this gives us 2401 relations, out of which 72 relations are
geometrically possible, see Prop.~\ref{prop:DRArelations} below.
They are listed on Fig.~\ref{fig:atomicrel}.

\begin{figure}[thb]
\begin{center}
\includegraphics[keepaspectratio, scale=0.85]{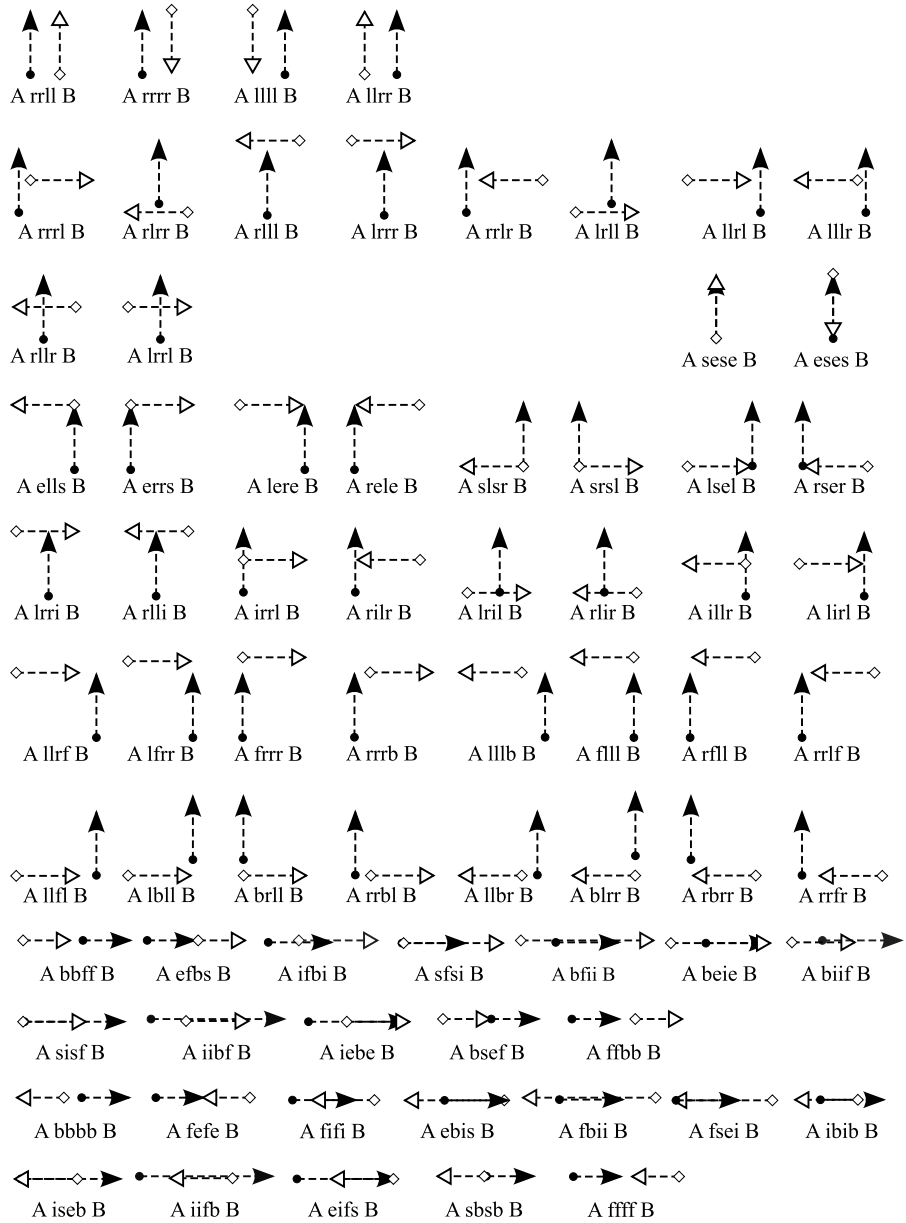}
\caption{\label{fig:atomicrel} The 72 atomic relations of the $\DRAf$ calculus.
In the dipole calculus, orthogonality is not defined, although the graphical representation may suggest this.}
\end{center}
\end{figure}

We introduce an operator that constructs a
relation between two dipoles out of four dipole-point-relations:
\FloatBarrier
\begin{defn}\sloppypar
The operator $\varrho$ takes the four $\LR$ relations between the start and
end points of two dipoles and constructs a relation between dipoles. It is
defined as the textual concatenation:
 $\varrho({\rm R_1}, {\rm R_2}, {\rm R_3},{\rm R_4}) = {\rm R_1 R_2 R_3 R_4}$.
By $\tau_i$ with $1 \le i \le 4$, we denote the projections to components
of the relations between dipoles, where
the identities
$\varrho(\tau_1 {\rm R}, \tau_2 {\rm R}, \tau_3 {\rm R}, \tau_4 {\rm R}) = R$
and $\tau_i \circ \varrho({\rm R_1}, {\rm R_2}, {\rm R_3}, {\rm R_4}) = R_i$
 hold.
\end{defn}

The relations that have been introduced above in an informal way can also be
defined in an algebraic way.
Every dipole $D$ on the plane
${\mathbb{R}}^2$ is an ordered pair
of two points ${\bf s}_D$ and ${\bf e}_D$, each of them being represented by its
Cartesian coordinates
$x$ and $y$, with
$x, y \in {\mathbb{R}}$ and
${\bf s}_D \not= {\bf e}_D$.
\begin{displaymath}
D = \left( {\bf s}_D, {\bf e}_D \right) ,   \qquad
{\bf s}_D  = \left( ({\bf s}_D)_x , ({\bf s}_D)_y \right)
\end{displaymath}

The basic relations are
then described
by equations with the coordinates as variables. The set of
solutions for a
system of equations describes all the possible coordinates for these points.
One first such specification was presented in Moratz et. al. \cite{Moratz00_QSRwithLineSegs}.

\FloatBarrier
\subsection{Several Versions of Sets of Dipole Base Relations}
\label{DRA-versions}

It is an unrealistic goal to provide a single set of qualitative base relations
which is suitable for all possible contexts.
In general, the desired granularity of a representation framework
depends on the specific application \cite{Hobbs85granularity}.
A coarse granularity only needs a small set of base relations.
Finer granularity can lead to a large number of base relations.
If it is desired to apply a spatial calculus to a problem, it is therefore advantageous when
a choice can be made between several versions of sets of base relations.
Then a calculus may be selected
which only has the necessary number of base relations
and thus has less representation complexity
but is fine-grained enough
to solve the particular spatial reasoning problem.
Focussing on the smallest number of base relations also fits better with the principle
of a vocabulary of concepts which is compatible with linguistic principles
\cite{MoratzTenbrink2006,moratz05_QSR_RPP}.
For this purpose,
several versions of sets of dipole base relations can be constructed based on the base relation set
of $\DRAf$.

In their paper on customizing spatial and temporal calculi, Renz and Schmid \cite{RenzSchmid2007}
investigated different methods for deriving variants of a given calculus that have better-suited
granularity for certain tasks. In the first variant, unions of base relations or so-called macro
relations were used as base relations. In the second variant, only a subset of base relations was
used as a new set of base relations.
In his pioneering work on dipole relations,
Schlieder \cite{Schlieder95} introduced a set of base relations in which no more than two
start or end points were on the same straight line.
As a result, only a subset of the $\DRAf$ base relations is
used,  which corresponds to Renz' and Schmid's second variant of
methods for deriving new base relation sets for qualitative calculi.
We refer to a calculus based on these base relations as $\DRAlr$ (where lr stands for left/right).
The following base relations are part of $\DRAlr$:
rrrr, rrll,
llrr, llll,
rrrl, rrlr, rlrr, rllr, rlll, lrrr, lrrl, lrll, llrl, lllr.

Moratz et al. \cite{Moratz00_QSRwithLineSegs} introduced an extension of
$\DRAlr$ which adds relations for representing polygons and polylines.
In this extension, two start or end points can share an identical location.
While in this calculus, three points at different locations cannot belong to the same straight line.
This subset of $\DRAf$ was named $\DRAc$
($c$ refers to {\it coarse}, $f$ refers to {\it fine}).
The set of base relations of $\DRAc$ extends the base relations of
$\DRAlr$
with the following relations:
ells, errs, lere, rele, slsr, srsl, lsel, rser, sese, eses.

Another method for deriving a new set of base relations from an existing set
merges unions of base relations to new base relations.
At a symbolic level, sets of base relations are used to form new
base relations.
In the context of $\DRAf$, this is done as shown in Fig.~\ref{fig:DRAop}
(the meaning of the names of the new base relations is explained in
the following paragraphs).

\begin{figure}
\begin{eqnarray*}
{\rm
\{ llll,\; lllb,\; lllr,\; lrll,\; lbll \}
}
& \mapsto & {\rm
LEFTleft
}
\\
{\rm
\{ ffff,\; eses,\; fefe,\; fifi,\; ibib,\; fbii,\; fsei,\; ebis,\; iifb,\;
eifs,\; iseb\}
}
& \mapsto & {\rm
FRONTfront
}
\\
{\rm
\{ bbbb \}
}
& \mapsto & {\rm
BACKback
}
\\
{\rm
\{ llbr \}
}
& \mapsto & {\rm
LEFTback
}
\\
{\rm
\{ llfl,\; lril,\;lsel \}
}
& \mapsto & {\rm
LEFTfront
}
\\
{\rm
\{ llrf,\; llrl,\; llrr,\; lfrr,\; lrrr,\; lere,\; lirl,\; lrri,\; lrrl \}
}
& \mapsto & {\rm
LEFTright
}
\\
{\rm
\{ rrrr,\; rrrl,\; rrrb,\; rbrr,\; rlrr \}
}
& \mapsto & {\rm
RIGHTright
}
\\
{\rm
\{ rrll,\; rrlr,\; rrlf,\; rlll,\; rfll,\; rllr,\; rele,\;
rlli,\; rilr \}
}
& \mapsto & {\rm
RIGHTleft
}
\\
{\rm
\{ rrbl \}
}
& \mapsto & {\rm
RIGHTback
}
\\
{\rm
\{ rrfr,\; rser,\; rlir \}
}
& \mapsto & {\rm
RIGHTfront
}
\\
{\rm
\{ ffbb,\; efbs,\; ifbi,\; iibf,\; iebe \}
}
& \mapsto & {\rm
FRONTback
}
\\
{\rm
\{ frrr,\; errs,\; irrl \}
}
& \mapsto & {\rm
FRONTright
}
\\
{\rm
\{ flll,\; ells,\; illr \}
}
& \mapsto & {\rm
FRONTleft
}
\\
{\rm
\{ blrr  \}
}
& \mapsto & {\rm
BACKright
}
\\
{\rm
\{ brll \}
}
& \mapsto & {\rm
BACKleft
}
\\
{\rm
\{ bbff,\; bfii,\; beie,\; bsef,\; biif \}
}
& \mapsto & {\rm
BACKfront
}
\\
{\rm
\{ slsr \}
}
& \mapsto & {\rm
SAMEleft
}
\\
{\rm
\{ sese,\; sfsi,\; sisf \}
}
& \mapsto & {\rm
SAMEfront
}
\\
{\rm
\{ sbsb  \}
}
& \mapsto & {\rm
SAMEback
}
\\
{\rm
\{ srsl  \}
}
& \mapsto & {\rm
SAMEright
}
\\
\end{eqnarray*}
\caption{Mapping from $\DRAf$ to $\DRAop$ relations\label{fig:DRAop}}
\end{figure}

\begin{figure}[htb]
\begin{center}
\includegraphics[keepaspectratio, scale=0.85]{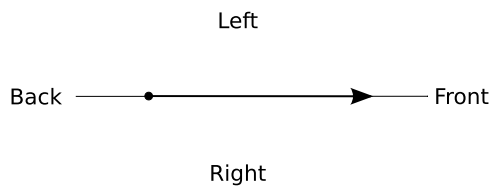}
\caption{\label{Regions} An oriented point and its qualitative spatial relative directions}
\end{center}
\end{figure}

\begin{figure}[htb]
\begin{center}
\includegraphics[keepaspectratio, scale=0.85]{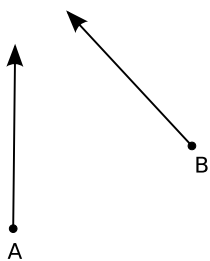}
\caption{\label{Configuration} Qualitative spatial relation
between two oriented points at different positions.
The qualitative spatial relation depicted here is $A$~RIGHTleft~$B$.}
\end{center}
\end{figure}

\begin{figure}[htb]
\begin{center}
\includegraphics[keepaspectratio, scale=0.85]{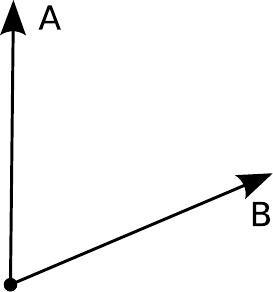}
\caption{\label{Configuration2} Qualitative spatial relation between two
  oriented points located at the same position.
The qualitative spatial relation depicted here is $A$~SAMEright~$B$.}
\end{center}
\end{figure}

$\DRAop$
is the name of the calculus which has the set of base relations
listed in Fig.~\ref{fig:DRAop}.
In \cite{Moratz06_ECAI}, a calculus $\OPRA_1$
which is isomorphic\footnote
{Since we have not introduced operations on QSR calculi yet,
we explain the details of the correspondence between
$\DRAop$ and
$\OPRA_1$ later in our paper, see
Prop.~\ref{prop:DRAop-OPRA-1}.}
to $\DRAop$ is defined in a complementary geometric way.
The transition from
oriented line segments with
well-defined lengths to line segments with infinitely small lengths is the core idea of this
geometric model.
In this conceptualization, the length
of objects no longer has any significance.
Thus, only the direction of the objects is modeled \cite{Moratz06_ECAI}.
These objects can then be conceptualized as oriented points.
An {\em o-point}, our term for an oriented point, is specified as a pair of a point with
a direction in the 2D-plane.
Then the "op" in the symbol $\DRAop$
stands for oriented points.
A single o-point induces the sectors depicted in
\nolinebreak{Fig.~\ref{Regions}}.
 ``Front'' and ``Back'' are linear sectors. ``Left'' and ``Right'' are
half-planes. The position of the point itself is denoted as ``Same''.
A qualitative spatial relative position relation between two o-points is represented by
the sector in which the second o-point lies in relation to the first one and by
the sector in which the first o-point lies in relation to the second one.
For the general case of two points having different positions, we use the concatenated string
of both sector names as the relation symbol. Then the configuration shown in
\nolinebreak{Fig.~\ref{Configuration}}
is expressed by the relation $A \; {\rm RIGHTleft} \; B$. If both points share the same
position, the relation symbol starts with the word ``Same'' and the second substring denotes
the direction of the second o-point relative to the first one
as shown in Fig.~\ref{Configuration2}.

Altogether we obtain 20 different atomic relations
(four times four general relations plus four with the
oriented points at the same positions).
The relation {\rm SAMEfront} is the identity
relation.
$\DRAop$ has fewer base relations and therefore is more compact than
$\DRAf$.
Focussing on a smaller set of base relations in this case also fits better with the principle
of using a vocabulary of concepts which is compatible with linguistic principles
\cite{MoratzTenbrink2006,moratz05_QSR_RPP}.
For this reason, many $\DRAop$ base relations
have simple corresponding linguistic expressions.
For example, the qualitative spatial configuration represented as
$A \; {\rm LEFTfront} \; B$ can be translated into the natural language expression
"B is left of A and A is in front of B". A and B in this example
would be oriented objects with an intrinsic front
like two cars A and B in a parking lot.
However, in general, the correspondence between QSR expressions and their linguistic counterparts
is only an approximation
\cite{MoratzTenbrink2006,moratz05_QSR_RPP}.

The two methods for deriving new sets of base relations which we applied above reduce the number of
base relations. Conversely, other methods extend the number of base relations.
For example, Dylla and Moratz \cite{DyllaMoratzSC04} have observed that
$\DRAf$ may not be sufficient for robot
navigation tasks, because the
dipole configurations that are
pooled in certain base relations are too diverse.
Thus, the representation has been extended with additional
orientation knowledge and
a more fine-grained
$\DRAfp$ calculus
with additional orientation
distinctions has been derived. It has slightly more base relations.

\begin{figure}[!h]
\begin{center} \includegraphics[keepaspectratio, scale=0.85]{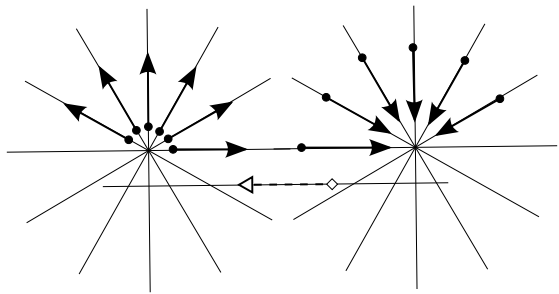}
\caption{Pairs of dipoles subsumed
by the same relation}
\label{StroboExample}
\end{center}
\end{figure}

\begin{figure}[ht]
\begin{center} \includegraphics[keepaspectratio, scale=0.55]{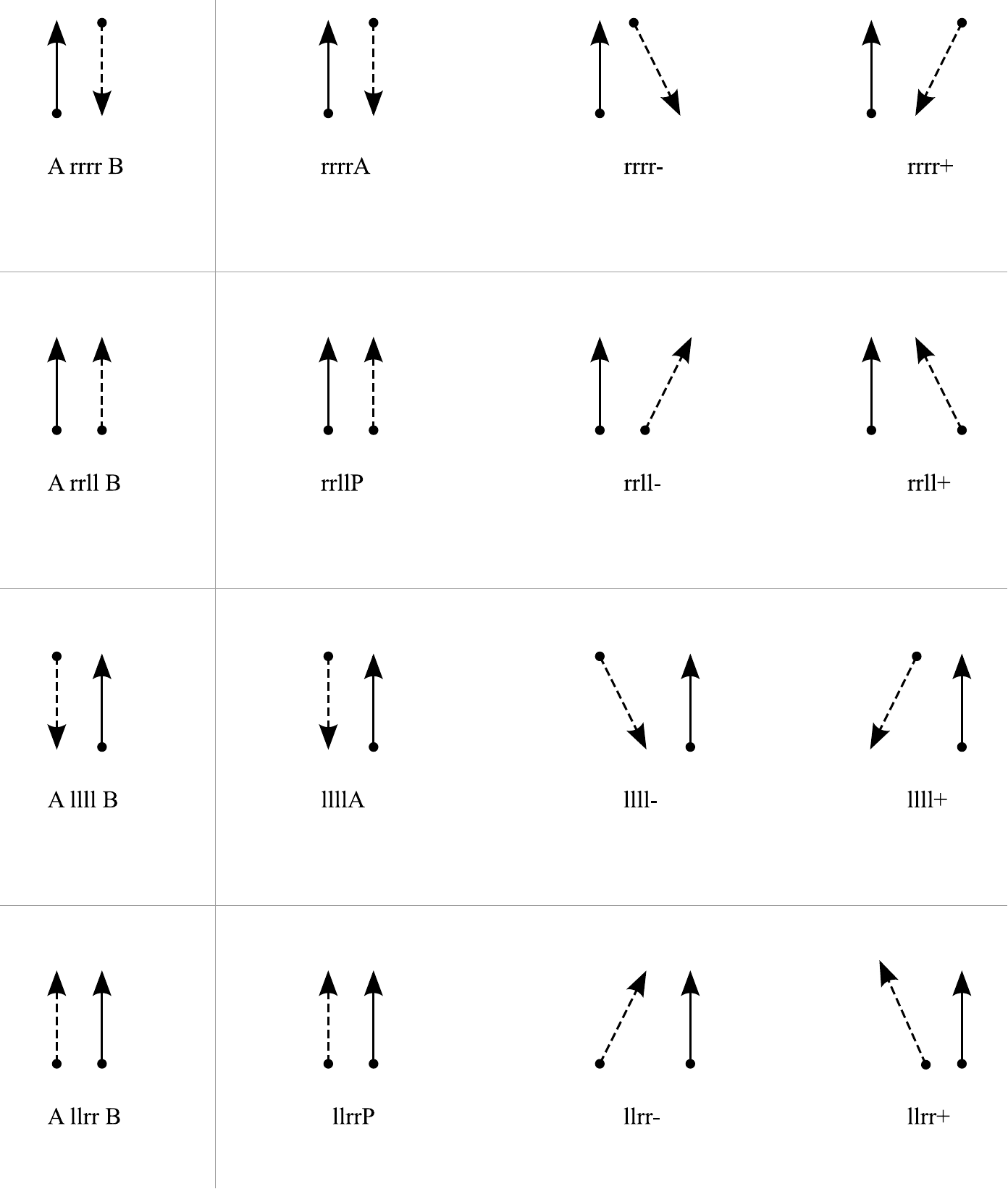}
\caption{Refined base relations in $\DRAfp$}
\label{SubRels}
\end{center}
\end{figure}

The large configuration space for the \relnsp{rrrr} relation is
visualized in Fig.~\ref{StroboExample}.
The other analogous relations which are extremely coarse are
\relnsp{llrr}, \relnsp{rrll} and \relnsp{llll}.
In many applications, this unwanted coarseness of four relations can
lead to problems\footnote{An investigation by Dylla and Moratz into the
first cognitive robotics applications of dipole relations integrated
in situation calculus \cite{DyllaMoratzSC04} showed
that the coarseness of $\DRAf$ compared to
$\DRAfp$ would indeed lead
to rather meandering paths for a spatial agent.}.
Therefore, we introduce an additional qualitative feature
by considering the angle spanned by the two
dipoles.
This gives us an important additional
distinguishing feature with four distinctive values.
These qualitative distinctions are parallelism (P) or anti-parallelism (A) and
mathematically
positive and negative angles between $A$ and $B$, leading to three
refining relations for each of the four above-mentioned
relations (Fig. \ref{SubRels}).

\FloatBarrier

We call this algebra $\DRAfp$ as it is an extension of the fine-grained
relation algebra $\DRAf$ with additional distinguishing features
due to ``parallelism''.
For the other relations, a '$+$' or '$-$', 'P' or 'A' respectively, is already
determined by the original base relation and does not have to be mentioned
explicitly. These base relations then have the same relation symbol as in
$\DRAf$.

The introduction of parallelism into dipole calculi not only has benefits
in certain applications. The algebraic features also benefit from this extension
(see Sect.~\ref{sec:props}).
For analogous reasons, a derivation of
$\DRAfp$
yields an oriented point calculus which explicitly contains the feature of parallelism,
which is isomorphic to the $\OPRA_1^*$ calculus\cite{dylladis}.
This calculus
$\DRAopp$ (opp stands for oriented points and parallelism)
has the same base relations as
$\DRAop$ with the exception of the relations
${\rm RIGHTright }$, ${\rm RIGHTleft }$, ${\rm LEFTleft }$, and ${\rm LEFTright }$.
The transformation from $\DRAfp$ to $\DRAopp$ is shown in Fig.~\ref{fig:DRAopp}.

\begin{figure}
\begin{eqnarray*}
{\rm
\{ llllA \}
}
& \mapsto & {\rm
LEFTleftA
}
\\
{\rm
\{ llll+,\; lllb+,\; lllr+ \}
}
& \mapsto & {\rm
LEFTleft+
}
\\
{\rm
\{ lrll,\; lbll \}
}
& \mapsto & {\rm
LEFTleft-
}
\\
{\rm
\{ ffff,\; eses,\; fefe,\; fifi,\; ibib,\; fbii,\; fsei,\; ebis,\; iifb,\;
eifs,\; iseb\}
}
& \mapsto & {\rm
FRONTfront
}
\\
{\rm
\{ bbbb \}
}
& \mapsto & {\rm
BACKback
}
\\
{\rm
\{ llbr \}
}
& \mapsto & {\rm
LEFTback
}
\\
{\rm
\{ llfl,\; lril,\;lsel \}
}
& \mapsto & {\rm
LEFTfront
}
\\
{\rm
\{ llrrP \}
}
& \mapsto & {\rm
LEFTrightP
}
\\
{\rm
\{ llrr+ \}
}
& \mapsto & {\rm
LEFTright+
}
\\
{\rm
\{ llrf,\; llrl,\; llrr-,\; lfrr,\; lrrr,\; lere,\; lirl,\; lrri,\; lrrl \}
}
& \mapsto & {\rm
LEFTright-
}
\\
{\rm
\{ rrrrA \}
}
& \mapsto & {\rm
RIGHTrightA
}
\\
{\rm
\{ rrrr+,\; rbrr,\; rlrr \}
}
& \mapsto & {\rm
RIGHTright+
}
\\
{\rm
\{ rrrr-,\; rrrl,\; rrrb \}
}
& \mapsto & {\rm
RIGHTright-
}
\\
{\rm
\{ rrllP \}
}
& \mapsto & {\rm
RIGHTleftP
}
\\
{\rm
\{ rrll+,\; rrlr,\; rrlf,\; rlll,\; rfll,\; rllr,\; rele,\;
rlli,\; rilr \}
}
& \mapsto & {\rm
RIGHTleft+
}
\\
{\rm
\{ rrll- \}
}
& \mapsto & {\rm
RIGHTleft-
}
\\
{\rm
\{ rrbl \}
}
& \mapsto & {\rm
RIGHTback
}
\\
{\rm
\{ rrfr,\; rser,\; rlir \}
}
& \mapsto & {\rm
RIGHTfront
}
\\
{\rm
\{ ffbb,\; efbs,\; ifbi,\; iibf,\; iebe \}
}
& \mapsto & {\rm
FRONTback
}
\\
{\rm
\{ frrr,\; errs,\; irrl \}
}
& \mapsto & {\rm
FRONTright
}
\\
{\rm
\{ flll,\; ells,\; illr \}
}
& \mapsto & {\rm
FRONTleft
}
\\
{\rm
\{ blrr  \}
}
& \mapsto & {\rm
BACKright
}
\\
{\rm
\{ brll \}
}
& \mapsto & {\rm
BACKleft
}
\\
{\rm
\{ bbff,\; bfii,\; beie,\; bsef,\; biif \}
}
& \mapsto & {\rm
BACKfront
}
\\
{\rm
\{ slsr \}
}
& \mapsto & {\rm
SAMEleft
}
\\
{\rm
\{ sese,\; sfsi,\; sisf \}
}
& \mapsto & {\rm
SAMEfront
}
\\
{\rm
\{ sbsb  \}
}
& \mapsto & {\rm
SAMEback
}
\\
{\rm
\{ srsl  \}
}
& \mapsto & {\rm
SAMEright
}
\\
\end{eqnarray*}
\caption{Mapping from $\DRAfp$ to $\DRAopp$ relations\label{fig:DRAopp}}
\end{figure}

Again, the mathematical properties of the oriented point calculus can be derived
from the corresponding dipole calculus, see Corollary~\ref{cor:DRAoppstrong}.

\subsection{Relation Algebras for Spatial Reasoning}
\label{ConstraintPropagationEtc}

Standard methods developed for finite domains generally do not apply
to constraint reasoning over infinite domains.  The theory of
relation algebras \cite{Ladkin94,Maddux2006} allows for a purely
symbolic treatment of constraint satisfaction problems involving relations over
infinite domains.  The corresponding constraint reasoning techniques were originally
introduced for temporal reasoning \cite{allen83} and later
proved to be valuable for spatial reasoning
\cite{renz99_RCC_Complexity,Isli00_2DOriOrdering}.  The central data
for a calculus is given by:
\begin{itemize}
\item a list of (symbolic names for) \emph{base
relations}, which are interpreted as relations over some domain,
having the crucial properties of \emph{pairwise disjointness} and \emph{joint exhaustiveness} (a general relation is then simply a set of base relations).
\item a table for the computation of the \emph{converses} of relations.
\item a table for the computation of the \emph{compositions} of relations.
\end{itemize}
 Then, the path consistency algorithm
\cite{DBLP:journals/isci/Montanari74} and backtracking techniques
\cite{DBLP:journals/jair/BeekM96} are the tools used to tackle the
problem of consistency of constraint networks and related problems.
These algorithms have been implemented in both generic reasoning tools \texttt{GQR}
 \cite{r4:aaai08ws} and \texttt{SparQ}
\cite{DBLP:conf/spatialCognition/WallgrunFWDF06}.  To integrate a new
calculus into these tools, only a list of base relations and
tables for compositions and converses really need to be provided.
Thereby, the qualitative reasoning facilities of these tools become
available for this calculus.\footnote{With more
information about a calculus, both of the tools can provide functionality that
goes beyond simple qualitative reasoning for constraint calculi.}
Since the
compositions and converses of general relations
can be reduced to
compositions and converses of base relations, these tables only need to
be given for base relations. Based on these tables, the tools provide a
means to approximate the consistency of constraint networks, list all
their atomic refinements, and more.

Let $b$ be the name of a base relation, and let $R_b$ be its set-theoretic
extension. The converse $\inv{(R_b)}=\{(x,y) | (y,x)\in R_b\}$
is often itself a base relation and is denoted by $\inv{b}$\footnote{In 
Freksa's double-cross calculus \cite{freksa_cosit0} the converses are
not necessarily base-relations, but for the calculi that we investigate 
this property holds.}.
In the dipole calculus, it is obvious that the converse of a relation can easily be
computed by exchanging the first two and second two letters of the name
of a relation, see Table \ref{TABLE:1}.
Also for the dipole calculus
$\DRAfp$
with additional orientation
distinctions a simple rule exchanges
'$+$' with '$-$', and vice versa.'P' and 'A' are invariant with respect to the converse
operation.
Since base relations generally are not closed under composition,
this operation is approximated by a \emph{weak composition}:
$$b_1;b_2=\{b \mid (R_{b_1}\circ R_{b_2})\cap R_b \not= \emptyset\}$$
where $R_{b_1}\circ R_{b_2}$ is the usual set theoretic composition
$$R_{b_1}\circ R_{b_2} = \{(x,z) | \exists y\,.\,(x,y)\in R_{b_1}, (y,z)\in R_{b_2}\}$$
The composition is said to be \emph{strong} if
$R_{b_1;b_2}=R_{b_1}\circ R_{b_2}$. Generally, $b_1;b_2$
over-approximates the set-theoretic composition.\footnote{The $R_\_$
operation naturally extends to sets of (names of) base relations.}
Computing the composition table
is much harder
 and will be the subject of Section ~\ref{condensedsemantics}.

\begin{table}
\begin{center}
\begin{tabular}{|l|l|l|l|l|l|l|l|l|}\hline
  $R$          & \relnsp{rrrr } & \relnsp{rrrl} & \relnsp{rrlr} & \relnsp{rrll} & \relnsp{rlrr} & \relnsp{rllr} & \relnsp{rlll} & \relnsp{lrrr} \\ \hline
  $R^{\smile}$ & \relnsp{rrrr } & \relnsp{rlrr} & \relnsp{lrrr} & \relnsp{llrr} & \relnsp{rrrl} & \relnsp{lrrl} & \relnsp{llrl} & \relnsp{rrlr} \\ \hline
\end{tabular}\\[2ex]
\end{center}
\caption{\label{TABLE:1}The converse ($\smile$) operation of
$\DRAf$ can be reduced to a simple permutation.}
\end{table}
\sloppypar
The mathematical background of composition in table-based reasoning is given by
the theory of \emph{relation algebras} \cite{Maddux2006,Renz2007}.
For many calculi, including the dipole calculus, a slightly weaker notion is needed, namely
that of a \emph{non-associative algebra} \cite{ligozatR-a:04-what}.
These algebras treat spatial relations as abstract entities that
can be combined by certain operations and governed by certain equations.
This allows algorithms and tools to operate at a symbolic level,
in terms of (sets of) base relations instead of their set-theoretic extensions.

\begin{defn}[\cite{ligozatR-a:04-what}]\label{def:non-assoc-alg}
\sloppypar
A \emph{non-associative algebra} $A$ is a tuple $A = (A,+,-,\cdot, 0, 1, ;, \inv{}, \Delta)$
such that:
\begin{enumerate}
\item $(A,+,-,\cdot, 0, 1)$ is a Boolean algebra.
\item $\Delta$ is a constant, $\inv{}$ a unary and $;$ a binary operation
such that, for any $a, b, c \in A$:
{\small
$$\begin{array}{lll}
(a)~ \inv{(\inv{a})} = a&
(b)~ \Delta ; a = a ; \Delta = a&
(c)~ a ; (b + c) = a ; b + a ; c\\
(d)~ \inv{(a + b)}= \inv{a}+ \inv{b}&
(e)~ \inv{(a - b)}= \inv{a}- \inv{b}&
(f)~ \inv{(a ; b)} = \inv{b} ; \inv{a}\\
\multicolumn{2}{l}
{(g)~ (a ; b) \cdot \inv{c} = 0 \mbox{ if and only if } (b ; c)\cdot \inv{a} = 0}
\end{array}$$}
\end{enumerate}
A non-associative algebra is called a \emph{relation algebra},
if the composition $;$ is associative.
\end{defn}

The elements of such an algebra will be called (abstract) relations.
We are mainly interested in finite non-associative algebras that are
\emph{atomic}, which means that there is a set of pairwise disjoint minimal relations,
called base relations, and all relations can be obtained as unions
of base relations. Then, the following fact is well-known and easy to prove:
\begin{prop}
An atomic non-associative algebra is uniquely determined by
its set of base relations, together with the converses and compositions
of base relations. (Note that the composition of two base
relations is in general not a base relation.)
\end{prop}

\begin{exmp}
The powerset of the 72 $\DRAf$ base relations forms a boolean algebra.
The relation \relnsp{sese} is the identity relation. The converse
and (weak) composition are defined as above.
We denote the resulting non-associative algebra by $\DRAf$.
The algebraic laws follow from general results about so-called
partition schemes, see \cite{ligozatR-a:04-what}.
Similarly, we obtain a non-associative algebra $\DRAfp$.

However, we do not obtain a non-associative algebra for $\DRAc$,
because $\DRAc$ does not provide a jointly exhaustive set of base
relations over the Euclidean plane. This leads to the lack of an 
identity relation, and more
severely, weak composition does not lead to an over-approximation (nor
an under-approximation) of set-theoretic composition, because e.g.\
\relnsp{ffbb} is missing from the composition of \relnsp{llll} with
itself. In particular, we cannot expect the algebraic laws of a
non-associative algebra to be satisfied.
\end{exmp}

For non-associative algebras, we define lax homomorphisms which allow
for both the embedding of a calculus into another one, and the
embedding of a calculus into its domain.

\begin{defn}[Lax homomorphism]
  Given non-associative algebras $A$ and $B$, a \emph{lax homomorphism}
  is a homomorphism $\mathop{\mathrm{h}}: A \longrightarrow B$ on the
  underlying Boolean algebras such that:
  \begin{itemize}
    \item $\mathop{\mathrm{h}}(\Delta_A) \ge \Delta_B$
    \item $\mathop{\mathrm{h}}(a^{\smile}) = \mathop{\mathrm{h}}(a)^{\smile}$
      for all $a \in A$
    \item $\mathop{\mathrm{h}}(a;b) \ge
      \mathop{\mathrm{h}}(a);\mathop{\mathrm{h}}(b)$
      for all $a, b \in A$
  \end{itemize}
\end{defn}

Dually to lax homomorphisms, we can define oplax homomorphisms\footnote
{The terminology is motivated by that for monoidal functors.}, which
enable us to define projections from one calculus to another.

\begin{defn}[Oplax homomorphism]
  Given non-associative algebras $A$ and $B$, an \emph{oplax homomorphism}
  is a homomorphism $\mathop{\mathrm{h}}: A \longrightarrow B$ on the
  underlying Boolean algebras such that:
  \begin{itemize}
    \item $\mathop{\mathrm{h}}(\Delta_A) \le \Delta_B$
    \item $\mathop{\mathrm{h}}(a^{\smile}) = \mathop{\mathrm{h}}(a)^{\smile}$
      for all $a \in A$
    \item $\mathop{\mathrm{h}}(a;b) \le
      \mathop{\mathrm{h}}(a);\mathop{\mathrm{h}}(b)$
      for all $a, b \in A$
  \end{itemize}
\end{defn}

A proper homomorphism (sometimes just called a homomorphism)
of non-associative algebras is a homomorphism that is
lax and oplax at the same time; the above inequalities then turn into equations.

An important application of homomorphisms is their use in the
definition of qualitative calculus.  Ligozat and Renz
\cite{ligozatR-a:04-what} define a qualitative calculus in terms of
a so-called \emph{weak representation} \cite{DBLP:conf/cosit/Ligozat05}:
\begin{defn}[Weak representation]
A weak representation is
an identity-preserving (i.e.\ $\mathop{\mathrm{h}}(\Delta_A) = \Delta_B$)
lax homomorphism $\varphi$ from a (finite atomic)
non-associative algebra into the relation algebra of a domain $\D$. The latter is given by the canonical relation algebra on the
powerset ${\cal P}(\D\times\D)$, where identity,
converse and composition (as well as the Boolean algebra operations)
are given by their set-theoretic interpretations.
\end{defn}

\begin{exmp}
Let $\dipole$ be the set of all dipoles in $\R^2$. Then the weak representation
of $\DRAf$ is
the lax homomorphism $\rf:\DRAf \to {\cal P}(\dipole\times\dipole)$ given
by
$$\rf(R)=\{R_b\, |\, b\in R\}.$$
\end{exmp}
We obtain a similar weak representation
$\rfp$ for $\DRAfp$.
  The following is straightforward:
\begin{prop}\label{fact:strong-composition}
A calculus has a strong composition if and only if its weak representation
is a proper homomorphism.
\end{prop}
\begin{pf}
Since a weak representation is identity-preserving, being proper means
that $\varphi(R_1;R_2)=\varphi(R_1)\circ \varphi(R_2)$, which is nothing but
$R_{R_1;R_2}=R_{R_1}\circ R_{R_2}$, which is exactly the strength of the composition.
\end{pf}

The following is straightforward \cite{DBLP:conf/cosit/Ligozat05}:
\begin{prop}
A weak representation $\varphi$ is injective if and only if
$\varphi(b)\not=\emptyset$ for each base relation $b$.
\end{prop}

The second main use of homomorphisms is relating different calculi.
For example,
the algebra over Allen's interval relations \cite{allen83} can be embedded
into $\DRAf$ ($\DRAfp$) via a homomorphism.

\begin{prop}\label{prop:allenDRA}
A homomorphism from Allen's interval algebra to $\DRAf$ ($\DRAfp$)
exists and is given by the following mapping of base relations.
\begin{displaymath}
\begin{array}{rclcrcl}
\qquad \qquad \qquad =  & \mapsto & \relnsp{sese}   & \qquad \qquad &   &         &\\
\relnsp{b}  & \mapsto & \relnsp{ffbb}   & \qquad & \relnsp{bi}  & \mapsto & \relnsp{bbff}   \\
\relnsp{m}  & \mapsto & \relnsp{efbs}   & \qquad & \relnsp{mi}  & \mapsto & \relnsp{bsef}   \\
\relnsp{o}  & \mapsto & \relnsp{ifbi}   & \qquad & \relnsp{oi}  & \mapsto & \relnsp{biif}   \\
\relnsp{d}  & \mapsto & \relnsp{bfii}   & \qquad & \relnsp{di}  & \mapsto & \relnsp{iibf}   \\
\relnsp{s}  & \mapsto & \relnsp{sfsi}   & \qquad & \relnsp{si}  & \mapsto & \relnsp{sisf}   \\
\relnsp{f}  & \mapsto & \relnsp{beie}   & \qquad & \relnsp{fi}  & \mapsto & \relnsp{iebe}
\end{array}
\end{displaymath}
\end{prop}
\begin{pf}
The identity relation $=$ is clearly mapped to the identity relation $\relnsp{sese}$.
For the composition and converse properties, we just inspect the composition and converse tables
for the two calculi.\footnote{This is a (non-circular) forward reference to Section~\ref{condensedsemantics},
where we compute the $\DRAf$ and $\DRAfp$ composition tables.} The mapping of the base-relation is then lifted directly to
a mapping of all relations, where the map is applied component-wise on the
relations. Using the laws of non-associative algebras,
the homomorphism property of these relations follows
from that of the base-relations.
\end{pf}

In cases stemming from the embedding of Allen's Interval Algebra,
the dipoles lie on the same straight lines and have the same
direction. $\DRAf$ and $\DRAfp$ also contain 13 additional relations which
correspond to the case with dipoles lying on a line but facing opposite directions.

As we shall see, it is very useful to extend the notion of homomorphisms
to weak representations:
\begin{defn}
Given weak representations $\varphi:A\to\mathcal{P}(\D\times\D)$
and $\psi:B\to\mathcal{P}(\V\times\V)$, a
\emph{lax (oplax, proper) homomorphism of weak representations}
$(h,\hD):\varphi \to \psi$ is given by
\begin{itemize}
\item a proper homomorphism of non-associative algebras $h:A\to B$, and
\item a map $\hD:\D\to \V$, such that the diagram
\end{itemize}
$$\xymatrix{
A \ar[rr]^{\varphi} \ar[dd]^h &&
\mathcal{P}(\D\times\D) \ar[dd]^{\mathcal{P}(\hD\times \hD)}\\
&&\\
B \ar[rr]^{\psi}&&
\mathcal{P}(\V\times\V)
}$$
commutes laxly (respectively\ oplaxly, properly). Here, lax commutation
means that for all $R\in A$,
$\psi(h(R))\subseteq \mathcal{P}(\hD\times \hD)(\varphi(R))$,
oplax commutation means the same with $\supseteq$, and proper
commutation with $=$.
Note that $\mathcal{P}(\hD\times \hD)$ is the obvious extension of $\hD$
to a function between relation algebras; note that (unless $\hD$
is bijective) this is not even a homomorphism of Boolean algebras
(it may fail to preserve top, intersections and complements),
although it satisfies the oplaxness property (and the laxness property
if $\hD$ is surjective).\footnote{The reader with background in
category theory may notice that the categorically more natural
formulation would use the contravariant powerset functor,
which yields homomorphisms of Boolean algebras. However, the
present formulation fits better with the examples.}
\end{defn}
Note that Ligozat \cite{DBLP:conf/cosit/Ligozat05} defines
a more special notion of morphism between weak representations;
it corresponds to our oplax homomorphism of weak representations
where the component $h$ is the identity.

\begin{exmp}\label{ex:IA-DRAf}
The homomorphism from Prop.~\ref{prop:allenDRA} can be extended
to a proper homomorphism of weak representations by letting
$\hD$ be the embedding of time intervals to dipoles on
the $x$-axis.
\end{exmp}

\begin{exmp}
Let $h$ map each $\DRAfp$ relation to the corresponding $\DRAf$ relation:
  \begin{eqnarray*}
    \relnsp{llll+} &\mapsto& \relnsp{llll} \\
    \relnsp{llll-} &\mapsto& \relnsp{llll} \\
    \relnsp{llllA} &\mapsto& \relnsp{llll} \\
    \relnsp{rrrr+} &\mapsto& \relnsp{rrrr} \\
    \relnsp{rrrr-} &\mapsto& \relnsp{rrrr} \\
    \relnsp{rrrrA} &\mapsto& \relnsp{rrrr} \\
    \relnsp{llrr+} &\mapsto& \relnsp{llrr} \\
    \relnsp{llrr-} &\mapsto& \relnsp{llrr} \\
    \relnsp{llrrP} &\mapsto& \relnsp{llrr} \\
    \relnsp{rrll+} &\mapsto& \relnsp{rrll} \\
    \relnsp{rrll-} &\mapsto& \relnsp{rrll} \\
    \relnsp{rrllP} &\mapsto& \relnsp{rrll} \\
  \end{eqnarray*}
Then $(h,id):\DRAfp\to\DRAf$ is a surjective oplax homomorphism of weak representations.
\end{exmp}
Although this homomorphism of weak representations is surjective,
it is not a quotient in the following sense (and in particular,
it does \emph{not} satisfy Prop.~\ref{prop:preserveStrength},
as will be shown in Sections~\ref{sec:weak-comp} and~\ref{sec:strong-comp}).
\begin{defn}
A homomorphism of non-associative algebras is said to be a
\emph{quotient homomorphism}\footnote{Maddux \cite{Maddux2006}
does not have much to say on this subject; instead, we suggest consulting a textbook
on universal algebra, e.g.\ \cite{Gratzer79}.} if it is proper and surjective.
A (lax, oplax or proper) homomorphism of weak representations is a
quotient homomorphism if it is surjective in both components.
\end{defn}

The easiest way to form a quotient of a weak representation is via
an equivalence relation on the domain:
\begin{defn}\label{def:quotient}
Given a weak representation $\varphi:A\to\mathcal{P}(\D\times\D)$
and an equivalence relation $\sim$ on $\D$, we obtain the
\emph{quotient representation} $\varphisim$ as follows:
$$\xymatrix{
A \ar[rr]^{\varphi} \ar[dd]^{q_A} &&
\mathcal{P}(\D\times\D) \ar[dd]^{\mathcal{P}(q\times q)}\\
&&\\
\Asim \ar[rr]^{\varphisim}&&
\mathcal{P}(\Dsim\times\Dsim)
}$$
\begin{itemize}
\item Let $q:\D\to\Dsim$ be the factorization of $\D$ by $\sim$;
\item $q$ extends to relations:  $\mathcal{P}(q \times q):\mathcal{P}(\D\times\D)\to\mathcal{P}(\Dsim\times\Dsim)$;
\item let $\sim_A$ be the congruence relation on $A$ generated by
$$\mathcal{P}(q \times q)(\varphi(b_1))\cap\mathcal{P}(q \times q)(\varphi(b_2))\not=\emptyset\ \Rightarrow\ b_1\sim_A b_2$$
for base relations $b_1,b_2\in A$. $\sim$ is called \emph{regular w.r.t.\ $\varphi$} if $\sim_A$ is the
 kernel of $\mathcal{P}(q \times q)\circ \varphi$ (i.e.\ the set of all
pairs made equal by $\mathcal{P}(q \times q)\circ \varphi$);
\item let $q_A:A\to \Asim$ be the quotient of $A$ by $\sim_A$ in the
sense of universal algebra \cite{Gratzer79}, which uses proper homomorphisms;
hence, $q_A$ is a proper homomorphism;
\item finally, the function $\varphisim$ is defined as
$$\varphisim(R) = \mathcal{P}(q \times q)(\varphi(q_A^{-1}(R))).$$
\end{itemize}
\end{defn}

\begin{prop}
The function $\varphisim$ defined in Def.~\ref{def:quotient} is
an oplax homomorphism of non-associative algebras.
\end{prop}
\begin{pf}
To show this, notice that an equivalent definition works
on the base relations of $\Asim$:
$$\varphisim(R) = \bigcup_{b\in R}
\mathcal{P}(q \times q)(\varphi(q_A^{-1}(b))).$$
It is straightforward to show that bottom and joins are preserved;
since $q$ is surjective, also top is preserved.\\ Concerning meets,
since general relations in $\Asim$ can be considered to be
sets of base relations, it suffices to show that
$b_1\wedge b_2=0$ implies $\mathcal{P}(q \times q)(\varphi(q_A^{-1}(b_1)))
\cap \mathcal{P}(q \times q)(\varphi(q_A^{-1}(b_2))) = \emptyset$.
Assume to the contrary that $\mathcal{P}(q \times q)(\varphi(q_A^{-1}(b_1)))
\cap \mathcal{P}(q \times q)(\varphi(q_A^{-1}(b_2))) \not= \emptyset$.
Then already $\mathcal{P}(q \times q)(\varphi(b'_1))
\cap \mathcal{P}(q \times q)(\varphi(b'_2)) \not= \emptyset$
for base relations $b'_i\in q_A^{-1}(b_i)$, $i=1,2$. But then
$b'_1 \sim_A b'_2$, hence $q_A(b'_1)=q_A(b'_2)\leq b_1\wedge b_2$,
contradicting $b_1\wedge b_2=0$.\\
Preservation of complement follows from this.\\
  Using properness of the quotient, it is
  then easily shown that the relation algebra part of the lax
  homomorphism property carries over from $\varphi$ to $\varphisim$:
  Concerning
  composition, by surjectivity of $q_A$, we know that any given relations
  $R_1,R_2\in \Asim$ are of the form $R_1=q_A(S_1)$ and $R_2=q_A(S_2)$.
  Hence, $\varphisim(R_1;R_2)=\varphisim(q_A(S_1);q_A(S_2))
  =\varphisim(q_A(S_1;S_2))=\mathcal{P}(q \times q)(\varphi(S_1;S_2)) \geq
  \mathcal{P}(q \times q)(\varphi(S_1);\varphi(S_2)) = \mathcal{P}(q \times
  q)(\varphi(S_1));\mathcal{P}(q \times q)(\varphi(S_2)) =
  \varphisim(q_A(S_1));\varphisim(q_A(S_2))=\varphisim(R_1);\varphisim(R_2)$.  The inequality of the
  identity is shown similarly.
\end{pf}

\begin{prop}
  $(q_A,q):\varphi\to \varphisim$ is an oplax quotient homomorphism of
  weak representations. If $\sim$ is regular w.r.t.\ $\varphi$, then
  the quotient homomorphism is proper, and satisfies the following
  universal property: if $(q_B,i):\varphi\to \psi$ is another oplax
  homomorphism of weak representations with $\psi$ injective and
  $\sim\subseteq \mathit{ker}(i)$, then there is a unique oplax
  homomorphism of weak representations $(h,k):\varphisim\to\psi$ with
  $(q_B,i)=(h,k)\circ(q_A,q)$.
\end{prop}
\begin{pf}
 The oplax homomorphism property for $(q_A,q)$ is
 $\mathcal{P}(q \times q)\circ\varphi\subseteq \varphisim\circ q_A$,
 which by definition of $\varphisim$ amounts to
 $$\mathcal{P}(q \times q)\circ\varphi\subseteq \mathcal{P}(q \times q)\circ\varphi\circ q_A^{-1}\circ q_A,$$
which follows from surjectivity of $q$.
 Regularity of $\sim$ is  w.r.t.\ $\varphi$ means that
 $\sim_A$ is the kernel of $\mathcal{P}(q \times q)\circ\varphi$,
 which turns the above inequation into an equality.
 Concerning the universal property, let $(q_B,i):\varphi\to \psi$
 with the mentioned properties be given. Since $\sim\subseteq \mathit{ker}(i)$,
 there is a unique function $k:\Dsim\to\V$ with $i=k\circ q$.
 The homomorphism $h$ we are looking for is determined uniquely by
 $h(q_A(b))=q_B(b)$; this also ensures the proper homomorphism property.
 All that remains to be shown is well-definedness.
 Suppose that $b_1\sim_A b_2$. By regularity, $\mathcal{P}(q \times q)(\varphi(b_1))=\mathcal{P}(q \times q)(\varphi(b_2))$. Hence also
 $\mathcal{P}(i \times i)(\varphi(b_1))=\mathcal{P}(i \times i)(\varphi(b_2))$
 and $\psi(q_B(b_1))=\psi(q_B(b_2))$.  By injectivity of $\psi$, we get
 $q_B(b_1)=q_B(b_2)$.
\end{pf}

\begin{exmp}\label{exa:DRAfpDRAopp}
Given dipoles $d_1,d_2\in\dipole$, let $d_1\sim d_2$ denote
that $d_1$ and $d_2$ have the same start point and point in the
same direction. (This is regular w.r.t. $\rf$.)
 Then $\dipole\!/\!\!\sim$ is the domain $\opra$
of oriented points in $\R^2$.
Let $\rop:\DRAop \to  {\cal P}(\opra\times\opra)$ and
$\ropp:\DRAopp \to {\cal P}(\opra\times\opra)$ be the weak
representations obtained as quotients of $\rf$ and $\rfp$, respectively,
see Fig.~\ref{fig:quotient}.
At the level of non-associative algebras, the quotient is given
by the tables in Figs.~\ref{fig:DRAop} and~\ref{fig:DRAopp}.
\end{exmp}
This way of constructing $\DRAop$ and $\DRAopp$ by a quotient
gives us their converse and composition tables for no extra effort;
we can obtain them by applying the respective congruences
to the tables for $\DRAf$ and $\DRAfp$, respectively. Moreover,
the next result shows that we also can use the quotient
to transfer an important property of calculi.

\begin{figure}
$$\xymatrix{
\DRAfp \ar[rr]^{\rfp} \ar[dd] &&
\mathcal{P}(\dipole\times\dipole) \ar[dd]\\
&&\\
\DRAopp \ar[rr]^{\ropp}&&
\mathcal{P}(\opra\times\opra)
}$$
\caption{Homomorphisms of weak representations from $\DRAfp$
to $\DRAopp$\label{fig:quotient}}
\end{figure}

\begin{prop}\label{prop:preserveStrength}
Quotient homomorphism of weak representations preserve strength
of composition.
\end{prop}
\begin{pf}
Let $(h,\hD):\varphi \to \psi$ with
$\varphi:A\to\mathcal{P}(\D\times\D)$ and
$\psi:B\to\mathcal{P}(\V\times\V)$
be a quotient homomorphism of weak representations.
According to Prop.~\ref{fact:strong-composition},
the strength of the composition is equivalent to $\varphi$ (respectively\ $\psi$) being
a proper homomorphism.
We assume that $\varphi$ is a proper homomorphism and need to show
that $\psi$ is proper as well.
We also know that $h$ and $\mathcal{P}(\hD\times \hD)$ are proper.
Let $R_2,S_2$ be two abstract relations in $B$. Because of the surjectivity of $h$,
there are abstract relations $R_1,S_1\in A$ with $h(R_1)=R_2$
and $h(S_1)=S_2$. Now $\psi(R_2;S_2)=\psi(h(R_1);h(S_1))=
\psi(h(R_1;S_1))=\mathcal{P}(\hD\times \hD)(\varphi(R_1;S_1))=
\mathcal{P}(\hD\times \hD)(\varphi(R_1));\mathcal{P}(\hD\times \hD)(\varphi(S_1))=
\psi(h(R_1));\psi(h(S_1))=\psi(R_2);\psi(S_2)$, hence $\psi$ is proper.
\end{pf}

The application of this Proposition must wait until
Section~\ref{condensedsemantics}, where we develop the necessary
machinery to investigate the strength of the calculi.
The domains of $\DRAop$ and $\OPRA_1$ obviously coincide.
An inspection of the converse and composition tables (that
of $\OPRA_1$ is given in \cite{Frommberger2007}) shows:

\begin{prop}\label{prop:DRAop-OPRA-1}
$\DRAop$ is isomorphic to $\OPRA_1$.
\end{prop}

We can also obtain a similar statement for $\DRAopp$.
The calculus $\OPRA^{*}_1$ \cite{dylladis} is
a refinement of  $\OPRA_1$ that is obtained along the same
features as $\DRAfp$ is obtained from $\DRAf$.
The method how to compute the composition table for
$\OPRA^{*}_1$ is described in \cite{dylladis} and a reference composition
table is provided with the tool \texttt{SparQ} \cite{SparQmanual}.

\begin{prop}\label{prop:DRAopp-OPRASTAR-1}
$\DRAopp$ is isomorphic to $\mathcal{OPRA}^*_1$.
\end{prop}

In the course of checking the isomorphism properties between $\DRAopp$ and
$\mathcal{OPRA}^*_1$, we discovered errors in $197$ entries of the composition table of
$\mathcal{OPRA}^*_1$ as it was shipped with the qualitative reasoner
SparQ \cite{SparQmanual}. This emphasizes our point how important it is to
develop a sound mathematical theory to compute a composition table and to
stay as close as possible with the implementation to the theory. In the composition
table for $\mathcal{OPRA}^*_1$ it was claimed that
\begin{eqnarray*}
\relnsp{SAMEright}; \relnsp{RIGHTrightA} &\Longrightarrow&
\{
\relnsp{LEFTright+}, \relnsp{LEFTrightP}, \relnsp{LEFTright-},\\ && \quad
\relnsp{BACKright},  \relnsp{RIGHTright+},\\ && \quad
\relnsp{RIGHTrightA},\relnsp{RIGHTright-}
\}
\end{eqnarray*}
were we use the $\DRAopp$ notation for the $\mathcal{OPRA}^*_1$-relations for
convenience. So the abstract composition $\relnsp{SAMEright}; \relnsp{RIGHTrightA}$
contains the base relation
$\relnsp{LEFTrightP}$, which however is not supported geometrically. Consider three oriented points $o_A$, $o_B$ and
$o_C$ with $o_A \rel{SAMEright} o_B$
\begin{figure}[htb]
\begin{center}
\includegraphics[keepaspectratio, scale=0.85]{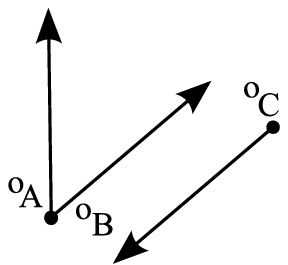}
\caption{\label{leftfig:OpraStar} $\mathcal{OPRA}^*_1$ configuration}
\end{center}
\end{figure}
and $o_B \rel{RIGHTrightA} o_C$, as depicted in Fig.~\ref{leftfig:OpraStar}. For the relation
$o_A \rel{LEFTrightP} o_C$ to hold, the carrier rays of $o_A$ and $o_C$ need to be parallel, but
because of $o_B \rel{RIGHTrightA} o_C$, the carrier rays of $o_B$ and $o_C$
and hence also those of  $o_A$ and $o_B$ need to
be parallel as well. Since the start point of $o_A$ and $o_B$ coincide, this can only be
achieved, if $o_A$ and $o_B$ are collinear, which is a contradiction to
$o_A \rel{SAMEright} o_B$.

\medskip

Altogether, we get the following diagram of calculi (weak representations)
and homomorphisms among them:
$$\xymatrix{
\mathcal{IA}\ar[rr]^{\textrm{\small proper}} &&  \DRAfp \ar[rr]^{\textrm{\small oplax}} \ar[dd]_{\textrm{\small oplax quotient}} && \DRAf \ar[dd]^{\textrm{\small oplax quotient}} && \mathcal{IA}\ar[ll]_{\textrm{\small proper}} \\
&&\\
\OPRA_1^* &\cong& \DRAopp \ar[rr]^{\textrm{\small oplax}} && \DRAop &\cong& \OPRA_1
}$$

\subsection{Constraint Reasoning}

Let us now apply the relation-algebraic method to constraint reasoning.
Dipole constraints are written as $xRy$, where $x,y$ are variables for the
dipoles and $R$ is a $\DRAf$ or $\DRAfp$ relation.
Given a set $\Theta$ of dipole constraints, an important reasoning
problem is to decide whether $\Theta$ is {\em consistent}, i.e.,
whether there is an assignment of all variables of $\Theta$ with
dipoles such that all constraints are satisfied (a {\em solution}).
We call this problem {\sf DSAT}. {\sf DSAT} is a Constraint Satisfaction
Problem (CSP) \cite{mackworth77}. We
rely on relation algebraic methods to check consistency,
namely the above mentioned path consistency algorithm.
For non-associative algebras, the abstract composition of relations
need not coincide with the (associative) set-theoretic composition.
Hence, in this case, the standard path-consistency algorithm does
not necessarily lead to path consistent networks, but only to algebraic closure \cite{RenzLigozat}:

\begin{defn}[Algebraic Closure]
    A CSP over binary relations is called \emph{algebraically closed} if for all variables
    $X_1, X_2, X_3$ and all relations $R_1, R_2, R_3$ the constraint relations
	$$ R_1(X_1,X_2),\quad R_2(X_2, X_3),\quad R_3(X_1, X_3) $$
    imply
    \[
      R_3 \le R_1; R_2.
    \]
\end{defn}

In general, algebraic closure is therefore only a one-sided approximation
of consistency: if algebraic closure detects an inconsistency, then we are
sure that the constraint network is inconsistent; however, algebraic
closure may fail to detect some inconsistencies: an algebraically
closed network is not necessarily consistent.  For some calculi, like
Allen's interval algebra, algebraic closure is known to exactly decide
consistency, for others it does not, see \cite{RenzLigozat}, where it
is also shown that this question is completely orthogonal to the
question as to whether the composition is strong.  We will examine these
questions for the dipole calculi in Section~\ref{condensedsemantics}
below.

Fortunately, it turns out that oplax homomorphisms preserve algebraic closure.

\begin{prop}\label{prop:hom-alg-closure}
Given non-associative algebras $A$ and $B$, an oplax homomorphism
$\mathop{\mathrm{h}}: A \longrightarrow B$ preserves algebraic closure.
If $\mathop{\mathrm{h}}$ is injective, it also reflects algebraic closure.
\end{prop}
\begin{pf}
  Since an oplax homomorphism is a homomorphism between Boolean algebras,
  it preserves the order. So for any
  three relations $R_1, R_2, R_3$ in the algebraically closed CSP
  over $A$, with
  \[
  R_3 \le R_1; R_2
  \]
  the preservation of the order implies:
  \[
  \mathop{\mathrm{h}}(R_3) \le \mathop{\mathrm{h}}(R_1; R_2).
  \]
  Applying the oplaxness property yields:
  \[
  \mathop{\mathrm{h}}(R_3) \le \mathop{\mathrm{h}}(R_1);
  \mathop{\mathrm{h}}(R_2).
  \]
  and hence the image of the CSP under $\mathop{\mathrm{h}}$ is also
  algebraically closed. If $\mathop{\mathrm{h}}$ is injective,
  it reflects equations and inequations, and the converse implication follows.
\end{pf}

\begin{defn}
  Following \cite{RenzLigozat}, a \emph{constraint network} over a
  non-associative algebra $A$ can be seen as a function $\nu:A\to
  {\cal P}(N\times N)$, where $N$ is the set of nodes (or variables), and
  $\nu$ maps each abstract relation $R$ to the set of pairs
  $(n_1,n_2)$ that are decorated with $R$.  (Note that $\nu$ is a weak
  representation only if the constraint network is algebraically
  closed.)

  Constraint networks can be translated along homomorphisms of
  non-associative algebras as follows: Given $h:A\to B$ and $\nu:A\to
  {\cal P}(N\times N)$, $h(\nu):B\to {\cal P}(N\times N)$ is the
  network that decorates $(n_1,n_2)$ with $h(R)$ whenever $\nu$
  decorates it with $R$

  A \emph{solution} for $\nu$ in a weak representation
  $\varphi:A\to\mathcal{P}(\D\times\D)$ is a function $j:N\to \D$ such
  that for all $R\in A$, ${\cal P}(j\times j)(\nu(R))\subseteq
  \varphi(R)$, or ${\cal P}(j\times j)\circ\nu\subseteq \varphi$ for
  short.

\end{defn}

\begin{prop}\label{prop:hom-solutions}
Oplax homomorphisms of weak representations preserve
solutions for constraint networks.
\end{prop}
\begin{pf}
Let weak representations $\varphi:A\to\mathcal{P}(\D\times\D)$
and $\psi:B\to\mathcal{P}(\V\times\V)$ and an
oplax homomorphism of weak representations
$(h,\hD):\varphi \to \psi$ be given.

A given solution $j:N\to \D$ for $\nu$ in $\varphi$ is defined by
${\cal P}(j\times j)\circ\nu\subseteq \varphi$. From this and the
oplax commutation property ${\cal P}(i\times i)\circ\varphi\subseteq
\psi\circ h$ we infer ${\cal P}(i\circ j\times i\circ
j)\circ\nu\subseteq \psi\circ h$, which implies that $i\circ j$ is a
solution for $h(\nu)$.
\end{pf}

An important question for a calculus (= weak representation)
is whether algebraic closure decides consistency.
We will now prove that this property is preserved under
certain homomorphisms.

\begin{prop}\label{prop:hom-alg-closure-consistency}
Oplax homomorphisms $(h,\hD)$ of weak representations with $h$
injective preserve the property that algebraic closure decides
consistency to the image of $h$.
\end{prop}
\begin{pf}
Let weak representations $\varphi:A\to\mathcal{P}(\D\times\D)$
and $\psi:B\to\mathcal{P}(\V\times\V)$ and an
oplax homomorphism of weak representations
$(h,\hD):\varphi \to \psi$ be given. Further assume that
for $\varphi$, algebraic closure decides
consistency.

Any constraint network in the image of $h$ can be written as
$h(\nu):B\to {\cal P}(N\times N)$. If $h(\nu)$ is algebraically
closed, by Prop.~\ref{prop:hom-alg-closure}, this carries over to
$\nu$.  Hence, by the assumption, $\nu$ is consistent, i.e.\ has a
solution. By Prop.~\ref{prop:hom-solutions}, $h(\nu)$ is consistent as
well.  Note that the converse directly always holds: any consistent
network is algebraically closed.
\end{pf}

For calculi such as RCC8, interval algebra etc., (maximal)
\emph{tractable subsets} have been determined, i.e.\ sets of
relations for which algebraic closure decides consistency.
We can apply Prop.~\ref{prop:hom-alg-closure-consistency} to the
homomorphism from interval algebra to $\DRAf$ (see
Example~\ref{ex:IA-DRAf}). We obtain that algebraic closure
in $\DRAf$ decides consistency of any constraint network
involving (the image of) a maximal tractable subset of
the interval algebra only.

On the other hand, the consistency problem for the $\DRAc$ calculus in
the base relations is already NP-hard, see \cite{LeeWolterAIJ}, and
hence algebraic closure does not decide consistency in this case.  We
will resume the discussion of consistency versus algebraic closure in
Sect.~\ref{reasoning}.

\section{A Condensed Semantics for the Dipole Calculus}
\label{condensedsemantics}

The $72$ base relations of $\DRAf$, or the 80 base relations of $\DRAfp$,
have so far been derived manually. This is a potentially erroneous procedure\footnote{For this reason, the
manually derived sets of base relations for the finer-grained dipole calculi described in
\cite{Schlieder95,Moratz00_QSRwithLineSegs} contained errors.}, especially if the calculus has
many base-relations like the $\DRAf$ and $\DRAfp$
calculi.
Therefore, it is necessary to use methods which yield more reliable results.
To start, we tried verifying the composition table of $\DRAf$ directly,
using the resulting quadratic inequalities as given in \cite{Moratz00_QSRwithLineSegs}.
However, it turned out that it is unfeasible to base the reasoning on these inequalities, even with
the aid of interactive theorem provers such as Isabelle/HOL \cite{NipPauWen02} and HOL-light \cite{DBLP:conf/tphol/Harrison09a}
(the latter is dedicated to proving facts about real numbers). This unfeasibility
is probably related to the above-mentioned NP-hardness of the consistency problem for $\DRAf$
base relations.
So, we developed a qualitative abstraction instead. A key
insight is that two configurations are qualitatively different if they
cannot be transformed into each other by maps that keep that part
of the spatial structure invariant that is essential for the calculus. In our case,
these maps are (orientation-preserving) affine bijections.
A set of configurations that can be transformed into each other
by appropriate maps is an \emph{orbit} of a suitable automorphism group.
Here, we use primarily the affine group $\mathbf{GA}(\R^2)$ and detail
how this leads to qualitatively different spatial configurations.

\subsection{Seven qualitatively different configurations}
\label{sec:seven}
Since the domains of most spatial calculi are infinite (e.g. the
Euclidean plane), it is impossible just to enumerate all possible configurations
relative to the composition operation when deriving a composition table. It is
still possible to enumerate a well-chosen subset of all configurations to
obtain a composition table, but it is difficult to show that this subset leads to
a complete table. We have experimented with the enumeration of all
$\DRAf$ scenarios with six points (which are the start- and end-points of three dipoles), 
which are equivalent to the entries
of the composition table, in a \emph{finite} grid over natural numbers. This
method led to a usable composition table, but its computation took
several weeks and it is unclear if it is complete. The goal remains the
efficient and automatic computation of a composition table. To obtain an efficient
method for computing the table, we introduce the \emph{condensed semantics} for
$\DRAf$ and $\DRAfp$. For these, we observe the
Euclidean plane with respect to all possible line configurations that are
distinguishable within the $\mathcal{DRA}$ calculi. With condensed
semantics, there is already a level of abstraction from the metrics of the underlying space. All we
can see are lines that are parallel or intersect. For the
binary composition operation of $\mathcal{DRA}$ calculi, we have to consider
all qualitatively different configurations of three lines.

In order to formalize ``qualitatively different configurations'',
we regard the $\mathcal{DRA}$ calculus as a first-order structure,
with the Euclidean plane as its domain, together with all the base relations.

\begin{prop}\label{thm:automorphisms}
All orientation-preserving affine bijections are $\DRAf$ and $\DRAfp$
automorphisms.
\end{prop}
(In \cite{MossakowskiWoelfl09}, the converse is also shown.)
\begin{pf}
It suffices to show that orientation preserving affine bijections
preserve the $\LR$ relations. Now, any orientation-preserving affine bijection
can be composed of translations, rotations, scalings and shears.
It is straightforward to see that these mappings preserve the $\LR$ relations.
\end{pf}

Recall that an affine map $f$ from Euclidean space to itself is given by
$$f(x,y) = A{{x}\choose{y}} + (b_x,b_y)$$
$f$ is a bijection iff $det(A)$ is non-zero.

Automorphisms and their compositions form a group which acts on
the set of points (and tuples of points, lines, etc.) by function application.
Recall that, if a group $G$ acts on a set, an \emph{orbit}
consists of the set reachable from a fixed element by performing the
action of all group elements: $O(x) = \{f(x) | f\in G\}$. The importance of this notion
is the following:
\begin{quote}
Qualitatively different configurations are orbits of the
automorphism group.
\end{quote}

Here, we start with configurations consisting of three lines,
i.e. we consider the orbits for all sets $\{l_1,l_2,l_3\}$ of (at most) three lines\footnote{We do
not require that $l_1$, $l_2$ and $l_3$ are distinct; hence, the set
$\{l_1,l_2,l_3\}$ may also consist of two elements or be a singleton.} in Euclidean space
with respect to the group of \emph{all} affine bijections (and not just the
orientation preserving ones -- orientations will come in at a later stage).
This group is usually called the affine group of $\R^2$ and denoted
by $\mathbf{GA}(\R^2)$.

A line in Euclidean space is given by the set of all points $(x,y)$ for which
$y=mx+b$. Given three lines $y=m_i x+b_i$ ($i=1,2,3$),
we list their orbits by giving a defining property.
In each case, it is fairly obvious that the defining property is
preserved by affine bijections. Moreover, in each case, we show
a \emph{transformation property}, namely that given two instances of the defining properties,
the first can be transformed into the second by an affine bijection.
Together, this means that the defining property exactly specifies an orbit.
The transformation property often follows from the following basic facts
about affine bijections, see \cite{Gallier2000}:
\begin{enumerate}
\item
An affine bijection is uniquely determined by its action on an affine
basis, the result of which is given by another affine basis.
Since an affine basis of the Euclidean plane is a
point triple in general position,
given any two point triples in general position,
there is a unique affine bijection mapping the first point triple to
the second.
\item
Affine maps transform lines into lines.
\item
Affine maps preserve parallelism of lines.
\end{enumerate}
That is, it suffices to show that an instance of the defining property is
determined by three points in general position and drawing lines and parallel lines.

We will consider the intersection of line $i$ with line $j$ ($i\not = j\in\{1,2,3\}$).
This is given by the system of equations:
$$\{y=m_i x+b_i,~ y=m_j x+b_j\}.$$
For $m_i\not = m_j$, this has a unique solution:
$$x=-\frac{b_i-b_j}{m_i-m_j}, ~ y=\frac{m_ib_j-m_jb_i}{m_i-m_j}.$$
For $m_i = m_j$, there is either is no solution ($b_i \not = b_j$; the lines are parallel),
or there are infinitely many solutions ($b_i=b_j$; the lines are identical).

We can now distinguish seven cases:

\begin{enumerate}
\item All $m_i$ are distinct and
the three systems of equations $\{y=m_i x+b_i,~ y=m_j x+b_j\}$
($i\not = j\in\{1,2,3\}$) yield three different solutions. Geometrically, this means that
  all three lines intersect with three different intersection points.
  The transformation property follows from the fact that the three intersection points
   determine the configuration.
  \bigskip \begin{center}
    \includegraphics[keepaspectratio, scale=0.85]{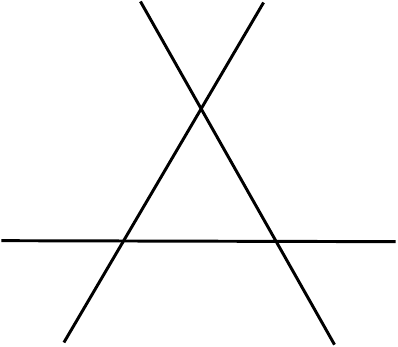}
  \end{center}  \bigskip
\item  All $m_i$ are distinct and
at least two of the three systems of equations $\{y=m_i x+b_i,~ y=m_j x+b_j\}$
($i\not = j\in\{1,2,3\}$) have a common solution. Then, obviously, the
single solution is common to all three equation systems. Geometrically, this means that
all three lines intersect at the same point.
  \bigskip \begin{center}
    \includegraphics[keepaspectratio, scale=0.85]{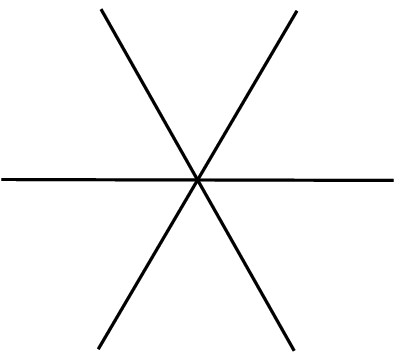}
  \end{center}  \bigskip
Take this point and a second
point on one of the lines. By drawing parallels through this second point,
we obtain two more points, one on each of the other two lines, such that
the four points form a parallelogram. The transformation property now
follows from the fact that any two non-degenerate parallelograms can be
transformed into each other by an affine bijection.
  \bigskip \begin{center}
    \includegraphics[keepaspectratio, scale=0.85]{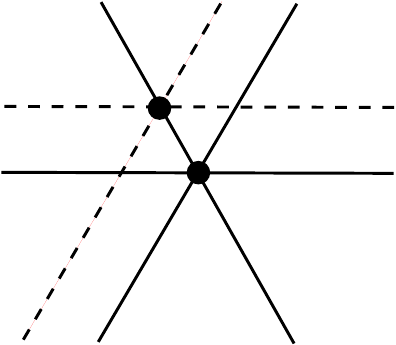}
  \end{center}  \bigskip

\item $m_i=m_j\not = m_k$ and $b_i\not = b_j$ for distinct $i,j,k\in\{1,2,3\}$.
Geometrically, this means that two lines are parallel, but not coincident, and the third line intersects them.
Such a configuration is determined by three points: the points of intersection, plus a further point on
one of the parallel lines. Hence, the transformation property follows.
  \bigskip \begin{center}
    \includegraphics[keepaspectratio, scale=0.85]{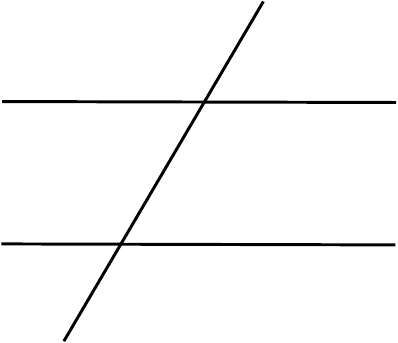}
  \end{center}  \bigskip
\item $m_i=m_j\not = m_k$  and $b_i= b_j$ for distinct $i,j,k\in\{1,2,3\}$.
Geometrically, this means that two lines are equal and a third one intersects them.
Again, such a configuration is determined by three points: the intersection point plus a further point on each
of the (two) different lines. Hence, the transformation property follows.
  \bigskip \begin{center}
    \includegraphics[keepaspectratio, scale=0.85]{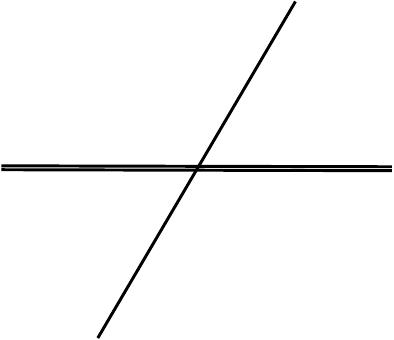}
  \end{center}  \bigskip

\item All $m_i$ are equal, but the $b_i$ are distinct.
Geometrically, this means that all three lines are parallel, but not coincident.
We cannot show the transformation property here, which means that this case comprises
several orbits. Actually, we get one orbit for each distance ratio
$$\frac{b_1-b_2}{b_1-b_3}.$$
An affine bijection $$f(x,y) = A{{x}\choose{y}} + (b_x,b_y)$$
transforms
a line $y=mx+b$ to $y=m'x+b'$, with $b'=c_1(m)b+c_2(m)$, where $c_1$ and $c_2$ depend non-linearly on $m$.
However, since $m=m_1=m_2=m_3$, this non-linearity does not matter.
This means that
$$\frac{b'_1-b'_2}{b'_1-b'_3}=\frac{c_1(m)b_1-c_1(m)b_2}{c_1(m)b_1-c_1(m)b_3}=\frac{b_1-b_2}{b_1-b_3},$$
i.e. the distance ratio is invariant
under affine bijections (which is well-known in affine geometry).
Given a fixed distance ratio, we can show the transformation property:
three points suffice to determine two parallel lines, and the position of the
third parallel line is then determined by the distance ratio.
For a distance ratio $1$, this configuration looks as follows:
 \bigskip \begin{center}
    \includegraphics[keepaspectratio, scale=0.85]{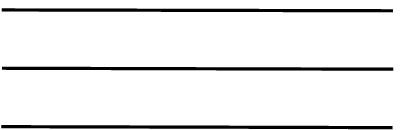}
  \end{center}  \bigskip
Actually, for the qualitative relations between dipoles placed on parallel lines,
their distance ratio does not matter. Hence, we will ignore distance ratios when computing
the composition table below. The fact that we get infinitely many orbits for this sub-case will
be discussed below.

\item All $m_i$ are equal and two of the $b_i$ are equal but different from the third.
Geometrically, this means that two lines are coincident, and a third one is parallel but not coincident.
Such a configuration is determined by three points: two points on the coincident lines and a third point on
the third line. Hence, the transformation property follows.
  \bigskip \begin{center}
    \includegraphics[keepaspectratio, scale=0.85]{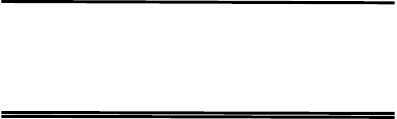}
  \end{center}  \bigskip
\item All $m_i$ are equal, and the $b_i$ are equal as well.
This means that all three lines are equal. The transformation property is obvious.
  \bigskip \begin{center}
    \includegraphics[keepaspectratio, scale=0.85]{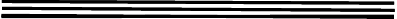}
  \end{center}  \bigskip
\end{enumerate}
Since we have exhaustively distinguished the various possible cases based on relations between
the $m_i$ and $b_i$ parameters,
this describes all possible orbits of three lines w.r.t.\ affine bijections.
Although we get infinitely many orbits for case (5), in contexts where the
distance ratio introduced in case (5) does not matter, we will speak
of seven qualitatively different configurations, and it is understood that
the infinitely many orbits for case (5) are conceptually combined into one
equivalence class of configurations.

\begin{figure}[htb]
  \begin{center}
    \includegraphics[keepaspectratio, scale=0.85]{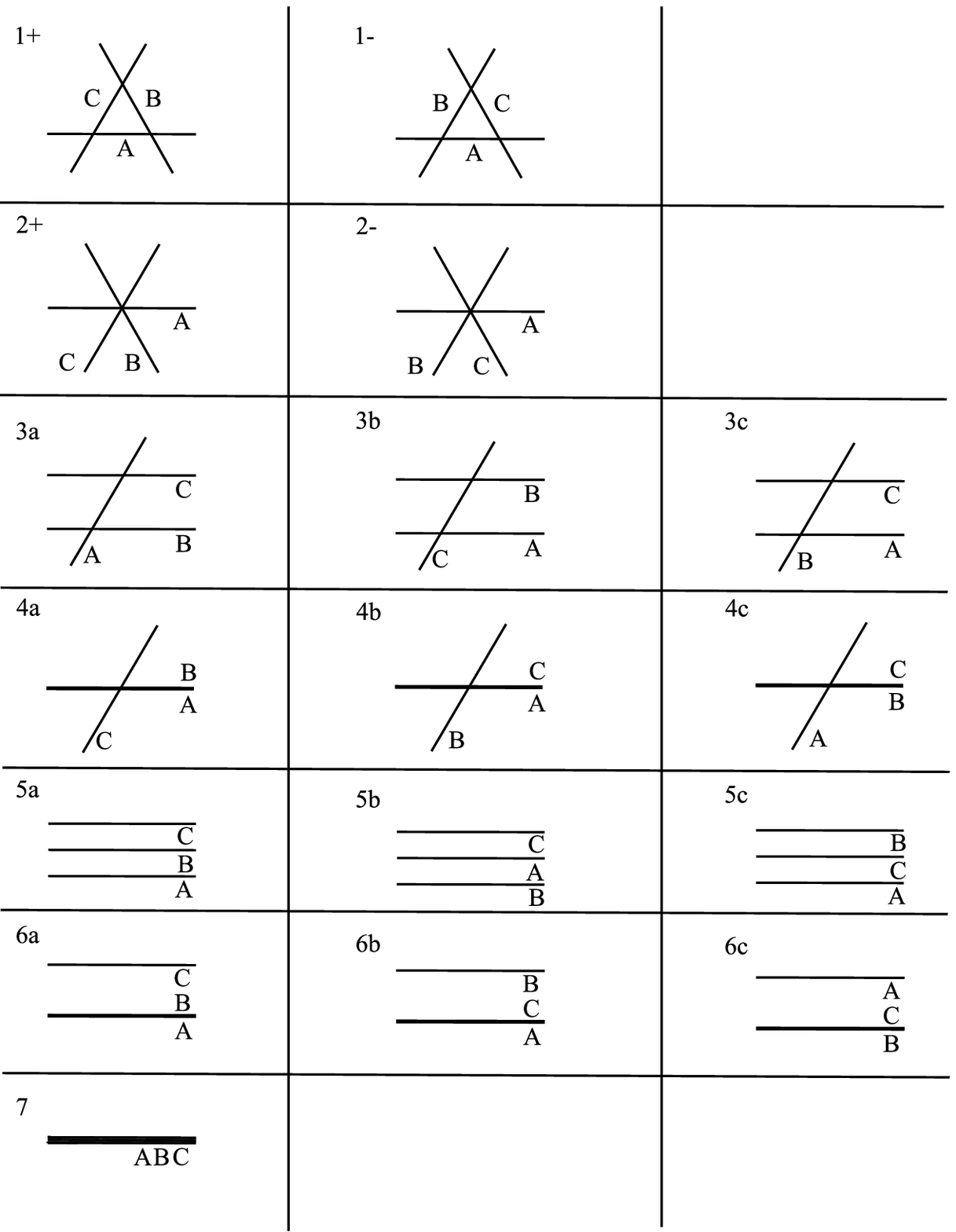}
    \caption{\label{fig:orientedOrbits}The $17$ qualitatively different configurations
      of triples of oriented lines w.r.t.\ orientation-preserving affine bijections}
  \end{center}
\end{figure}

Recall that we have considered \emph{sets} of (up to) three lines.  If
we consider \emph{triples} of lines instead, cases (3) to (6) split up
into three sub-cases, because they feature distinguishable lines.  We
then get 15 different configurations, which we name 1, 2, 3a, 3b, 3c,
4a, 4b, 4c, 5a, 5b, 5c, 6a, 6b, 6c and 7.  While 5a, 5b and 5c correspond to
case (5) above and therefore are comprised of infinitely many orbits,
the remaining configurations are comprised of a single orbit.

The next split appears at the point when we consider qualitatively 
different configurations of triples
of unoriented lines with respect to \emph{orientation-preserving}
affine bijections.
An affine map $f(x,y) = A{{x}\choose{y}} + (b_x,b_y)$
is orientation-preserving if $det(A)$ is positive.
In the above arguments, we now have to consider
oriented affine bases. Let us call an affine base $(p_1,p_2,p_3)$
positively ($+$) oriented, if the angle
$\angle(\vect{p_1}{p_2},\vect{p_1}{p_3})$ is positive, otherwise, it
is negatively ($-$) oriented.  Two given affine bases with the same
orientation determine a unique orientation-preserving affine
bijection transforming the first one into the second.  Thus, the
orientation of the affine base matters, and hence cases 1 and 2 above are split
into two sub-cases each.  For all the other cases, we have the freedom
to choose the affine bases such that their orientations coincide.
In the end, we get $17$ different orbits of triples of oriented lines:
1+, 1-, 2+, 2-, 3a, 3b, 3c, 4a, 4b, 4c, 5a, 5b, 5c, 6a, 6b, 6c and 7.
They are shown in Fig.~\ref{fig:orientedOrbits}
\FloatBarrier

\begin{figure}[htb]
  \begin{center}
  \includegraphics[keepaspectratio, scale=0.85]{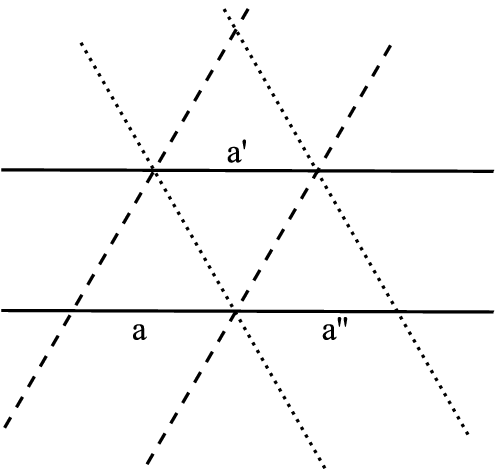}
    \caption{\label{fig:para}Replication of a given distance in $\DRAfp$}
  \end{center}
\end{figure}

The structure of the orbits already gives us some insight into the nature
of the dipole calculus. The fact that sub-case (1) corresponds to one orbit
means that neither angles nor ratios of angles can be measured in
the dipole calculus. By way of contrast, the presence of infinitely many orbits
in sub-case (5) means that ratios of distances
in a specific direction, not distances, \emph{can} be measured in the dipole calculus.
Indeed, in $\DRAfp$, it is even possible to replicate
a given distance arbitrarily many times, as indicated in Fig.~\ref{fig:para}.

That is, $\DRAfp$ can be used to generate a
one-dimensional coordinate system. Note however that, due to the
lack of well-defined angles, a two-dimensional coordinate system
cannot be constructed.

Note that Cristani's 2DSLA calculus \cite{Cristani03}, which can be
used to reason about sets of lines, is too coarse for our purposes:
cases (1) and (2) above cannot be distinguished in 2DSLA.

\subsection{Computing the composition table with Condensed Semantics}

For the composition of (oriented) dipoles, we use the seventeen
different configurations for triples of (unoriented) lines for the
automorphism group of orientation-preserving affine bijections that
have been identified in the previous section
(Fig.~\ref{fig:orientedOrbits}).  A \emph{qualitative composition
 configuration} consists of a qualitative configuration for a triple
of lines (the lines will serve as carrier lines for dipoles), carrying
qualitative location information for the start and end points of
three dipoles, as detailed in the sequel.  While the notion of
qualitative configuration composition is motivated by geometric
notions, it is purely abstract and symbolic and does not refer
explicitly to geometric objects.  This ensures that it can be directly
represented in a finite data structure.

Each of the three (abstract) lines $l^a_A, l^a_B, l^a_C$ of a qualitative
composition configuration carries two abstract segmentation
points $S_X$ and $E_X$ ($X\in\{A,B,C\}$). $\mathbf{P} = \left\{S_A,
  S_B, S_C, E_A, E_B, E_C\right\}$ is the set of all abstract
segmentation points.

\begin{figure}[h!]
  \begin{center}
    \includegraphics[keepaspectratio, scale=0.85]{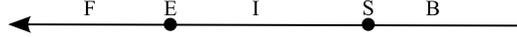}
    \caption{\label{fig:linesegmentation}Segmentation on the line.}
  \end{center}
\end{figure}

In the geometric interpretation of these abstract entities (which will be
defined precisely later on), the segmentation points
lead to a segmentation of the lines. So, we introduce
five abstract segments $F$, $E$, $I$, $S$, $B$ (the letters are
borrowed from the $\mathcal{LR}$ calculus). The set of abstract
segments is denoted by $\mathcal{S}$. It is ordered in the following sequence:
\[
F > E > I > S > B.
\]
The geometric intuition behind this is shown in Fig.~\ref{fig:linesegmentation}.

Having this segmentation of line configurations, we can introduce
qualitative configurations for \emph{abstract dipoles} by
qualitatively locating their start and end points based on the above
segmentation.  In the case that two or more points fall onto the same
segment, information on the relative location of points within
that segment is needed; this is provided by an ordering relation
denoted by $<_p$.

By $\mathcal{D}$, we denote the set $\mathcal{S} \times \mathcal{S}
\setminus \{(S,S), (E,E)\}$ (the exclusion of $\{(S,S), (E,E)\}$
is motivated by the fact that the start and end points of a dipole
cannot coincide). By $st(dp)$ and $ed(dp)$, we denote the
projections to the first and second components of each tuple,
respectively. For convenience, we call the elements of the co-domains of $st$ and $ed$
abstract points.


Finally, we need information on the points of intersection of lines.
Depending on orbit, there may be none, one, two or three
points of intersection.
Hence, we introduce sets
$\hat \mathcal{S}(i)$ with $i \in \left\{1+, 1-, 2+, 2-, 3a, 3b, 3c, 4a, 4b, 4c, 5a, 5b, 5c, 6a, 6b, 6c, 7 \right\}$ which give names to each abstract point of intersection.
These sets are defined as:
\begin{eqnarray*}
\hat \mathcal{S}(1+) &:=& \left\{\hat s_{AB}, \hat s_{BC}, \hat s_{AC} \right\} \\
\hat \mathcal{S}(1-) &:=& \left\{\hat s_{AB}, \hat s_{BC}, \hat s_{AC} \right\} \\
\hat \mathcal{S}(2+) &:=& \left\{\hat s_{ABC} \right\} \\
\hat \mathcal{S}(2-) &:=& \left\{\hat s_{ABC} \right\} \\
\hat \mathcal{S}(3a) &:=& \left\{\hat s_{AB}, \hat s_{AC} \right\} \\
\hat \mathcal{S}(3b) &:=& \left\{\hat s_{AC}, \hat s_{BC} \right\} \\
\hat \mathcal{S}(3c) &:=& \left\{\hat s_{AB}, \hat s_{BC} \right\} \\
\hat \mathcal{S}(4a) &:=& \left\{\hat s_{ABC} \right\} \\
\hat \mathcal{S}(4b) &:=& \left\{\hat s_{ABC} \right\} \\
\hat \mathcal{S}(4c) &:=& \left\{\hat s_{ABC} \right\} \\
\hat \mathcal{S}(5a) &:=& \emptyset \\
\hat \mathcal{S}(5b) &:=& \emptyset \\
\hat \mathcal{S}(5c) &:=& \emptyset \\
\hat \mathcal{S}(6a) &:=& \emptyset \\
\hat \mathcal{S}(6b) &:=& \emptyset \\
\hat \mathcal{S}(6c) &:=& \emptyset \\
\hat \mathcal{S}(7) &:=& \emptyset \\
\end{eqnarray*}
where $\hat s_{XY}$ denotes the point of intersection of abstract lines $l^a_X$ and $l^a_Y$ and $\hat s_{XYZ}$
denotes the the point of intersection of the three abstract lines $l^a_X$, $l^a_Y$ and $l^a_Z$.

In the geometric interpretation, we require segmentation points that
coincide with points of intersection whenever possible.  This
coincidence is expressed via an \emph{assignment mapping}, which is a
partial mapping $a: \mathbf{P} \rightharpoonup \hat\mathcal{S}(i)$
subject to the following properties:
\begin{itemize}
\item if $a (S_X) = \hat s_y$, then $y$ contains $X$;
\item if $a (E_X) = \hat s_y$, then $y$ contains $X$;
\item if both $a (S_X)$ and $a (E_X)$ are defined, then
  $a (S_x) \neq a (E_x)$, for all $X \in \left\{ A, B, C \right\}$;
\item the domain of $a$ has to be maximal.
\end{itemize}
The first two conditions express that each abstract segmentation point
is mapped to the correspondingly named abstract point of intersection.
The third condition requires that the abstract segmentation points
of a line cannot be mapped to the same abstract point of intersection.
The last condition ensures that abstract segmentation points
are mapped to abstract points of intersection whenever possible.

We now arrive at a formal definition:
\begin{defn}[Qualitative Composition Configuration]\label{def:qcc}
  A \emph{qualitative composition configuration} (qcc) consists of:
  \begin{itemize}
  \item \sloppypar An identifier $i$ from the set \\
    $\left\{1+, 1-, 2+, 2-, 3a, 3b, 3b, 4a, 4b, 4c, 5a, 5b, 5c, 6a, 6b, 6c,
      7 \right\}$
    denoting one of the qualitatively different configurations
    of line triples as introduced in Section~\ref{sec:seven};
  \item An assignment mapping $a: \mathbf{P} \rightharpoonup \hat\mathcal{S}(i)$;
  \item A triple $(dp_A, dp_B, dp_C)$ of elements from $\mathcal{D}$, where we call each such element an \emph{abstract dipole};
  \item A relation $<_{p}$ on all points, i.e. the start and end points of the abstract dipoles, which is compatible with $<$.
  \end{itemize}
\end{defn}

\begin{defn}[Abstract direction]
  For any abstract dipole $dp$, we say that $dir(dp) = +$ if and only if
  $ed(dp) >_{p} st(dp)$, otherwise $dir(dp) = -$.
\end{defn}


\subsubsection{Geometric Realization\label{sec:geomreal}}

In this section, we claim that each qcc has a realization, first of all, we
need to define what such a realization is.

\begin{defn}[Order on ray]
	Given a ray $l$, for two points $A$ and $B$, we say that
	$A <_r B$, if $B$ lies further in the positive direction than
	$A$.
\end{defn}

We construct a map on each ray that reflects the abstract segments shown
in Fig.~\ref{fig:linesegmentation} to provide a link between a qcc and a
compatible line scenario.

\begin{defn}[Segmentation map]
	Given a ray $r$ and two points $\tilde S$ and $\tilde E$ on it, the segmentation map
	$seg: r \longrightarrow \left\{\tilde F, \tilde E, \tilde I, \tilde S, \tilde B\right\}$ is defined as:
	\begin{eqnarray*}
		r(x) &=& \left\{
			\begin{array}{@{\quad}r@{\quad}l}
			\textnormal{if } \tilde S <_r \tilde E &
			\left\{
				\begin{array}{rl}
					\tilde F & \textnormal{if } \tilde E <_r x \\
					\tilde E & \textnormal{if } \tilde E =_r x \\
					\tilde I & \textnormal{if } \tilde x <_r \tilde E \wedge \tilde S <_r x\\
					\tilde S & \textnormal{if } \tilde S =_r x\\
					\tilde B & \textnormal{if } x <_r \tilde S\\
				\end{array}\right.
				\\
				\textnormal{if } \tilde E <_r \tilde S &
				\left\{
				\begin{array}{rl}
					\tilde F & \textnormal{if } x <_r \tilde E \\
					\tilde E & \textnormal{if } x =_r \tilde E \\
					\tilde I & \textnormal{if } \tilde E <_r x \wedge x <_r \tilde S \\
					\tilde S & \textnormal{if } x =_r \tilde S \\
					\tilde B & \textnormal{if } \tilde S <_r x \\
				\end{array}\right.
			\end{array}
		\right.
	\end{eqnarray*}
	for any point on $x$ on $r$.
\end{defn}

When it is clear that we are talking about segments on an actual ray, we often omit the $\tilde \_$.

\begin{defn}[Geometric Realization]\label{def:geometricreal}
	For any qcc $Q$ a \emph{geometric realization} $R(Q)$
	consists of a triple of dipoles $(d_A, d_B, d_C)$ in $\R^2$,
	three carrier rays $l_A$, $l_B$, $l_C$
	of the dipoles, and two points $\tilde S_X$ and $\tilde E_X$ on $l_X$ for each $X\in\{A,B,C\}$,
	such that:
	\begin{itemize}
		\item $(l_A, l_B, l_C)$ (more precisely, the corresponding triple of unoriented lines) belongs 
                  to the configuration denoted by the identifier $i$ of $Q$;
		\item the angle between $l_a$ and the other two rays must lie in the
                  interval $(\pi, 2 \cdot \pi]$;
		\item for any $x,y\in\tilde\mathbf{P}$, if $a(p(x))$
                      and  $a(p(y))$ are both defined and equal, then $x=y$
                      (where $p:\tilde \mathbf{P} =\{\tilde S_A,\tilde S_B,\tilde S_C,
                      \tilde E_A,\tilde E_B,\tilde E_C\}\to\mathbf{P}$ be the obvious bijection);
		\item for all $X$, $st(dp_X) = seg(st(d_X))$ and
			  $ed(dp_X) = seg(ed(d_X))$;
                \item for all points $x$ and $y$ on $l_X$, if $seg(x) < seg(y)$,
                  then $x <_r y$;
		\item if $l_X=l_Y$,
                  the order $<_p$
                  must be preserved for points
                  $st(d_X)$, $ed(d_X)$, $st(d_Y)$, $ed(d_Y)$, in such a way that:
                  if $st(dp_X) <_p st(dp_Y)$, then $st(d_Y) <_r st(d_X)$ and in
                  the same manner between all other points.
	\end{itemize}
	must hold.
\end{defn}

\begin{prop}\label{prop:recover}
Given three dipoles in $\R^2$, there is a qcc $Q$ and a geometric realization
of $R(Q)$ which uses these three dipoles.
\end{prop}

\begin{pf}
  For this proof, we construct a qcc from a scenario of
  three dipoles in $\R^2$.
  Given three dipoles $d_A$, $d_B$, $d_C$ in $\R^2$, we determine their
  carrier rays $l_A$, $l_B$, $l_C$ in such a way that the angles between
  $l_A$ and $l_B$ as well as $l_A$ and $l_C$ lie in the interval $(\pi, 2\cdot\pi]$.
  We determine the identifier of the configuration
  in which the the scenario lies. We determine the points of intersection of the
  rays and identify them with $\hat s_{XY}$ in $\hat\mathcal{S}(i)$.
  For all points $X$ in $\mathcal{P}$, for which $a$ is undefined, the points
  $\hat X$ are placed in such a way, that $S_X <_r E_X$ (which is equivalent to $S_X < E_X$).
  We identify $st(dp_X)$ and
  $ed(dp_X)$ according to the segmentation map on these rays. If
  two carrier rays coincide, we define the order $<_p$ w.r.t. $<_r$,
  otherwise it is arbitrary. This clearly gives a $qcc$.

  An example of this construction is given in Fig.~\ref{fig:raytrafo}.
  \begin{figure}[ht]
    \begin{center}
      \includegraphics[keepaspectratio, scale=0.85]{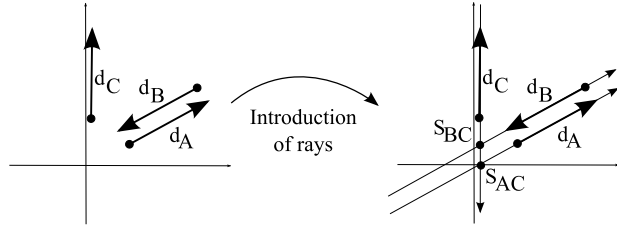}
      \caption{\label{fig:raytrafo}Construction of qcc}
    \end{center}
\end{figure}
  On the left-hand-side of Fig.~\ref{fig:raytrafo}, there is a scenario
  with three dipoles, lying somewhere in $\R^2$. On the right hand side, rays
  and points of intersection are added.
  Comparison with orbits and placement of lines determine the identifier
  $3b$ for this scenario. The map $a$ can be defined as
  \begin{eqnarray*}
    a(S_A) &=& \hat \mathcal{S}_{AC} \\
    a(S_B) &=& \hat \mathcal{S}_{BC} \\
    a(E_C) &=& \hat \mathcal{S}_{AC} \\
    a(S_B) &=& \hat \mathcal{S}_{BC}
  \end{eqnarray*}
  where the assignment is only free for $E_A$ and $E_B$. $E_A$ and $E_B$
  are lying at the start point of dipole $d_A$ and at the end point of dipole
  $d_B$. In this way, we get:
  \begin{eqnarray*}
    st(dp_A) = E \\
    ed(dp_A) = F \\
    st(dp_B) = E \\
    ed(dp_B) = I \\
    st(dp_C) = B \\
    ed(dp_C) = B
  \end{eqnarray*}
  and
  \begin{eqnarray*}
    dir(dp_A) = + \\
    dir(dp_B) = - \\
    dir(dp_C) = -
  \end{eqnarray*}
  In this case the assignment of $<_p$ is arbitrary. \\
  \noindent
  This construction gives us the desired qcc and a realization of it.
\end{pf}

\FloatBarrier

\subsection{Primitive Classifiers\label{sec:primclass}}

The last and most crucial point is the computation of $\mathcal{DRA}$
relations between three dipoles. We can decompose this task into subtasks,
since each $\DRAf$ relation comprises four $\mathcal{LR}$ relations
between a dipole and point; these are obtained from a qualitative composition configuration
using so-called \emph{primitive classifiers}. The \emph{basic classifiers} apply
the \emph{primitive classifiers} to the abstract dipoles in each qualitative composition configuration
in an adequate manner. For
$\DRAfp$ relations an extension of the \emph{basic classifiers}
is used in cases where the qualitative angle between several dipoles has to be determined.
Finally, the resulting data is collected in a (composition) table.

\begin{defn}[Primitive Qualitative Composition Configuration]
A \emph{primitive qualitative composition configuration} (pqcc) is a sub-configuration
of a qualitative composition configuration (see Def.~\ref{def:qcc}) containing two
abstract dipoles (where for the second one, only the start or end point is used for
classification). All other data are the same as in Def.~\ref{def:qcc}.
\end{defn}

\begin{notation}
	To simplify the explanation of large classifiers, we shall write:
	\begin{eqnarray*}
	f(x) &=& \left\{
		\begin{array}{rcl}
			cond_1 &\longrightarrow& value_1\\
			cond_2 &\longrightarrow& value_2
		\end{array}\right.
	\end{eqnarray*}
	instead of
	\begin{eqnarray*}
	f(x)&=&\left\{
		\begin{array}{ll}
			value_1 & \mbox{if } cond_1\\value_2 & \mbox{if } cond_2.
		\end{array}\right.
	\end{eqnarray*}
	If it is clear which function we are defining, we even omit the ``$f(x) = $''.
\end{notation}

Given a primitive qualitative composition configuration $Q$, \emph{primitive classifiers}
map the qualitative locations of a dipole $dp_1$ and a point $pt$
(which is the start or end point of another dipole $dp_2$) to a letter
indicating the $\mathcal{LR}$
relation between the dipole and point.  We say that the dipole has
positive $pos$ orientation if $dir(dp) = +$, otherwise the orientation is
negative $neg$.

We need three different types of primitive classifiers for our algorithm.

\begin{figure}[ht]
  \begin{center}
    \includegraphics[keepaspectratio, scale=0.85]{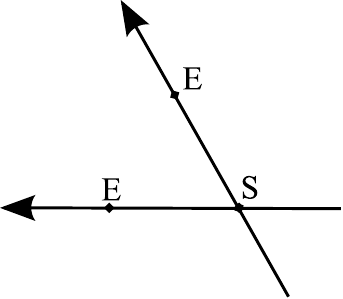}
  \caption{\label{fig:primitive}Line configuration for primitive Classifier}
  \end{center}
\end{figure}

Given two arbitrary dipoles $dp_1$ and $dp_2$,
we construct a primitive classifier for a pqcc with intersecting carrier
rays in its realization.
The classifier itself only works on $dp_1$ and $pt$, where $pt$ is either the
start or end point of $dp_2$.
A realization of this pqcc is given in Fig.~\ref{fig:primitive} for the reader's convenience,
the actual dipoles are omitted from the figure, since they can be placed arbitrarily.

To realize the dipole, this classifier takes dipole $dp_1$ and
the start or end point of $dp_2$ called $pt$ as well as information on whether
$dp_1$ is pointing in the same direction as the ray ($pos$) or
against it ($neg$) for both dipoles. The classifier returns an
$\mathcal{LR}$-relation determining the relation between $dp_1$ and $pt$.

In this case,
the classifier $cli_{x,y} (dp_1,pt)$ is given by:

\begin{eqnarray*}
  pos & \longrightarrow & \left\{
    \begin{array}{rcl}
      pt > y & \longrightarrow & R \\
      pt = y & \longrightarrow & \left\{
        \begin{array}{rcl}
          st(dp_1) < x \wedge ed(dp_1) < x & \longrightarrow & F \\
          st(dp_1) < x \wedge ed(dp_1) = x & \longrightarrow & E \\
          st(dp_1) < x \wedge ed(dp_1) > x & \longrightarrow & I \\
          st(dp_1) = x \wedge ed(dp_1) > x & \longrightarrow & S \\
          st(dp_1) > x \wedge ed(dp_1) > x & \longrightarrow & B
        \end{array} \right. \\
      pt < y & \longrightarrow & L \\
    \end{array}
  \right. \\
  neg &  \longrightarrow & \left\{
    \begin{array}{rcl}
      pt < y & \longrightarrow & R \\
      pt = y & \longrightarrow & \left\{
        \begin{array}{rcl}
          st(dp_1) > x \wedge ed(dp_1) > x & \longrightarrow & F \\
          st(dp_1) > x \wedge ed(dp_1) = x & \longrightarrow & E \\
          st(dp_1) > x \wedge ed(dp_1) < x & \longrightarrow & I \\
          st(dp_1) = x \wedge ed(dp_1) < x & \longrightarrow & S \\
          st(dp_1) < x \wedge ed(dp_1) < x & \longrightarrow & B
        \end{array} \right. \\
      pt > y & \longrightarrow & L \\
    \end{array}
  \right.
\end{eqnarray*}
The subscripts on the classifier denote the point of intersection of the
two lines. For the case shown in Fig.~\ref{fig:primitive}, we have
$x = y = S$.
We see that the table for $neg$ is exactly the complement of $pos$. This
primitive classifier assumes that, in the geometric realization,
the second dipole (containing
point $pt$) points to the right w.r.t. dipole $d$.
If the second dipole points to the left in the realization,
it is sufficient to apply an operation that interchanges
$L$ with $R$ on this classifier, in order to obtain the correct results. We will call
this operation $com$. This is the only primitive classifier needed for
intersecting lines.

\begin{figure}[ht]
\begin{center}
    \includegraphics[keepaspectratio, scale=0.85]{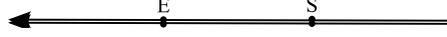}
    \caption{\label{fig:primSameline}Primitive classifier for same line.}
  \end{center}
\end{figure}

Secondly, we give a primitive classifier $cls (dp_1,pt)$
for two lines that coincide, see Fig.~\ref{fig:primSameline}.
{\tiny
\begin{eqnarray*}
  pos & \longrightarrow & \left\{
    \begin{array}{rcl}
      pt = F & \longrightarrow & \left\{
        \begin{array}{rcl}
          st(dp_1) < F \wedge ed(dp_1) < F &\longrightarrow& F \\
          st(dp_1) < F \wedge ed(dp_1) = F & \longrightarrow & \left\{
            \begin{array}{rcl}
              ed(dp_1) <_{p} pt &\longrightarrow& F \\
              ed(dp_1) =_{p} pt &\longrightarrow& E \\
              ed(dp_1) >_{p} pt &\longrightarrow& I
            \end{array}
          \right.\\
          st(dp_1) = F \wedge ed(dp_1) = F & \longrightarrow & \left\{
            \begin{array}{rcl}
              st(dp_1) <_p pt \wedge ed(dp_1) <_{p} pt &\longrightarrow& F \\
              st(dp_1) <_p pt \wedge ed(dp_1) =_{p} pt &\longrightarrow& E \\
              st(dp_1) <_p pt \wedge ed(dp_1) >_{p} pt &\longrightarrow& I \\
              st(dp_1) =_p pt \wedge ed(dp_1) >_{p} pt &\longrightarrow& S \\
              st(dp_1) >_p pt \wedge ed(dp_1) >_{p} pt &\longrightarrow& B \\
            \end{array}
          \right.\\
        \end{array}
      \right.\\
      pt = E & \longrightarrow & \left\{
        \begin{array}{rcl}
          st(dp_1) < E \wedge ed(dp_1) < E &\longrightarrow&  F \\
          st(dp_1) < E \wedge ed(dp_1) = E &\longrightarrow&  E \\
          st(dp_1) < E \wedge ed(dp_1) > E &\longrightarrow&  I \\
          st(dp_1) = E \wedge ed(dp_1) > E &\longrightarrow&  S \\
          st(dp_1) > E \wedge ed(dp_1) > E &\longrightarrow&  B \\
        \end{array}
      \right. \\
      pt = I & \longrightarrow & \left\{
        \begin{array}{rcl}
          st(dp_1) < I \wedge ed(dp_1) < I &\longrightarrow&  F \\
          st(dp_1) < I \wedge ed(dp_1) = I &\longrightarrow& \left\{
            \begin{array}{rcl}
              ed(dp_1) <_p pt &\longrightarrow&  F \\
              ed(dp_1) =_p pt &\longrightarrow&  E \\
              ed(dp_1) >_p pt &\longrightarrow&  I \\
            \end{array}
          \right. \\
          st(dp_1) < I \wedge ed(dp_1) > I &\longrightarrow&  I \\
          st(dp_1) = I \wedge ed(dp_1) = I &\longrightarrow& \left\{
            \begin{array}{rcl}
              st(dp_1) <_p pt \wedge ed(dp_1) <_p pt &\longrightarrow&  F \\
              st(dp_1) <_p pt \wedge ed(dp_1) =_p pt &\longrightarrow&  E \\
              st(dp_1) <_p pt \wedge ed(dp_1) >_p pt &\longrightarrow&  I \\
              st(dp_1) =_p pt \wedge ed(dp_1) >_p pt &\longrightarrow&  S \\
              st(dp_1) >_p pt \wedge ed(dp_1) >_p pt &\longrightarrow&  B \\
            \end{array}
          \right. \\
          st(dp_1) = I \wedge ed(dp_1) > I &\longrightarrow& \left\{
            \begin{array}{rcl}
              st(dp_1) <_p pt &\longrightarrow&  I \\
              st(dp_1) =_p pt &\longrightarrow&  S \\
              st(dp_1) >_p pt &\longrightarrow&  B \\
            \end{array}
          \right.\\
          st(dp_1) > I \wedge ed(dp_1) > I &\longrightarrow& B
        \end{array}
      \right. \\
      pt = S & \longrightarrow & \left\{
        \begin{array}{rcl}
          st(dp_1) < S \wedge ed(dp_1) < S &\longrightarrow& F \\
          st(dp_1) < S \wedge ed(dp_1) = S &\longrightarrow& E \\
          st(dp_1) < S \wedge ed(dp_1) > S &\longrightarrow& I \\
          st(dp_1) = S \wedge ed(dp_1) > S &\longrightarrow& S \\
          st(dp_1) > S \wedge ed(dp_1) > S &\longrightarrow& B \\
        \end{array}
      \right. \\
      pt = B & \longrightarrow & \left\{
        \begin{array}{rcl}
          st(dp_1) = B \wedge ed(dp_1) = B &\longrightarrow& \left\{
            \begin{array}{rcl}
              st(dp_1) <_p pt \wedge ed(dp_1) <_p pt &\longrightarrow&  F \\
              st(dp_1) <_p pt \wedge ed(dp_1) =_p pt &\longrightarrow&  E \\
              st(dp_1) <_p pt \wedge ed(dp_1) >_p pt &\longrightarrow&  I \\
              st(dp_1) =_p pt \wedge ed(dp_1) >_p pt &\longrightarrow&  S \\
              st(dp_1) >_p pt \wedge ed(dp_1) >_p pt &\longrightarrow&  B \\
            \end{array}
          \right. \\
          st(dp_1) = B \wedge ed(dp_1) > B &\longrightarrow& \left\{
            \begin{array}{rcl}
              st(dp_1) <_p pt &\longrightarrow&  I \\
              st(dp_1) =_p pt &\longrightarrow&  S \\
              st(dp_1) >_p pt &\longrightarrow&  B \\
            \end{array}
          \right. \\
          st(dp_1) > B \wedge ed(dp_1) > B &\longrightarrow& B
        \end{array}
      \right.
    \end{array}
  \right. \\
  neg &  \longrightarrow & \left\{
    \begin{array}{rcl}
      pt = B & \longrightarrow & \left\{
        \begin{array}{rcl}
          st(dp_1) > B \wedge ed(dp_1) > B &\longrightarrow& F \\
          st(dp_1) > B \wedge ed(dp_1) = B & \longrightarrow & \left\{
            \begin{array}{rcl}
              ed(dp_1) <_{p} pt &\longrightarrow& I \\
              ed(dp_1) =_{p} pt &\longrightarrow& E \\
              ed(dp_1) >_{p} pt &\longrightarrow& F
            \end{array}
          \right.\\
          st(dp_1) = B \wedge ed(dp_1) = B & \longrightarrow & \left\{
            \begin{array}{rcl}
              st(dp_1) <_p pt \wedge ed(dp_1) <_{p} pt &\longrightarrow& B \\
              st(dp_1) =_p pt \wedge ed(dp_1) <_{p} pt &\longrightarrow& S \\
              st(dp_1) >_p pt \wedge ed(dp_1) <_{p} pt &\longrightarrow& I \\
              st(dp_1) >_p pt \wedge ed(dp_1) =_{p} pt &\longrightarrow& E \\
              st(dp_1) >_p pt \wedge ed(dp_1) >_{p} pt &\longrightarrow& F \\
            \end{array}
          \right.\\
        \end{array}
      \right.\\
      pt = S & \longrightarrow & \left\{
        \begin{array}{rcl}
          st(dp_1) > S \wedge ed(dp_1) > S &\longrightarrow&  F \\
          st(dp_1) > S \wedge ed(dp_1) = S &\longrightarrow&  E \\
          st(dp_1) > S \wedge ed(dp_1) < S &\longrightarrow&  I \\
          st(dp_1) = S \wedge ed(dp_1) < S &\longrightarrow&  S \\
          st(dp_1) < S \wedge ed(dp_1) < S &\longrightarrow&  B \\
        \end{array}
      \right. \\
      pt = I & \longrightarrow & \left\{
        \begin{array}{rcl}
          st(dp_1) > I \wedge ed(dp_1) > I &\longrightarrow&  F \\
          st(dp_1) > I \wedge ed(dp_1) = I &\longrightarrow& \left\{
            \begin{array}{rcl}
              ed(dp_1) >_p pt &\longrightarrow&  F \\
              ed(dp_1) =_p pt &\longrightarrow&  E \\
              ed(dp_1) <_p pt &\longrightarrow&  I \\
            \end{array}
          \right. \\
          st(dp_1) > I \wedge ed(dp_1) < I &\longrightarrow&  I \\
          st(dp_1) = I \wedge ed(dp_1) = I &\longrightarrow& \left\{
            \begin{array}{rcl}
              st(dp_1) >_p pt \wedge ed(dp_1) >_p pt &\longrightarrow&  F \\
              st(dp_1) >_p pt \wedge ed(dp_1) =_p pt &\longrightarrow&  E \\
              st(dp_1) >_p pt \wedge ed(dp_1) <_p pt &\longrightarrow&  I \\
              st(dp_1) =_p pt \wedge ed(dp_1) <_p pt &\longrightarrow&  S \\
              st(dp_1) <_p pt \wedge ed(dp_1) <_p pt &\longrightarrow&  B \\
            \end{array}
          \right. \\
          st(dp_1) = I \wedge ed(dp_1) < I &\longrightarrow& \left\{
            \begin{array}{rcl}
              st(dp_1) >_p pt &\longrightarrow&  I \\
              st(dp_1) =_p pt &\longrightarrow&  S \\
              st(dp_1) <_p pt &\longrightarrow&  B \\
            \end{array}
          \right.\\
          st(dp_1) < I \wedge ed(dp_1) < I &\longrightarrow& B
        \end{array}
      \right. \\
      pt = E & \longrightarrow & \left\{
        \begin{array}{rcl}
          st(dp_1) > E \wedge ed(dp_1) > E &\longrightarrow& F \\
          st(dp_1) > E \wedge ed(dp_1) = E &\longrightarrow& E \\
          st(dp_1) > E \wedge ed(dp_1) < E &\longrightarrow& I \\
          st(dp_1) = E \wedge ed(dp_1) < E &\longrightarrow& S \\
          st(dp_1) < E \wedge ed(dp_1) < E &\longrightarrow& B \\
        \end{array}
      \right. \\
      pt = F & \longrightarrow & \left\{
        \begin{array}{rcl}
          st(dp_1) = F \wedge ed(dp_1) = F &\longrightarrow& \left\{
            \begin{array}{rcl}
              st(dp_1) >_p pt \wedge ed(dp_1) >_p pt &\longrightarrow&  F \\
              st(dp_1) >_p pt \wedge ed(dp_1) =_p pt &\longrightarrow&  E \\
              st(dp_1) >_p pt \wedge ed(dp_1) <_p pt &\longrightarrow&  I \\
              st(dp_1) =_p pt \wedge ed(dp_1) <_p pt &\longrightarrow&  S \\
              st(dp_1) <_p pt \wedge ed(dp_1) <_p pt &\longrightarrow&  B \\
            \end{array}
          \right. \\
          st(dp_1) = F \wedge ed(dp_1) < F &\longrightarrow& \left\{
            \begin{array}{rcl}
              st(dp_1) >_p pt &\longrightarrow&  I \\
              st(dp_1) =_p pt &\longrightarrow&  S \\
              st(dp_1) <_p pt &\longrightarrow&  B \\
            \end{array}
          \right. \\
          st(dp_1) < F \wedge ed(dp_1) < F &\longrightarrow& B
        \end{array}
      \right.
    \end{array}
  \right.
\end{eqnarray*}
}

This classifier looks a little cumbersome, but we decided to use it in this way,
so that all impossible cases w.r.t. the ordering of the line are
excluded. This gives better error handling capabilities in an implementation of
it, since impossible cases can be detected. A more compressed version is possible,
but it cannot detect impossible cases anymore. All cases that are
not listed in the above classifier are cases where the ordering $>_p$ is not
compatible with the segmentation, and so they are impossible.
This is the only classifier for coinciding lines.

\begin{figure}[ht]
\begin{center}
    \includegraphics[keepaspectratio, scale=0.85]{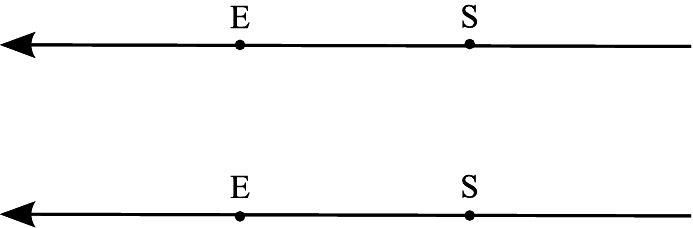}
    \caption{\label{fig:primParallel}Primitive classifier for parallel lines.}
  \end{center}
\end{figure}

The third classifier is for parallel lines, i.e. a configuration like that
in Fig.~\ref{fig:primParallel}. Let the lower line be the line the dipole lies on.
The information about the line on which the dipole lies is handled by a basic classifier
which uses this primitive classifier and exchanges $L$ and $R$ appropriately.
Fortunately this classifier $clpar (dp_1,pt)$ is simple:
\begin{eqnarray*}
  pos &\longrightarrow& R \\
  neg &\longrightarrow& L \\
\end{eqnarray*}
This is the only classifier for parallel lines.

This is a complete list of the basic classifiers that are needed.

\subsection{Basic Classifiers\label{sec:basicclass}}
\begin{figure}[ht]
  \begin{center}
    \includegraphics[keepaspectratio, scale=0.85]{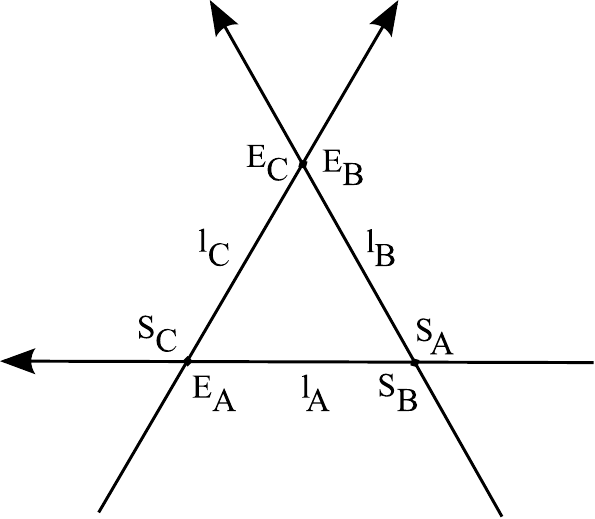}
  \caption{\label{fig:Basic}Line configuration for Basic Classifier}
  \end{center}
\end{figure}

Based on the primitive classifiers introduced in Sect.~\ref{sec:primclass}, we
construct the \emph{basic classifiers} to determine the $\mathcal{DRA}$
relations in scenarios. For $\DRAf$, we always need exactly four
primitive classifiers to determine the relation. For $\DRAfp$, in
some cases we need an additional fifth classifier to determine the qualitative
angle. We will first focus on the $\DRAf$ case. Given a qcc,
we apply four basic classifiers three times: namely (1) to
the first and second abstract dipole, (2) to the second and third and (3) to the
first and third. Thus, we obtain an entry in the composition table.
Consider a qcc with $i = 1+$ and $a(S_A) = \hat s_{AB}$, $a(S_B) = \hat s_{AB}$ and
$a(s_C) = \hat S_{AC}$. Such a configuration has a realization as in
Fig.~\ref{fig:Basic}.
The dipole $d_X$ lies on the ray $l_X$ for $X \in \left\{A, B, C\right\}$.
We now apply primitive classifiers to this scenario in the way defined in
Section~\ref{basic}. Hence, we get the basic classifier for such a configuration:
\begin{eqnarray*}
  R(dp_A, st_B) &=& cli_{s,s} (dp_A,st_B) \\
  R(dp_A, ed_B) &=& cli_{s,s} (dp_A,ed_B) \\
  R(dp_B, st_A) &=& com \circ cli_{s,s} (dp_B,st_A) \\
  R(dp_B, ed_A) &=& com \circ cli_{s,s} (dp_B,ed_A) \\ \\
  R(dp_B, st_C) &=& cli_{e,e} (dp_B, st_C) \\
  R(dp_B, ed_C) &=& cli_{e,e} (dp_B, ed_C) \\
  R(dp_C, st_B) &=& com \circ cli_{e,e} (dp_C, st_B) \\
  R(dp_C, st_B) &=& com \circ cli_{e,e} (dp_C, ed_B) \\ \\
  R(dp_A, st_C) &=& cli_{e,s} (dp_A, st_C) \\
  R(dp_A, ed_C) &=& cli_{e,s} (dp_A, ed_C) \\
  R(dp_C, st_A) &=& com \circ cli_{s,e} (dp_C, st_A) \\
  R(dp_C, ed_A) &=& com \circ cli_{s,e} (dp_C, ed_A) \\
\end{eqnarray*}\sloppypar
and we obtain the relation between $dp_A$ and $dp_B$:
$\varrho(R(dp_A, st_B),R(dp_A, ed_B),R(dp_B, st_A),R(dp_B, ed_A))$.
The relations between $d_B$ and $d_C$ as well as between
$dp_A$ and $dp_C$ are derived analogously.
The basic classifiers depend on the configuration in which
the qcc realization lies and on the angle between the rays
in the realization. They are constructed for an angle between the rays in the interval
$(\pi, 2\cdot\pi]$. If the angle is in the interval
$(0, \pi]$, the $\mathcal{LR}$ relation between any line on the first ray
and a point on the second just swaps. We capture this by introducing the
operation $com$ which is applied in this case. With it, we can limit the
number of necessary primitive classifiers.
The construction of the other basic classifiers is
done analogously.

\subsection{Extended Basic Classifiers for $\DRAfp$
  \label{sec:basicclassfp}}

\begin{figure}[ht]
  \begin{center}
    \begin{tabular}{ccc}
    \includegraphics[keepaspectratio, scale=0.85]{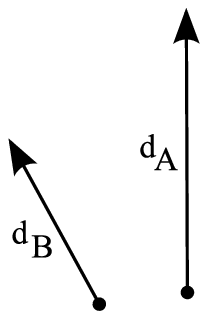} &~\hspace{3cm}~&
    \includegraphics[keepaspectratio, scale=0.85]{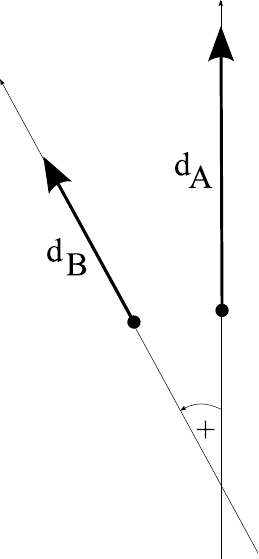}
    \end{tabular}
    \caption{\label{fig:drascen+}$\mathcal{DRA}$ Scenario}
  \end{center}
\end{figure}

For $\DRAfp$, basically the same classifiers as
described for $\DRAf$ in Section~\ref{sec:basicclass} are used. We simply extend them for
the relations \relnsp{rrrr}, \relnsp{rrll}, \relnsp{llll} and \relnsp{llrr} to
classify the information about qualitative angles.
For this purpose, we have to have a look at the angles between dipoles in the
realization of a given qcc.
The qualitative angle between two dipoles $d_A$ and $d_B$
is called positive
$+$ (negative $-$) if the angle from the carrier ray of $d_A$ called $l_A$ to
the carrier ray of $d_B$ called $l_B$ lies in the interval
$(0,\pi)$ ($(\pi, 2\cdot\pi)$).
 We give an example of this. Consider
the configuration of a $\mathcal{DRA}$ scenario in Fig.~\ref{fig:drascen+} on
the left hand side.
On the right-hand side of Fig.~\ref{fig:drascen+}, the carrier rays are
introduced and we can see that the angle clearly lies in the interval
$(0,\pi)$ and hence the qualitative angle is positive. The definitions of
parallel $P$ and anti-parallel $A$ are straightforward.
The set $a^{-1}(\hat S_{xy})$ always contains exactly two elements, if
$\hat S_{xy} \in \hat \mathcal{S}(i)$. To continue, we need functions
$proj_x: \mathcal{P} (\mathbf{P}) \longrightarrow
\mathcal{P}(\mathbf{P})$
defined as
\[
proj_x = \left\{a \mid idx_x(a) = x \right\}
\]
which form the set of all elements with index ($idx$) x. $\mathcal{P}$ denotes
powerset formation. By the definition of $a$ and the sets $\hat \mathcal{S}(i)$,
these sets are always singletons, if $proj_x \circ a^{-1}$ is applied to an
intersection point and if $a^{-1}$ contains an element with index $x$, otherwise
the set is empty. We shall write $a_x^{-1}$ for $proj_x \circ a^{-1}$.

We observed that the qualitative angles between two dipoles can be classified
very easily once the $\DRAf$ relations between the dipoles $d_A$ and
$d_B$ are known. All
we need to do is to find out if the ray $l_B$ intersects $l_A$ in
front of or behind $d_A$. In the language of qcc and abstract dipoles
$dp_A$ and $dp_B$, we can say that,
if $a^{-1}_A (\hat S_{AB}) > ed(dp_A)$ for $dir(dp_A) = +$,
or if $a^{-1}_A (\hat S_{AB}) <  ed(dp_A)$, if $dir(dp_A) = -$, then the abstract
point of intersection lies ``in front of $dp_A$'' and, if
$st(dp_A) > a^{-1}_A (\hat S_{AB})$ for $dir(dp_A) = +$ or $a^{-1}_A (\hat S_{AB}) >  st(dp_A)$ for
$dir(dp_A) = -$, the abstract point of intersection lies ``behind $dp_A$''.

\begin{prop}\label{prop:intersectfront}
  In a realization $R(Q)$ of a qcc $Q$, the carrier rays of any
  two dipoles $d_1$ and $d_2$ intersect in front of $d_1$ if and only if,
  in $Q$ the property
  \[
  (a^{-1}_1 (\hat S_{12}) > ed(dp_1) \wedge dir(dp_1) = +) \vee
  (a^{-1}_1 (\hat S_{12}) <  ed(dp_1) \wedge dir(dp_1) = -)
  \]
  is fulfilled.
\end{prop}
\begin{pf}
  This is immediate by inspection of the property and respective scenarios.
\end{pf}

\begin{prop}\label{prop:intersectback}
  In a realization $R(Q)$ of a qcc $Q$, the carrier rays of any
  two dipoles $d_1$ and $d_2$ intersect behind $d_1$ if and only if,
  in $Q$ the property
  \[
  (st(dp_1) > a^{-1}_1 (\hat S_{12}) \wedge dir(dp_1) = +) \vee
  (a^{-1}_1 {\hat S_{12}} >  st(dp_1) \wedge dir(dp_1) = -)
  \]
  is fulfilled.
\end{prop}
\begin{pf}
  This is immediate by inspection of the property and respective scenarios.
\end{pf}

The complete
extension for the Basic Classifiers is given as:
\begin{eqnarray*}
\relnsp{rrrr} & \longrightarrow & \left\{
  \begin{array}{rcl}
    a^{-1}_A (\hat S_{AB}) > ed(dp_A) \wedge dir(dp_A) = + & \longrightarrow & - \\
    a^{-1}_A (\hat S_{AB}) <  ed(dp_A) \wedge dir(dp_A) = - & \longrightarrow & - \\
    st(dp_A) > a^{-1}_A (\hat S_{AB}) \wedge dir(dp_A) = + & \longrightarrow & + \\
    a^{-1}_ A (\hat S_{AB}) >  st(dp_A) \wedge dir(dp_A) = - & \longrightarrow & +
  \end{array}
  \right.  \\
\relnsp{rrll} & \longrightarrow & \left\{
  \begin{array}{rcl}
    a^{-1}_A (\hat S_{AB}) > ed(dp_A) \wedge dir(dp_A) = + & \longrightarrow & + \\
    a^{-1}_A (\hat S_{AB}) <  ed(dp_A) \wedge dir(dp_A) = - & \longrightarrow & + \\
    st(dp_A) > a^{-1}_A (\hat S_{AB}) \wedge dir(dp_A) = + & \longrightarrow & - \\
    a^{-1}_A (\hat S_{AB}) >  st(dp_A) \wedge dir(dp_A) = - & \longrightarrow & -
  \end{array}
  \right. \\
\relnsp{llll} & \longrightarrow & \left\{
  \begin{array}{rcl}
    a^{-1}_A(\hat S_{AB}) > ed(dp_A) \wedge dir(dp_A) = + & \longrightarrow & + \\
    a^{-1}_A (\hat S_{AB}) <  ed(dp_A) \wedge dir(dp_A) = - & \longrightarrow & + \\
    st(dp_A) > a^{-1}_A (\hat S_{AB}) \wedge dir(dp_A) = + & \longrightarrow & - \\
    a^{-1}_A (\hat S_{AB}) >  st(dp_A) \wedge dir(dp_A) = - & \longrightarrow & -
  \end{array}
  \right. \\
\relnsp{llrr} & \longrightarrow & \left\{
  \begin{array}{rcl}
    a^{-1}_A (\hat S_{AB}) > ed(dp_A) \wedge dir(dp_A) = + & \longrightarrow & - \\
    a^{-1}_A (\hat S_{AB}) <  ed(dp_A) \wedge dir(dp_A) = - & \longrightarrow & - \\
    st(dp_A) > a^{-1}_A (\hat S_{AB}) \wedge dir(dp_A) = + & \longrightarrow & + \\
    a^{-1}_A (\hat S_{AB}) >  st(dp_A) \wedge dir(dp_A) = - & \longrightarrow & +
  \end{array}
  \right.
\end{eqnarray*}
Constructing the classifiers for qccs based on configurations with parallel
lines is easy, depending on the $\DRAf$-relations, the dipoles can either
be parallel or anti-parallel in such cases, but never both at the same
time.

\begin{lem}\label{lem:same-relations-intersection}
  Given two intersecting lines, the $\mathcal{LR}$-relations between
  a dipole on a first line and a point on the second line are stable under
  the movement of the point along the line,
  unless it moves through the point of intersection of the two lines.
\end{lem}
\begin{pf}
  By the definition of $\mathcal{LR}$-relations, the point can be in one of three
  different relative positions to the carrier ray of the dipole. The point can lie on either
  side of the point of intersection, yielding the relation $L$ or $R$,
  or on the point of intersection itself, yielding exactly one
  relation on the line.
\end{pf}

\begin{lem}\label{lem:same-relations-same-line}
  Given a dipole and a point lying on its carrier line, the
  $\mathcal{LR}$-relations between the dipole and point
  are stable under the movement of the point along the line,
  unless it is moved over the start
  or end point of the dipole.
\end{lem}
\begin{pf}
  Inspect the definition of $\mathcal{LR}$-relations on a line.
\end{pf}

\begin{lem}\label{lem:dipole-stable-intersection}
  For dipoles lying on intersecting rays, the $\mathcal{DRA}$ relations are
  stable under the movement of the start and end points of the dipoles
  along the rays, as long as the segments for the start and end points and the
  directions of the dipoles do not change.
\end{lem}
\begin{pf}
  We observe that the segmentation is a stronger property than the
  one used in Lemma~\ref{lem:same-relations-intersection}. For
  $\DRAf$ relations it suffices to apply Lemma~\ref{lem:same-relations-intersection}
  four times. For $\DRAfp$ relations, we also need to take the intersection
  property of Prop.~\ref{prop:easyclass} into account.
\end{pf}

\begin{lem}\label{lem:dipole-stable-same-line}
  For dipoles on the same line, the $\mathcal{DRA}$-relations are stable
  under the movement of the start and end points of the dipoles
  along the rays,
  so long as the relation $<_r$ does not change.
\end{lem}
\begin{pf}
  Apply Lemma~\ref{lem:same-relations-same-line} four times.
\end{pf}

\begin{lem}\label{lem:affine-dipole-trafo-preserves}
  \begin{enumerate}
  \item Transforming a given realization of a qcc along an orientation-preserving
    affine transformation preserves the segmentation map.
  \item If two dipoles are
    on the same line, affine transformations also preserve
    $<_r$.
  \end{enumerate}
\end{lem}
\begin{pf}
  \noindent
  1) According to Prop.~\ref{thm:automorphisms}, any orientation-preserving
    affine transformation preserves the $\LR$ relations.\\
  \noindent
  2) This follows from the preservation of length ratios by
  affine transformations, i.e. the length ratios between the
  start and end points of the dipoles and points
  $S$ and $E$ on the ray.
\end{pf}

\begin{lem}\label{lem:equvialent-realisations}
Given a qcc, any two geometric realizations exhibit the same
$\mathcal{DRA}$-relations among their dipoles.
\end{lem}
\begin{pf}
  Let two geometric realizations $R_1$, $R_2$ of a qcc $Q$ be given.
  Since the line triples of $R_1$ and $R_2$ belong to the same
  orbit, there is an orientation-preserving affine bijection $f$
  transforming the line triple of $R$ into that of $R'$.
  In case of configurations 5a, 5b and 5c, we assume that all distance
  ratios are adjusted to 1 in order to reach the same orbit.
  Note that this adjustment, although not an affine transformation,
  does not affect the relations between dipoles.

  Since $f$ maps $R_1$'s line triple to $R_2$'s line triple, it also
  maps the corresponding points of intersection to each other.  For
  orbits $1+$ and $1-$, all segmentation points are points of intersection.
  Hence, $f$ does not change the segments given by $r(x)$ in which the
  start and end points of the dipoles lie.  For the rest of the
  argument, apply Lemma~\ref{lem:dipole-stable-intersection}.

  For cases $2+$ and $2-$, we just have a single point of intersection,
  but the relative directions of the rays are restricted by the
  definition of a realization and so is the location of all segmentation
  points w.r.t. the intersection point, as are the locations of the
  start and end points of the dipoles w.r.t. the segmentation points. For the
  rest of the argument, apply Lemma~\ref{lem:dipole-stable-intersection}.

  In cases $3a$, $3b$ and $3c$, we have two intersection points and
  two segmentation points that are not points of intersection but, as before,
  the directions of the rays and the locations of all segmentation points are
  restricted and hence the locations of the start and end points of the dipoles,
  and again, we can apply Lemma~\ref{lem:dipole-stable-intersection}.

  In cases $4a$, $4b$ and $4c$, we have one point of intersection and $3$
  segmentation points that are not points of intersection. First, we can argue
  to restrict the location and direction. In the end, we can apply
  Lemma~\ref{lem:dipole-stable-same-line} and Lemma~\ref{lem:dipole-stable-intersection}.

  In cases $5a$, $5b$ and $5c$, we only have segmentation points that
  are not points of intersection, but all rays have the same directions and
  the relative orientations of segmentation points on the line are restricted.
  Hence, the directions of the dipoles do not change during the mapping and
  the relative direction between dipoles is all that is necessary
  to determine their $\mathcal{DRA}$-relations in the case of parallel
  dipoles.

  The proof of cases $6a$, $6b$ and $6c$ is similar to cases $4$ and
  $5$, with the argument based on Lemma~\ref{lem:dipole-stable-same-line} for
  dipoles on the same line, and the arguments of cases $5$ for
  parallel lines.

  For case $7$, we need to apply Lemma~\ref{lem:dipole-stable-same-line}.

  For additional arguments for $\DRAfp$-relations, please refer to the proof
  of Prop.~\ref{prop:easyclass}.

\end{pf}

\begin{thm}[Correctness of the Construction]\label{thm:correctness}
 Given a qcc $Q$ and an arbitrary geometric realization $R(Q)$ of it, the
 $\DRAf$ relation in $R(Q)$ is the same
 as that computed by the basic classifiers on $Q$.
\end{thm}
\begin{pf}
  According to Lemma~\ref{lem:equvialent-realisations}, we can focus on one
  geometric realization per qcc.

  For this proof, we need to inspect once more the construction of the
  basic classifiers above the primitive classifiers.
  The actual values of $a$, $dir$ and the start and end points
  of the abstract dipoles as well as the order $<_p$ are not
  directly used by basic classifiers\footnote{With the exception of the
  extended classifiers, but we will discuss these later}. They
  are passed through to primitive classifiers. The only
  information that is directly used in basic classifiers
  is the identifier $i$ of the configuration.

  We divide this proof in two steps. In the first step,
  we show that the primitive classifiers are correct
  and, in the second step, we do the same for basic classifiers.
  \noindent
  We will show a proof for the classifier $cli_{S,S}(dp_1,pt)$
  and a pqcc with $dir_{dp_1} = +$, $dp_1=(I,I)$ and $pt=I$. A realization
  of this configuration is shown in Fig.~\ref{fig:pfConf}
  \begin{figure}[htb]
    \begin{center}
      \includegraphics[keepaspectratio, scale=0.85]{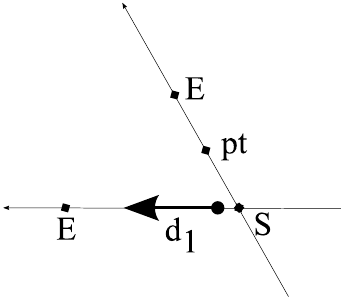}
      \caption{\label{fig:pfConf}A realization}
    \end{center}
  \end{figure}
  and we can easily see that $d_1 \relnsp{R} pt$ has to be true.
  By observing $cli_{S,S}(dp_1,pt)$, we see that we are in the
  case $pos$ and that $pt > S$ and so the primitive classifier
  also yields $dp_1 \rel{R} pt$ as expected. All other proofs
  for pqccs are done in an analogous way by inspection of the
  relations yielded by the primitive classifiers and their realizations.
  \noindent
  With primitive classifiers working correctly, we need to focus on the
  basic classifiers. Here, we will show this for the case $i = 1+$, all other
  cases are handled in an analogous fashion. First we take any
  realization of $i = 1+$ and add directions to the lines as
  described in the section about geometric realizations of qccs. For example,
  the one depicted in Fig.~\ref{fig:pfOrbitReal}.
  \begin{figure}[htb]
    \begin{center}
      \includegraphics[keepaspectratio, scale=0.85]{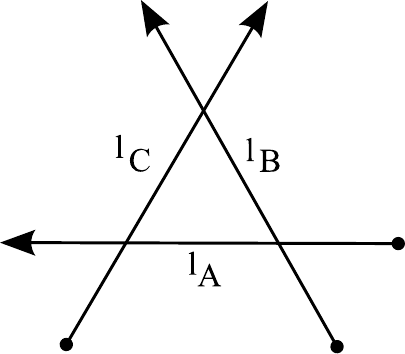}
      \caption{\label{fig:pfOrbitReal}A realization for a qcc}
    \end{center}
  \end{figure}
  In the next step, this realization is decomposed according to the
  definition of $\DRAf$-relations and the basic classifiers shown in
  Fig.~\ref{fig:pfDecomposition}.
  \begin{figure}[htb]
    \begin{center}
      \includegraphics[keepaspectratio, scale=0.85]{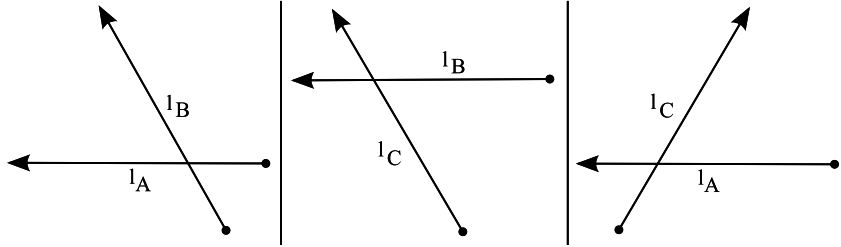}
      \caption{\label{fig:pfDecomposition}Decomposition of line configuration}
    \end{center}
  \end{figure}
  The various parts of the decomposed line configuration need to be matched
  with the realization of the primitive classifier, here 
  the realization of Fig.~\ref{fig:primitive}. In our case, the
  classifier matches directly with the orientations from $l_A$ to
  $l_B$, $l_B$ to $l_C$ and $l_A$ to $l_C$. In the other cases, the
  angle between the lines may be inverted. Then, we need to
  swap $R$ and $L$ which is done by the operation $com$. Furthermore, we
  see that the lines $l_C$ and $l_B$ both intersect in
  segment $E$, whereas $l_A$ and $l_B$ intersect both in $S$. The intersection
  for $l_A$ and $l_C$ is $E$ for $l_A$ and $S$ for $l_C$, we need to
  parameterize the respective primitive classifiers with that information.
  But in the end, our arguments yield exactly the basic classifier shown in
  Section~\ref{sec:basicclass}. The arguments for the other $16$ basic classifiers
  are analogous.
\end{pf}
\begin{prop}\label{prop:easyclass}
Given any qcc $Q$ and its geometric realization $R(Q)$, the extended basic
classifiers determine the same $\DRAfp$ relation as in the
realization.

\end{prop}
\begin{pf}
We assume that the $\DRAf$ relation is determined correctly. All
we need to consider here are the ``extended'' relations.

We will give the proof for \relnsp{rrrr-}, the proof for the other cases is
analogous. Consider two dipoles $d_A$ and $d_B$ in an \relnsp{rrrr}
configuration on the rays $l_A$ and $l_B$.
There are two classes of qualitatively distinguishable configurations for
$(d_A \rel{rrrr} d_B)$:
\begin{center}
  \includegraphics[keepaspectratio, scale=0.85]{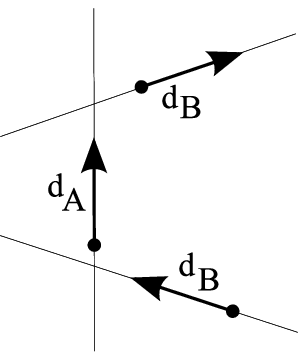}
\end{center}
We can see that $l_B$ intersects $l_A$ either in front of or
behind $d_A$. If the intersection point lies in
front of $d_A$, we are in a situation like
\begin{center}
  \includegraphics[keepaspectratio, scale=0.85]{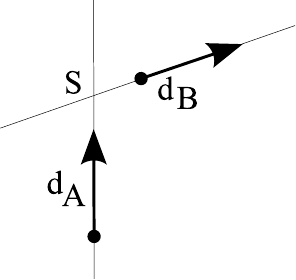}
\end{center}
where $S$ is the intersection point. We can further see that the angle
from $l_A$ to $l_B$ lies clearly in the interval $(\pi,2\cdot\pi)$. Furthermore,
$l_B$ can be rotated in the whole interval $(\pi,2\cdot\pi)$ without changing
the $\DRAf$ relation. Using this, we obtain the $\DRAfp$-relation
\relnsp{rrrr-} between $d_A$ and $d_B$ if the point of intersection $S$
lies in front of $d_A$. For any qcc belonging to such a scenario, the rest of
the proof follows from Prop.~\ref{prop:intersectfront} and
Prop.~\ref{prop:intersectback} as well as the inspection of the extended
classifiers:
\begin{eqnarray*}
    \hat S_{AB} > ed(dp_A) \wedge dir(dp_A) = + & \longrightarrow & - \\
    \hat S_{AB} <  ed(dp_A) \wedge dir(dp_A) = - & \longrightarrow & - \\
\end{eqnarray*}
But these also yield $(dp_A \rel{rrrr-} dp_B)$.
By the same arguments, we show that $(d_A \rel{rrrr+} d_B)$ if the point of
intersection of $l_A$ and $l_B$ lies behind $d_A$.
The proof for all other cases
is analogous.
\end{pf}

\begin{cor}\label{prop:DRArelations}
The 72 relations in Fig.~\ref{fig:atomicrel} are those out of the
2401 formal combinations of four $\LR$ letters that are geometrically
possible.
\end{cor}
\begin{pf}
  By an exhaustive inspection of the primitive classifiers which
  occur in the basic classifiers for all pqccs. For the decomposition,
  we refer to the proof of Thm.~\ref{thm:correctness}.
\end{pf}

\begin{thm}\label{thm:correctnessdrafp}
 Given a qcc $Q$ and an arbitrary geometric realization $R(Q)$ of it, the
 $\DRAfp$ relation in $R(Q)$ is the same
 as that computed by the basic classifiers on $Q$.
\end{thm}
\begin{pf}
  Follows from Thm.~\ref{thm:correctness} and Prop.~\ref{prop:easyclass}.
\end{pf}

\subsection{Implementation of the Classification Procedure\label{sec:implementation}}

Qualitative composition configurations can be naturally represented as
a finite datatype. The classifiers are implemented as simple programs
(mainly case distinctions) that operate on $qccs$ in the sense of
Def.~\ref{def:qcc}. The classifiers are chosen with respect to the
identifier $i$ and the assignment mapping $a$ of the $qcc$.  In our
particular implementation, we exploited some symmetries to limit the
number of classifiers that we had to implement.

With the condensed semantics, we are able to compute the composition
tables of the $\mathcal{DRA}$ calculi in an efficient way. In fact we
have implemented the computation of composition tables for both
$\DRAf$ and $\DRAfp$ as Haskell programs, making use of {Haskell}'s
parallelism extensions.
The Haskell implementations of the basic classifiers for
$\DRAf$ and $\DRAfp$ are written in such a way that they
share a library of primitive classifiers. In these
programs, we further generate all qccs in an optimized way, i.e. we
only generate the order $<_p$ if it is needed, and classify them
with our basic classifiers. In the end, we compose our results into
composition tables. For the case where three lines are collinear, we simply
decided to enumerate all possible locations of points in a
certain interval for reasons of simplicity and this did not increase the
overall runtime too much.

The computation of the
composition table for $\DRAf$ takes less than one minute on a Notebook
with an Intel Core 2 T7200 with $1.5$ Gbyte of RAM, and the computation
of the composition table for $\DRAfp$ takes less than two minutes on
the same computer. This is a great advancement compared to the
enumeration of scenarios on a grid, which took several weeks to
compute only an approximation to the composition table.

\subsection{Properties of the Composition\label{sec:props}}
We have investigated several properties of the composition tables for
$\DRAf$ and $\DRAfp$. For both tables the properties
\begin{eqnarray*}
  id^{\smile} &=& id \\
  {\left(R^{\smile}\right)}^{\smile} &=& R \\
  id \circ R &=& R \\
  R \circ id &=& R \\
  {\left(R_1 \circ R_2\right)}^{\smile} &=& R_2^{\smile} \circ R_1^{\smile} \\
  R_1^{\smile} \in R_2 \circ R_3 &\iff& R_3^{\smile} \in R_1 \circ R_2
\end{eqnarray*}
hold with $R$, $R_1$, $R_2$, $R_3$ being any base-relation and $id$ the
identical relation. These properties can be automatically tested by the
\texttt{GQR} and \texttt{SparQ} qualitative reasoners. The other properties for a non-associative
algebra follow trivially.
Furthermore, we have tested the associativity
of the composition. For $\DRAf$, we have $373248$ triples of
relations to consider of which $71424$ are not associative. So the composition
of $19.14\%$ of all possible triples of relations is not associative\footnote{In the master thesis of
one of our students, a detailed analysis of a specific
non-associative dipole configuration is presented \cite{FlorianMossakowskiDA2007}}, e.g.
\begin{eqnarray*}
  (\relnsp{rrrl} ; \relnsp{rrrl}) ; \relnsp{llrl} &\neq& \relnsp{rrrl} ; (\relnsp{rrrl} ; \relnsp{llrl}).
\end{eqnarray*}
For $\DRAfp$ all $512000$ triples of base-relations are
associative w.r.t. composition. With this result, we obtain that
$\DRAfp$ is a relation algebra in a strict sense.

\subsection{$\DRAf$ composition is weak}
\label{sec:weak-comp}

The failure of $\DRAf$ to be associative may imply that its
composition is also weak. We will investigate this in this section. First,
recall the definition of strong composition. Furthermore, the composition of
$\mathcal{OPRA}_1$ is known to be weak \cite{Frommberger2007}, but by
Ex.~\ref{exa:DRAfpDRAopp} and Prop.~\ref{prop:preserveStrength}, then
$\DRAf$ also has a weak composition.

\begin{defn}\label{def:strong_comp}
A Qualitative Composition is called \emph{strong} if, for any arbitrary pair
of objects $A$, $C$ in the domain in relation $A r_{ac} C$, there is for
every entry in the composition table that contains $A r_{ac} C$ on the right
hand side, an object $B$ such that $A r_{ab} B$ and $B r_{bc} C$ reconstruct this
entry.
\end{defn}

We will show now that the defining property of strong composition
(see Sect.~\ref{ConstraintPropagationEtc}) is
violated for $\DRAf$.

\begin{prop}
The composition of $\DRAf$ is weak.
\end{prop}
\begin{pf}
Consider the $\DRAf$ composition $A \rel{BFII} B;
B \rel{LLLB} C \mapsto A \rel{LLLL} C$. We show that there are
dipoles $A$ and $B$ such that there is no dipole $B$ which reflects the
composition. Consider dipoles $A$ and $B$ as shown in Fig.~\ref{fig:weakcomp}.
\begin{figure}[ht]
\begin{center} \includegraphics[keepaspectratio, scale=0.85]{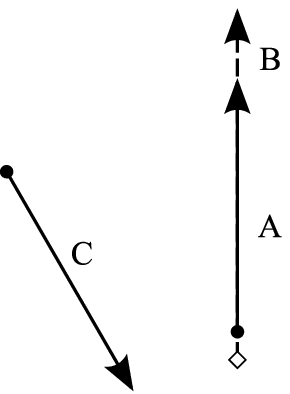}
\caption{$\DRAf$ weak composition
\label{fig:weakcomp}}
\end{center}
\end{figure}
We observe that they are in the $\DRAfp$ relation $\relnsp{LLLL-}$
with the dipole $C$ pointing towards the line dipole $A$ lies on. Because of
$A \rel{BFII} B$, dipole $B$ has to lie on the same line as $A$. But, since
$C$ is a straight line and lines $A$ and $B$ lie in front of $C$,
the endpoint of $B$ cannot lie behind $C$.
\end{pf}

As expected, the composition of $\DRAf$ turns out to be weak. Let
us have a closer look at the composition of $\DRAfp$ in the next
section.

\subsection{Strong Composition}
\label{sec:strong-comp}

We are now going to prove that $\DRAfp$ has a strong composition.
The following lemma will be crucial; note that it does \emph{not} hold
for $\DRAf$.
\begin{lem}\label{lemma:betweenness-preservation}
Let $R$ be a $\DRAfp$ base relation.
For $\DRAfp$ base relations $R$ not involving parallelism or anti-parallelism,
betweenness and equality among $\{{\bf s}_A,{\bf e}_A,S_{A,B}\}$\footnote{Please remember that
${\bf s}_A = st(dp_A)$ and ${\bf e}_A = ed(dp_A)$.} for given
dipoles $A\,R\,B$ are independent of the choice of $A$ and $B$, hence uniquely determined by $R$ alone.
\end{lem}
\begin{pf}
Let $R=r_1 r_2 r_3 r_4 r_5$, where $r_5\in\{+,-\}$ even if $r_5$ this is omitted in the
standard notation. Note that the assumption $r_5\in\{+,-\}$ implies that $S_{A,B}$ is defined.
If $r_3\in\{b,s,i,e,f\}$, ${\bf e}_A\not ={\bf s}_A=S_{A,B}$,
hence there is no betweenness. Analogously, ${\bf s}_A\not ={\bf e}_A=S_{A,B}$
if $r_4\in\{b,s,i,e,f\}$. The remaining possibilities for $r_3 r_4 r_5$ are:
\begin{enumerate}
 \item ll+, rr-: in these cases, ${\bf e}_A$ is between ${\bf s}_A$ and $S_{A,B}$;
 \item ll-, rr+: in these cases, ${\bf s}_A$ is between ${\bf e}_A$ and $S_{A,B}$;
 \item rl-, lr+: in these cases, $S_{A,B}$ is between ${\bf s}_A$ and ${\bf e}_A$.
\end{enumerate}
Note that cases 1 and 2 cannot be distinguished in $\DRAf$.
\begin{center}
\includegraphics[keepaspectratio, scale=0.85]{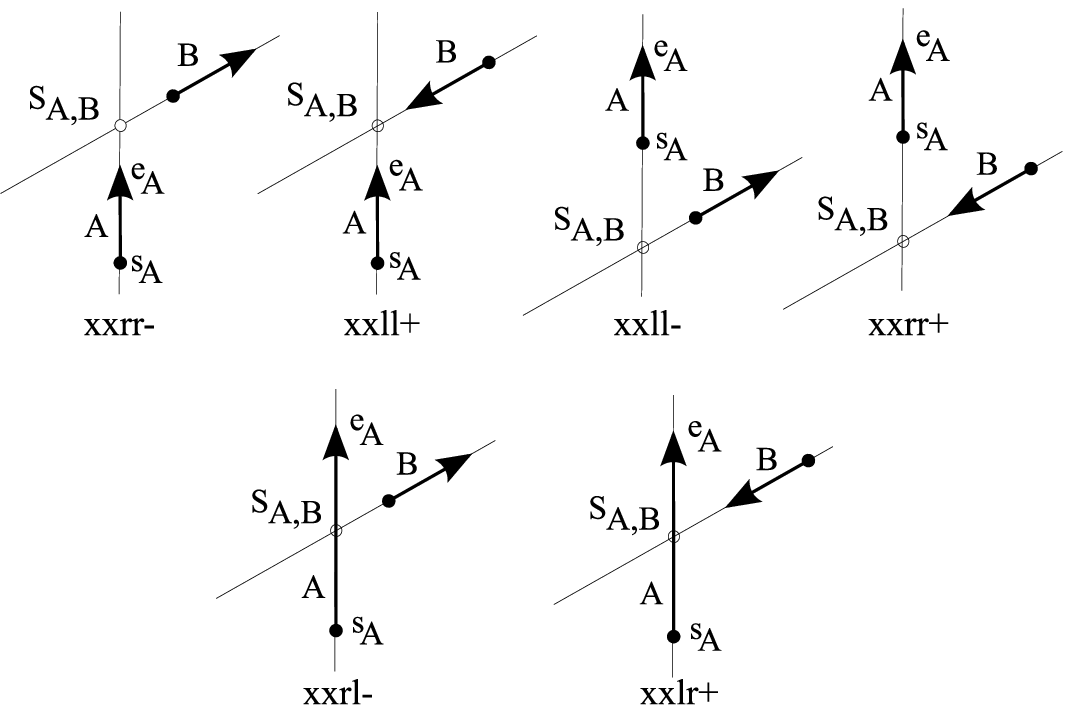}
\end{center}
\end{pf}

\begin{cor}\label{cor:betweenness-preservation}
Let $R$ be a $\DRAfp$ base relation not involving parallelism or anti-parallelism.
Let $A\,R\,B$ and $A'\,R\,B'$. Then, the map
$\{{\bf s}_A \mapsto {\bf s}_{A'}; {\bf e}_A\mapsto {\bf e}_{A'}; S_{A,B} \mapsto S_{A',B'}\}$ preserves betweenness
and equality.
\end{cor}

\begin{lem}\label{lemma:orientation-preservation}
Let $R$ be a $\DRAfp$ base relation not involving parallelism or anti-parallelism.
Given dipoles $A\,R\,C$ and $A'\,R\,C'$ and points $p_A$, $p_{A'}$, $p_C$ and $p_{C'}$
on the lines carrying $A$, $A'$, $C$ and $C'$ respectively, if the maps
$\{{\bf s}_A\mapsto {\bf s}_{A'}, {\bf e}_A\mapsto {\bf e}_{A'}, S_{A,C}\mapsto S_{A',C'}, p_A\mapsto p_{A'}\}$
and
$\{{\rm s}_C\mapsto {\bf s}_{C'}, {\bf e}_C\mapsto {\bf e}_{C'}, S_{A,C}\mapsto S_{A',C'}, p_C\mapsto p_{C'}\}$
preserve betweenness and equality, then the angles $\angle(\vect{S_{A,C}}{p_A},\vect{S_{A,C}}{p_C})$ and
$\angle(\vect{S_{A,C}}{p_A},\vect{S_{A,C}}{p_C})$ have the same sign.
\end{lem}
\begin{pf}
Since $A\,R\,C$ and $A'\,R\,C'$, the angles $\angle(\vect{{\bf s}_A}{{\bf e}_A},\vect{{\rm s}_C}{{\bf e}_C})$ and
$\angle(\vect{{\bf s}_{A'}}{{\bf e}_{A'}},\vect{{\bf s}_{C'}}{{\bf e}_{C'}})$ have the same sign.
By the assumption of the preservation of betweenness and equality, this carries over
to angles $\angle(\vect{S_{A,C}}{p_A},\vect{S_{A,C}}{p_C})$ and
$\angle(\vect{S_{A,C}}{p_A},\vect{S_{A,C}}{p_C})$.
\end{pf}

\begin{thm}
Composition in $\DRAfp$ is strong.
\end{thm}
\begin{pf}
Let $r_{ac} \in r_{ab} \circ r_{bc}$ be an entry in the composition table,
with $r_{ac}$, $r_{ab}$ and $r_{bc}$ base relations.
Given dipoles $A$ and $C$ with $A r_{ac} C$, we need to show the existence
of a dipole $B$ with  $A r_{ab} B$ and $B r_{bc} C$.

Since $r_{ac} \in r_{ab} \circ r_{bc}$, we know that there are dipoles
$A'$, $B'$ and $C'$ with $A' r_{ab} B'$, $B' r_{bc} C'$ and $A' r_{ac} C'$.
Given dipoles $X$ and $Y$, let $S_{X,Y}$ denote the point of intersection
of the lines carrying $X$ and $Y$; it is only defined if $X$ and $Y$
are not parallel. Consider now the three lines carrying $A'$,
$B'$ and $C'$, respectively.
According to the results of Section~\ref{sec:seven}, for the configuration of these
three lines, there are fifteen qualitatively different cases
1, 2, 3a, 3b, 3c, 4a, 4b, 4c, 5a, 5b, 5c, 6a, 6b, 6c and 7:
\begin{enumerate}
\item
The three points of intersection $S_{A',B'}$, $S_{B',C'}$ and $S_{A',C'}$
exist and are different. Since $A r_{ac} C$ and $A' r_{ac} C'$,
by Corollary~\ref{lemma:betweenness-preservation}, the point sets
$\{{\bf s}_A,{\bf e}_A,S_{A,C}\}$ and $\{{\bf s}_{A'},{\bf e}_{A'},S_{A',C'}\}$ are ordered in
corresponding ways on their lines.
Hence, it is possible to choose $S_{A,B}$ in such
a way that the point sets $\{{\bf s}_A,{\bf e}_A,S_{A,C},S_{A,B}\}$ and $\{{\bf s}_{A'},{\bf e}_{A'},S_{A',C'},S_{A',B'}\}$
are ordered in corresponding ways on their lines. In a similar way
(interchanging $A$ and $C$), $S_{B,C}$ can be chosen.
  \bigskip \begin{center}
    \includegraphics[keepaspectratio, scale=0.85]{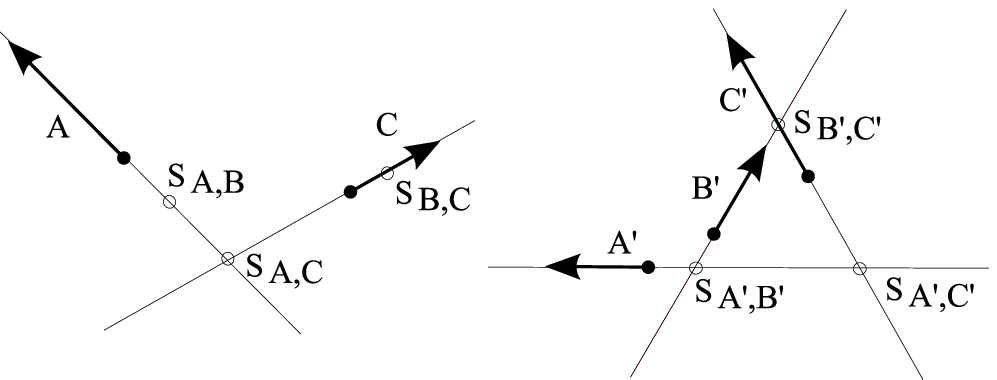}
  \end{center}  \bigskip
Since both $\{S_{A,B},S_{A,C},S_{B,C}\}$ and $\{S_{A',B'},S_{A',C'},S_{B',C'}\}$
are affine bases, there is a unique affine bijection $h\map{\R^2}{\R^2}$
with $h(S_{A',B'})=S_{A,B}$, $h(S_{A',C'})=S_{A,C}$ and $h(S_{B',C'})=S_{B,C}$.
By Lemma~\ref{lemma:orientation-preservation}, $h$ preserves orientation,
and thus by Thm.~\ref{thm:automorphisms} also the $\DRAfp$ relations.
Hence, by choosing $B=h(B')$, we get $h(A') r_{ab} B$ and $B r_{bc} h(C')$.
Since the sets $\{{\bf s}_A,{\bf e}_A,S_{A,C},S_{A,B}\}$ and
$\{h({\bf s}_{A'}),h({\bf e}_{A'}),S_{A,C},S_{A,B}\}$ are on the same line and have corresponding
qualitative (betweenness) relations, and the same holds for
the sets $\{{\bf s}_C,{\bf e}_C,S_{A,C},S_{B,C}\}$ and
$\{h({\bf s}_{C'}),h({\bf e}_{C'}),S_{A,C},S_{B,C}\}$, we also have
$A r_{ab} B$ and $B r_{bc} C$ (even though $h(A')=A$ and $h(C')=C$ do not necessarily hold).
\item
The three intersection points $S_{A',B'}$, $S_{B',C'}$ and $S_{A',C'}$
exist and coincide, i.e. $S_{A',B'}=S_{B',C'}=S_{A',C'}=:S'$.
Let $S=S_{A,C}$.
Let $x_A$ be ${\bf s}_A$ and $x_{A'}$ be ${\bf s}_{A'}$ if ${\bf s}_A\not= S$ (and therefore
${\bf s}_{A'}\not= S'$), otherwise, let $x_A$ be ${\bf e}_A$ and $x_{A'}$ be ${\bf e}_{A'}$.
$x_C$ and $x_{C'}$ are chosen in a similar way.
Since both $\{S,x_A,x_C\}$ and $\{S',x_{A'},x_{C'}\}$
are affine bases, there is a unique affine bijection $h\map{\R^2}{\R^2}$
with $h(S')=S$, $h(x_{A'})=x_A$ and $h(x_{C'})=x_C$.
The rest of the argument is similar to case (1).
\item
(Two lines are parallel and intersect with the third one.)
In the sequel, we will just specify how two affine bases are
chosen; the rest of the argument (as well as the choice of points on the unprimed
side in such a way that qualitative relations are preserved) is then similar to the previous cases.\\
Subcases (3a), (3b): The lines carrying $A$ and $C$ intersect. Choose $x_A$ and $x_{A'}$
as in case (2), and chose an appropriate point $S_{B,C}$. Then use the affine
bases $\{x_A,S_{A,C},S_{B,C}\}$ and $\{x_{A'},S_{A',C'},S_{B',C'}\}$.\\
Subcase (3c): The lines carrying $A$ and $C$ are parallel.
Choose appropriate points $S_{A,B}$ and $S_{B,C}$ and use the affine
bases $\{{\bf s}_A,S_{A,B},S_{B,C}\}$ and $\{{\bf s}_{A'},S_{A',B'},S_{B',C'}\}$.
\item
(Two lines are identical and intersect with the third one.)\\
Subcases (4a) and (4b): The lines carrying $A$ and $C$ intersect. Choose $x_A$, $x_{A'}$,
$x_C$ and $x_{C'}$ as in case (2) and use the affine bases
$\{S_{A,C},x_A,x_C\}$ and $\{S_{A',C'},x_{A'},x_{C'}\}$.\\
Subcase (4c): The lines carrying $A$ and $C$ are identical.
This means that $S_{A',B'}=S_{A',C'}=:S'$. Choose an appropriate
point $S$ and $x_A$, $x_{A'}$ as in case (2). Moreover, in a similar
way, choose $x_{B'}\not= S'$, and then some corresponding $x_B$
being in the same $\mathcal{LR}$-relation
to $A$ as $x_{B'}$ has to $A'$. Then use the affine bases
$\{S,x_A,x_B\}$ and $\{S,x_{A'},x_{B'}\}$.
\item
(All three lines are distinct and parallel.) Subcases (5a), (5b)
and (5c) can all be treated in the same way: Use the affine bases
$\{{\bf s}_A,{\bf e}_A,{\bf s}_C\}$ and $\{{\bf s}_{A'},{\bf e}_{A'},{\bf s}_{C'}\}$. Note that the
distance ratio does not matter here.
\item
(Two lines are identical and are parallel to the third one.)\\
Subcases (6a) and (6b): The lines carrying $A$ and $C$ are parallel.
Proceed as in case (5).\\
Subcase (6c): The lines carrying $A$ and $C$ are identical.
Choose some ${\bf s}_B$
in the same $\mathcal{LR}$-relation to $A$ as ${\bf s}_{B'}$ is to $A'$.
Then use the affine bases
$\{{\bf s}_A,{\bf e}_A,{\bf s}_B\}$ and $\{{\bf s}_{A'},{\bf e}_{A'},{\bf s}_{B'}\}$.
\item
(All three lines are identical.)
For this case, the result follows from the fact that Allen's interval algebra has strong composition (refer to \cite{RenzLigozat}).
\end{enumerate}
\end{pf}

\begin{cor}\label{cor:DRAoppstrong}
Composition in $\DRAopp$ is strong as well.
\end{cor}
\begin{pf}
By Example~\ref{exa:DRAfpDRAopp} and Prop.~\ref{prop:preserveStrength}.
\end{pf}

\section{Constraint Reasoning with the Dipole Calculus}
  \label{reasoning}

\subsection{Consistency\label{sec:consistency}}

We now consider the question of whether algebraic closure decides consistency.
We call the set of constraints between all dipoles at hand a
\emph{constraint network}. If no constraint between two dipoles is given, we
agree that they are in the universal relation. By
\emph{scenario}, we denote a constraint network in which all constraints are
base-relations\footnote{In this case, a base-relation between every pair of
distinct dipoles has to be provided}. We
construct constraint-networks which are geometrically unrealizable but still
algebraically closed. We do this by constructing constraint networks
that are consistent and algebraically closed, and then we will change a relation
in them in such a way that they remain algebraically closed but become inconsistent.
We
follow the approach of \cite{Rohrig1997} in using a simple geometric shape
for which scenarios exist, where algebraic closure fails to decide consistency.
In our case, the basic shape is a convex hexagon, similar to a screw head.

\begin{figure}[ht]
  \begin{center}
    \includegraphics[keepaspectratio, scale=0.85]{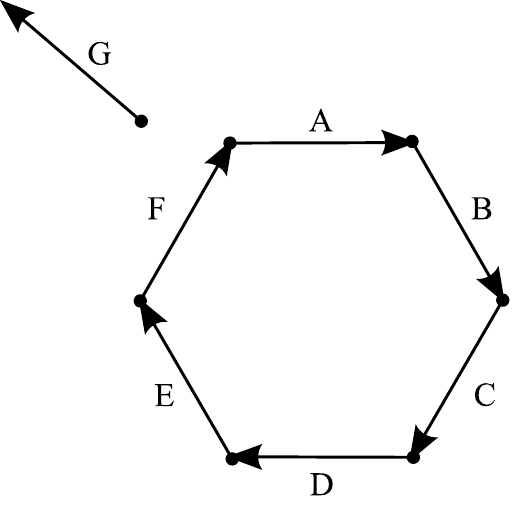}
    \caption{\label{fig:dra-c-ex}Convex hexagon}
  \end{center}
\end{figure}

Consider a convex hexagon consisting of the dipoles $A$, $B$, $C$, $D$, $E$ and
$F$. Such an object is described as
\[
(A \rel{errs} B) (B \rel{errs} C) (C \rel{errs} D) (D \rel{errs} E)
(E \rel{errs} F) (F \rel{errs} A)
\]
where the components $r$ of the relations ensure convexity, since they
enforce an angle between $0$ and $\pi$ between the respective first and
second dipole, i.e., the
endpoint of consecutive dipoles always lies to the right of the preceding
dipole. Such an
object is given in Fig.~\ref{fig:dra-c-ex}
To this scenario we add a seventh dipole $G$ with the relations
\[
(G \rel{rrll} A) (G \rel{lrll} F) (G \rel{llrr} D) (G \rel{rlrr} C)
\]
We have the overall constraint network:
\[
\begin{array}{l}
(A \rel{errs} B) (B \rel{errs} C) (C \rel{errs} D) (D \rel{errs} E)
(E \rel{errs} F) (F \rel{errs} A)
\\
(G \rel{rrll} A) (G \rel{lrll} F) (G \rel{llrr} D) (G \rel{rlrr} C)
\end{array}
\]

Because of the relations $(G \rel{lrll} F)$ and $(G \rel{rlrr} C)$,
line $l_G$ intersects line $l_F$ as well as line $l_C$. Because of
the first two components of the relations, dipoles $F$ and $C$ are
oriented into qualitatively antipodal directions. This network is consistent
and is of course algebraically closed.

\begin{figure}[ht]
  \begin{center}
    \includegraphics[keepaspectratio, scale=0.85]{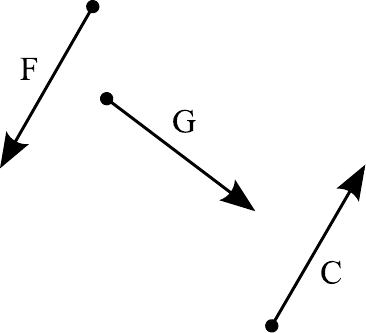}
    \caption{\label{fig:dra-c-ex2}Position of $G$}
  \end{center}
\end{figure}

\begin{figure}[ht]
  \begin{center}
    \includegraphics[keepaspectratio, scale=0.85]{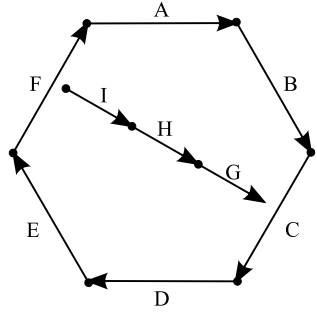}
    \caption{\label{fig:dra-fp}Construction of the counterexample}
  \end{center}
\end{figure}

To construct an inconsistent network,
we change the relation $(G \rel{rlrr} C)$ to $(G \rel{rlll} C)$ and obtain
the constraint network:
\[
\begin{array}{l}
(A \rel{errs} B) (B \rel{errs} C) (C \rel{errs} D) (D \rel{errs} E)
(E \rel{errs} F) (F \rel{errs} A)
\\
(G \rel{rrll} A) (G \rel{lrll} F) (G \rel{llrr} D) (G \rel{rlll} C)
\end{array}
\]
The relations
$(G \rel{rlll} C)$ and $(G \rel{lrll} F)$ enforce that $G$ must lie in
between $F$ and $C$ as shown in Fig.~\ref{fig:dra-c-ex2}.
In this case, the all convex hexagons
$A$, $B$, $C$, $D$, $E$, $F$ have the endpoints of
consecutive dipoles lying to the left of the preceding one, they are of
the form:
\[
(A \rel{ells} B) (B \rel{ells} C) (C \rel{ells} D) (D \rel{ells} E)
(E \rel{ells} F) (F \rel{ells} A)
\]
which is a contradiction of the required form of hexagon in the
scenario. In fact there is no affine transformation which preserves
the relative orientations between dipoles $A$, $B$, $C$, $D$, $E$, $F$,
and maps a hexagon of Fig.~\ref{fig:dra-c-ex} to any that can be constructed
along dipoles $C$ and $F$ in Fig.~\ref{fig:dra-c-ex2}
in such a way that the edges $C$ and $F$ of both hexagons coincide.
Still algebraic closure with
$\DRAf$ yields the refinement:
\[
\begin{array}{l}
(F \rel{lllr} G)(E\,\,(\relnsp{flll}, \relnsp{llll}, \relnsp{rfll}, \relnsp{rlll}, \relnsp{rrll})\,\,G)(D \rel{errs} E)
(D \rel{rrll} G)\\
(D\,\,(\relnsp{rbrr}, \relnsp{rllr}, \relnsp{rlrr}, \relnsp{rrrr})\,\,F)
(C \rel{llrl} G)(C\,\,(\relnsp{lrrl}, \relnsp{rllr})\,\,F)(E \rel{errs} F)\\
(C\,\,(\relnsp{rllr}, \relnsp{rrfr}, \relnsp{rrlr}, \relnsp{rrrr})\,\,E)(C\,\,\rel{errs}\,\,D)
(B\,\,(\relnsp{llrr}, \relnsp{rrrr})\,\,G)\\
(B\,\,(\relnsp{blrr}, \relnsp{llll}, \relnsp{llrf}, \relnsp{llrl}, \relnsp{llrr}, \relnsp{rfll}, \relnsp{rlll}, \relnsp{rlrr,
rrbl}, \relnsp{rrll}, \relnsp{rrrl}, \relnsp{rrrr})\,\,E)\\
(B\,\,(\relnsp{rbrr}, \relnsp{rlrr}, \relnsp{rrfr}, \relnsp{rrlr}, \relnsp{rrrr})\,\,D)
(B \rel{errs} C)(A \rel{llrr} G)(A \rel{rser} F)\\
(A\,\,(\relnsp{frrr}, \relnsp{lrrr}, \relnsp{rrrb}, \relnsp{rrrl}, \relnsp{rrrr)\,\,E)(A\,\,(rllr}, \relnsp{rlrr}, \relnsp{rrrr})\,\,C)\\
(A\,\,(\relnsp{lfrr}, \relnsp{llbr}, \relnsp{llll}, \relnsp{lllr}, \relnsp{llrr}, \relnsp{lrll}, \relnsp{lrrr}, \relnsp{rrlf}, \relnsp{rrll}, \relnsp{rrlr}, \relnsp{rrrr})\,\,D)\\
(B\,\,(\relnsp{frrr}, \relnsp{lrrl}, \relnsp{lrrr}, \relnsp{rrrr})\,\,F)(A \rel{errs} B)
\end{array}
\]
A scenario,
\[
 \begin{array}{l}
(F \rel{lllr} G)(E \rel{flll} G)(E \rel{errs} F)(D \rel{rrll} G)(D \rel{rrrr} F)(D \rel{errs} E)\\
(C \rel{llrl} G)(C \rel{rllr} F)(C \rel{rrlr} E)(C \rel{errs} D)(B \rel{llrr} G)(B \rel{lrrl} F)\\
(B \rel{llrl} E)(B \rel{rlrr} D)(B \rel{errs} C)(A \rel{llrr} G)(A \rel{rser} F)(A \rel{rrrr} E)\\
(A \rel{lrrr} D)(A \rel{rllr} C)(A \rel{errs} B)
\end{array}
\]
can be derived from this algebraically closed network. It is still deemed
algebraically closed, even though it is not consistent with the same argument
given above. Hence algebraic closure does not decide consistency for
$\DRAf$-scenarios.
On the other hand, algebraic closure with  $\DRAfp$ detects all
possible extensions of this network to that calculus as being inconsistent.
Extending the consistent case with the relation $(G \rel{rlrr} C)$ yields
three possible extensions for $\DRAfp$ scenarios, of which all are
consistent. In fact, we get the three following consistent refinements.
\[
\begin{array}{c|c|c|c}
DRA_f\textnormal{-relation} &
~~~\textnormal{refinement} 1 ~~~&
~~~\textnormal{refinement} 2 ~~~&
~~~\textnormal{refinement} 3~~~
\\\hline
(G \rel{rrll} A)&(G \rel{rrll-} A)&(G \rel{rrll-} A)&(G \rel{rrll-} A)
\\\hline
(G \rel{llrr} D)&(G \rel{llrr-} D)&(G \rel{llrr+} D)&(G \rel{llrrP} D)
\end{array}
\]
We have found an example that shows that algebraic closure for
$\DRAfp$ finds inconsistencies in constraint networks where
it fails for $\DRAf$. Does algebraic closure for
$\DRAfp$ decide consistency? We can also give a negative
result for this. To construct a counterexample, we begin with a
configuration as in Fig.~\ref{fig:dra-fp}
We ensure with the constraints
\[
(A \rel{errs} B) (B \rel{errs} C) (C \rel{errs} D) (D \rel{errs} E)
(E \rel{errs} F) (F \rel{errs} A)
\]
that the dipoles $A$, $B$, $C$, $D$, $E$ and $F$ form a convex hexagon.
Furthermore, we ensure that the dipoles $I$, $H$ and $G$ form a continuous line
by
\[
(I \rel{efbs} H)(H \rel{efbs} G).
\]
The constraints
\[
(F  \rel{rrrl} I)(C \rel{rrlr} G)
\]
state that the line has to lie inside the hexagon, since its start point
and end point lie inside. To construct the counterexample, we just
claim that the end point of $H$ lies outside the hexagon by
$(A \rel{rllr} H)$, i.e. the lines $A$ and $H$ intersect, this is a
contradiction of the convexity of the hexagon. This network
can be refined to a scenario
\begin{eqnarray*}
  \mathtt{SCEN} &:=&(H \rel{efbs} G)(I \rel{ffbb} G)(I \rel{efbs} H)(F \rel{rrrl} G)(F \rel{rrrl} H)\\ &&
(F \rel{rrrl} I)(E \rel{rrrr-} G)(E \rel{rrrr-} H)(E \rel{rrrr-} I)\\ &&
(E \rel{errs} F)(D \rel{rrrrA} G)(D \rel{rrrrA} H)(D \rel{rrrrA} I)\\ &&
(D \rel{rrrr-} F)(D \rel{errs} E)(C \rel{rrlr} G) (C \rel{rrlr} H) \\ &&
(C \rel{rrlr} I)(C \rel{rrlr} F)(C \rel{rrrr+} E) (C \rel{errs} D) \\ &&
(B \rel{rrrl} G)(B \rel{rrrl} H)(B \rel{rrrl} I) (B \rel{lrrl} F)  \\ &&
(B \rel{llrr-} E)(B \rel{rlrr} D)(B \rel{errs} C) (A \rel{lllr} G) \\ &&
(A \rel{rllr} H)(A \rel{rrlr} I)(F \rel{errs} A) (A \rel{rrrrA} E) \\ &&
(A \rel{lrrr} D)(A \rel{rlrr} C)(A \rel{errs} B).
\end{eqnarray*}
which is still algebraically closed w.r.t. $\DRAfp$, even though
it is not consistent. We see that algebraic-closure does not decide
consistency even for $\DRAfp$-scenarios.

We have run several tests to get some quantitative information on how much better
the $\DRAfp$ calculus performs with respect to the $\DRAf$
calculus. We have generated several scenarios of size $\le n$ with
$n \in \left\{30,40,50,60,70\right\}$ randomly to obtain this information.
It turns out that a number of $10^{\frac{n}{10}+1}$ scenarios yield usable
data. In fact, we have generated $\DRAfp$ scenarios and checked them
with an algebraic reasoner, then we have projected them to $\DRAf$
and checked these with the same reasoner. In the end, we compared the
per-scenario results. The results are as follows:
{\tiny
\begin{center}
\begin{tabular}{l|l|l|l|l|l}
Scenarios                              & $10000$ & $100000$ &$1000000$ & $10000000$ & $100000000$ \\\hline
Maximum Size                           & $30$ & $40$ & $50$ & $60$ & $70$ \\\hline
Algebraically Closed                   & $691$ & $5295$ & $40820$ & $346164$ & $3048063$ \\\hline
A-closed w.r.t. $\DRAf$ only  & $11$ & $149$ & $1061$ & $8839$ & $78792$\\\hline
A-closed w.r.t. $\DRAfp$ only & $0$ & $0$ & $0$ & $0$ & $0$
\end{tabular}
\end{center}}
Roughly $2.5\%$ of the scenarios that are algebraically
closed w.r.t. to $\DRAf$ are not algebraically closed w.r.t.
$\DRAfp$. Still, for the smallest checked maximum scenario size $30$
the factor is only $1.5\%$.

We also investigate the question if algebraic closure decides consistency
for $\DRAop$ and $\DRAopp$.

\begin{figure}[htb]
\begin{center}
\includegraphics[keepaspectratio, scale=0.85]{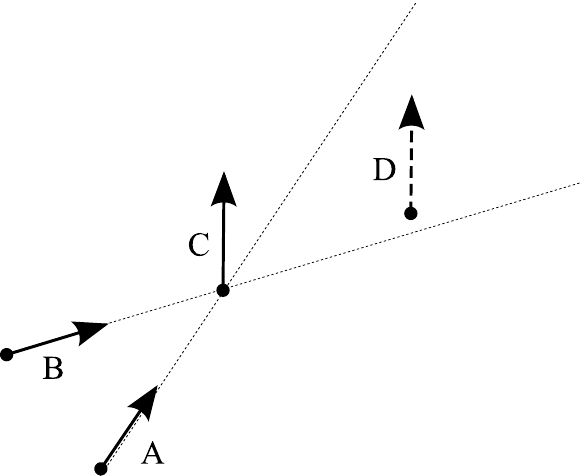}
\caption{\label{fig:draopcons} $\DRAopp$ scenario}
\end{center}
\end{figure}

\begin{prop}\label{prop:DRAOppAclosurevsconsistency}
  For $\DRAopp$ algebraic closure does not decide consistency.
\end{prop}
\begin{pf}
  This proof is inspired by the one that shows that algebraic
  closure does not decide consistency for $\mathcal{OPRA}$ (ref. to \cite{Frommberger2007}).
  Consider a $\DRAopp$ constraint network in three points $A$, $B$ and
  $C$ as shown in Fig.~\ref{fig:draopcons}. Both $A$ and $B$
  point at $C$. These three points are in the relations:
  \[
  A \rel{LEFTright-} B \quad\quad
  A \rel{FRONTleft} C \quad\quad
  B \rel{FRONTleft} C.
  \]
  We add a point $D$ to our constraint satisfaction problem with
  $C \rel{RIGHTleftP} D$.  We claim that $D$ also lies
  in front of $A$ and $B$ by introducing the constraints
  $A \rel{FRONTleft} D$ and $B \rel{FRONTleft} D$. By
  inspecting the composition table of $\DRAopp$, we can
  see that it is consistent. Since by the constraint
  $A \rel{LEFTright-} B$ the points $A$ and $B$ are not
  collinear,
  $D$ has to lie on the intersection point of
  the rays $l_A$ and $l_B$, but by $A \rel{FRONTleft} C$
  and $B \rel{FRONTleft} C$, $C$ also has to lie on that
  intersection point. Hence, $C$ and $D$ have to have
  the same position, what is a contradiction to the constraint
  $C \rel{RIGHTleftP} D$. Hence this scenario is algebraically
  closed, but inconsistent.
\end{pf}

\begin{prop}
  For $\DRAop$ algebraic closure does not decide consistency.
\end{prop}
\begin{pf}
  This proof is analogous to the one of Prop.~\ref{prop:DRAOppAclosurevsconsistency},
  with substituting $\relnsp{LEFTright-}$ by $\relnsp{LEFTright}$ and
  $\relnsp{RIGHTleftP}$ by $\relnsp{RIGHTleft}$.
\end{pf}

\section{A Sample Application of the Dipole Calculus \label{application}}

\begin{figure}[t]
\begin{center}
\includegraphics[keepaspectratio, scale=0.85]{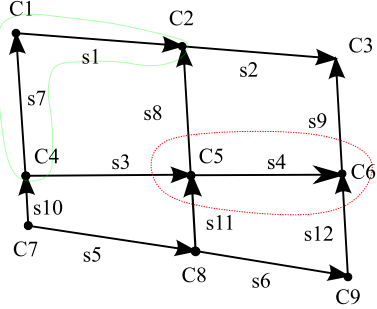}
\caption{\label{streetsDRAfp} A street network and two local observations}
\end{center}
\end{figure}

In this section, we want to demonstrate with an example how spatial knowledge expressed in $\DRAfp$
can be used for deductive reasoning based on constraint propagation (algebraic closure), resulting
in the generation of useful indirect knowledge from partial observations in a spatial scenario.
In our sample application, a spatial agent (a simulated robot, cognitive simulation of a
biological system etc.) explores a spatial scenario.
The agent collects local observations and wants to generate survey knowledge.

Fig.~\ref{streetsDRAfp} shows our spatial environment. It consists of
a street network in which some streets continue straight after a crossing and
some streets run parallel. These features are typical of real-world street networks.
Spatial reasoning in our example uses constraint propagation (e.g. algebraic closure computation)
to derive indirect constraints between the relative location of streets which are
further apart from local observations between neighboring streets.
The resulting survey knowledge can be used for several tasks including navigation tasks.

The environment is represented as
streets $s_i$ and crossings $C_j$.
The streets and crossings have unique names
(e.g. $s_1$, ... ,  $s_{12}$, and $C_1$, ..., $C_9$ in one concrete example).
The local observations are modeled in the following way, based on specific visibility rules
(we want to simulate prototypical features of visual perception):
Both at each crossing and at each straight street segment we have an observation.
At each crossing the agent observes the neighboring crossings.
At the middle of each straight street segment the agent can observe
the direction of the outgoing streets at the adjacent crossings (but not at their other ends).
Two specific examples of observations are marked in Fig.~\ref{streetsDRAfp}.
The observation "s1 errs s7" is marked green at crossing C1.
The observation "s8 rrllP s9" is marked red at street s4.

These observations relate spatially neighboring streets to each other in a pairwise manner, using
 $\DRAfp$ base relations.
The agent has no additional knowledge about the specific environment.
The spatial world knowledge of the agent is expressed in the converse and composition tables of
$\DRAfp$ .

The following sequence of partial observations could be the result of a tour made by the spatial agent,
exploring the street network of our example (see Fig.~\ref{streetsDRAfp}):
\begin{figure}[ht]
  \begin{center}
    \includegraphics[keepaspectratio, scale=0.85]{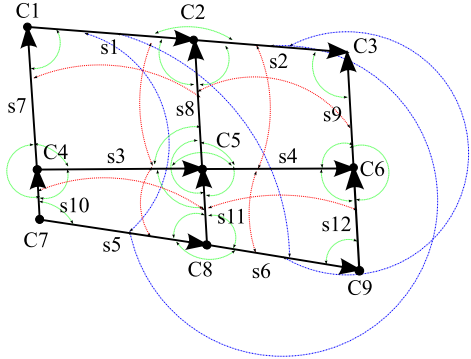}
    \caption{\label{streetsDRAfpInput}All observation and resulting uncertainty}
  \end{center}
\end{figure}
{\tt \small
\\
\begin{tabbing}
s11:\= \kill \\
Observations at crossings\\
C1: \> (s7 \rel{errs} s1)\\
C2: \> (s1 \rel{efbs} s2) (s8 \rel{errs} s2) (s1 \rel{rele} s8)\\
C3: \> (s2 \rel{rele} s9)\\
C4: \> (s10 \rel{efbs} s7) (s10 \rel{errs} s3) (s7 \rel{srsl} s3)\\
C5: \> (s3 \rel{efbs} s4) (s11 \rel{efbs} s8) (s11 \rel{errs} s4) (s3 \rel{ells} s8)\\
    \> (s3 \rel{rele} s11) (s8 \rel{srsl} s4)\\
C6: \> (s12 \rel{efbs} s9) (s4 \rel{ells} s9) (s4 \rel{rele} s12)\\
C7: \> (s10 \rel{srsl} s5)\\
C8: \> (s5 \rel{efbs} s6) (s5 \rel{ells} s11) (s11 \rel{srsl} s6)\\
C9: \> (s6 \rel{ells} s12)\\
\\
Observations at streets\\
s1: \> (s7 \rel{rrllP} s8)\\
s2: \> (s8 \rel{rrllP} s9)\\
s3: \> (s10 \rel{rrllP} s11)\\
s4: \> (s11 \rel{rrllP} s12)\\
s8: \> (s3 \rel{llrr-} s1)\\
s9: \> (s4 \rel{llrr-} s2)\\
s10:\> (s3 \rel{rrll-} s5)\\
s11:\> (s4 \rel{rrll-} s6)\\
\end{tabbing}
}

The result of the algebraic closure computation/constraint propagation is a refined network with
the same solution set (the results are computed with the publicly available SparQ reasoning tool
supplied with our newly computed $\DRAfp$ composition table \cite{SparQmanual}).
We have listed the results in the appendix.
Three different models are the only remaining consistent interpretations
(see the appendix for a list of all the resulting data).
The three different models agree on all but four relations.
The solution set can be explained with the help of the diagram in Fig.~\ref{streetsDRAfpInput}.
The input crossing observations are marked with green arrows, the input street observations are
marked with red arrows.
The result shows that for all street pairs which could not be observed directly,
the algebraic closure algorithm deduces a strong constraint/precise information.
Typically, the resulting spatial relation between street
pairs comprises just one $\DRAfp$ base relation.
The exception consists of four relations between streets in which the three models differ
(marked with dashed blue arrows in Fig.~\ref{streetsDRAfpInput}).
For these four relations each model from the solution set agrees on the same $\DRAf$ base relation for
a given relation, but
the three consistent models differ on the finer granularity level of $\DRAfp$ base relations.
Since the refinement of one of these four underspecified relations on a single interpretation
($\DRAfp$ base relation) as a logical consequence
also assigns a single base relation to the other three relations, only three interpretations are valid models.
The uncertainty/indeterminacy is the result of the specific street configuration
in our example. The streets in a North-South direction are parallel, but the streets in an East-West direction
are not parallel resulting in fewer constraint composition results.
However, the small solution set of consistent models agrees on most of the relative position
relations between street pairs and the differences between models are small.
In our judgement, this means that the system has generated
the relevant survey knowledge about the whole street network
from local observations alone.

\section{Summary and Conclusion}
We have presented different variants of qualitative spatial reasoning calculi about oriented
straight line segments which we call dipoles.
We have derived calculi for oriented points from dipole calculi, which turned out to be
isomorphic to some versions of the $\mathcal{OPRA}$ calculi.
These spatial calculi provide a basis for
representing and reasoning about
qualitative position information in intrinsic reference systems.

We have computed the composition table for dipole calculi
by a new method based on the algebraic semantics
of the dipole relations. We have used a so-called condensed semantics
which uses the orbits of the affine group $\mathbf{GA}(\R^2)$ to provide
an abstract symbolic notion of qualitative composition configuration.
This can be used to compute the composition table in a computer-assisted way.
The correctness of this computation is ensured by letting the computer
program directly operate with qualitative composition configurations.

This has been the first computation of the composition table for
$\DRAfp$.  So far, only composition tables for $\DRAc$ and $\DRAf$
exist, which contain many errors \cite{ThomasSollerDA2005}.  We have
analysed the algebraic features of the various dipole calculi.  We
have proved the result that $\DRAfp$ has strong composition.  This is
an interesting result, because in this case an application-motivated
calculus extension has been found to also have a certain mathematical
elegance.  Moreover, the strength of composition carries over to
$\DRAopp$, the $\OPRA$ variant introduced in this paper.  This
transfer of properties from one calculus to another calculus is an
important new general result on quotients of qualitative calculi. To
our knowledge, also the notion of quotient of a qualitative calculus
(defined using methods from universal algebra) appears for the first
time in this paper.

We have demonstrated a prototypical application of reasoning about
qualitative position information in relative reference
systems.  In this scenario about cognitive spatial agents and
qualitative map building, coarse locally perceived street
configuration information has to be integrated by constraint
propagation in order to get survey knowledge.  The well-known path-consistency
method which is implemented with standard QSR tools can make use of our
new dipole calculus composition table and compute the desired result
in polynomial time.  Such concrete but generalizable application
scenarios for relative position calculi are the
more important since a recent result by Wolter and Lee
\cite{LeeWolterAIJ} showed that relative position calculi are intractable
even in base relations.  For this reason, it is necessary to gain
experience in which application contexts the unavoidable approximate
reasoning is effective and produces relevant inference results. With
our street network example, we have a test case which puts emphasis
on deriving implicit knowledge as the output of qualitative spatial
reasoning based on observed data.  This is a prototypical application
scenario which in the future can also be applied to other relative
position calculi.

Since the observed data in the case
of error-free perception leads to consistent input constraints, the
general consistency problem can be avoided --
we instead rely on logical consequence. Now both problems
are intractable and need to be approximated using algebraic closure;
however, in our setting, approximate losses are less harmful,
since we do not risk working with inconsistent scenarios.

Our future work will address the question of how in general the quality of
approximations for relative position reasoning can also be assessed
with quantitative measures.  Another part of our future QSR research
will apply our new condensed semantics method to other calculi.

\section*{Acknowledgment}
The authors would like to thank
Diedrich Wolter,
Jochen Renz,
Frank Dylla,
Christian Freksa,
Franz Kalhoff,
Stefan W{\"o}lfl,
Lutz Schr\"oder,
and
Brandon Bennett
for interesting and helpful
discussions related to the topic of the paper.
Our work was supported by the DFG Transregional Collaborative
Research Center SFB/TR~8 {}``Spatial Cognition''.

\bibliographystyle{elsarticle-num}
\bibliography{oslsa}

\begin{appendix}

\section*{Appendix: Computation for the street network application with the
SparQ tool}

In this appendix, we demonstrate how to use the publicly available
SparQ QSR toolbox \cite{SparQmanual} to compute
the algebraic closure by constraint propagation
for the street network example from Section~\ref{application}.
For successful relative position reasoning, the SparQ tool has to be
supplied with our newly computed $\DRAfp$ composition table \cite{SparQmanual}.

The local street configuration observations by the spatial agent are listed in Section~\ref{application}.
The direct translation of these logical propositions into a SparQ spatial reasoning command
looks as follows\footnote{For technical details of SparQ we refer the reader
to the SparQ manual \cite{SparQmanual}}:

{\tt
sparq constraint-reasoning dra-fp path-consistency
"(  (s7 errs s1) (s1 efbs s2) (s8 errs s2) (s1 rele s8) (s2 rele s9) (s10 efbs s7)
(s10 errs s3) (s7 srsl s3) (s3 efbs s4) (s11 efbs s8) (s11 errs s4) (s3 ells s8)
(s3 rele s11) (s8 srsl s4) (s12 efbs s9) (s4 ells s9) (s4 rele s12) (s10 srsl s5)
(s5 efbs s6) (s5 ells s11) (s11 srsl s6) (s6 ells s12) (s7 rrllP s8) (s8 rrllP s9)
(s10 rrllP s11) (s11 rrllP s12) (s3 llrr- s1) (s4 llrr- s2) (s3 rrll- s5) (s4 rrll- s6) )"

}
\footnote{SparQ refers to $\DRAfp$ with the symbol dra-80.
SparQ does not accept line breaks which we have inserted here for better readability.
All the data for this sample application including the new composition table can be obtained from
the URL \url{http://www.informatik.uni-bremen.de/~till/Oslsa.tar.gz} (which also provides the composition table and other data for the GQR reasoning tool
{\tt
https://sfbtr8.informatik.uni-freiburg.de/R4LogoSpace/Resources/}).}

The result of this reasoning command is a refined network with
the same solution set derived by the application of the algebraic closure/constraint propagation algorithm
(see Section~\ref{ConstraintPropagationEtc}).

{\tt
Modified network. \\
((S5 (\relnsp{EFBS}) S6)(S12 (\relnsp{LSEL}) S6)(S12 (\relnsp{LLFL}) S5)(S11 (\relnsp{SRSL}) S6)(S11 (\relnsp{LSEL}) S5)(S11 (\relnsp{RRLLP}) S12)
(S4 (\relnsp{RRLL-}) S6)(S4 (\relnsp{RRLL-}) S5)(S4 (\relnsp{RELE}) S12)(S4 (\relnsp{RSER}) S11)(S3 (\relnsp{RRLL-}) S6)(S3 (\relnsp{RRLL-}) S5)
(S3 (\relnsp{RFLL}) S12)(S3 (\relnsp{RELE}) S11)(S3 (\relnsp{EFBS}) S4)(S10 (\relnsp{RRBL}) S6)(S10 (\relnsp{SRSL}) S5)(S10 (\relnsp{RRLLP}) S12)
(S10 (\relnsp{RRLLP}) S11)(S10 (\relnsp{RRRB}) S4)(S10 (\relnsp{ERRS}) S3)(S9 (\relnsp{LBLL}) S6)(S9 (\relnsp{LLLL-}) S5)(S9 (\relnsp{BSEF}) S12)
(S9 (\relnsp{LLRRP}) S11)(S9 (\relnsp{LSEL}) S4)(S9 (\relnsp{LLFL}) S3)(S9 (\relnsp{LLRRP}) S10)(S8 (\relnsp{BRLL}) S6)(S8 (\relnsp{LBLL}) S5)
(S8 (\relnsp{RRLLP}) S12)(S8 (\relnsp{BSEF}) S11)(S8 (\relnsp{SRSL}) S4)(S8 (\relnsp{LSEL}) S3)(S8 (\relnsp{LLRRP}) S10)(S8 (\relnsp{RRLLP}) S9)
(S2 (\relnsp{RRLL+ RRLL- RRLLP}) S6)(S2 (\relnsp{RRLL+ RRLL- RRLLP}) S5)(S2 (\relnsp{RRLF}) S12)(S2 (\relnsp{RRFR}) S11)(S2 (\relnsp{RRLL+}) S4)
(S2 (\relnsp{RRLL+}) S3)(S2 (\relnsp{RRRR+}) S10)(S2 (\relnsp{RELE}) S9)(S2 (\relnsp{RSER}) S8)(S1 (\relnsp{RRLL+ RRLL- RRLLP}) S6)
(S1 (\relnsp{RRLL+ RRLL- RRLLP}) S5)(S1 (\relnsp{RRLL+}) S12)(S1 (\relnsp{RRLF}) S11)(S1 (\relnsp{RRLL+}) S4)(S1 (\relnsp{RRLL+}) S3)
(S1 (\relnsp{RRFR}) S10)(S1 (\relnsp{RFLL}) S9)(S1 (\relnsp{RELE}) S8)(S1 (\relnsp{EFBS}) S2)(S7 (\relnsp{RRLL-}) S6)(S7 (\relnsp{BRLL}) S5)
(S7 (\relnsp{RRLLP}) S12)(S7 (\relnsp{RRLLP}) S11)(S7 (\relnsp{RRBL}) S4)(S7 (\relnsp{SRSL}) S3)(S7 (\relnsp{BSEF}) S10)(S7 (\relnsp{RRLLP}) S9)
(S7 (\relnsp{RRLLP}) S8)(S7 (\relnsp{RRRB}) S2)(S7 (\relnsp{ERRS}) S1))
}

SparQ can output all path-consistent scenarios (i.e. constraint networks in base relations) via the command:

{\tt
sparq constraint-reasoning dra-fp scenario-consistency all
"(  (s7 errs s1) (s1 efbs s2) (s8 errs s2) (s1 rele s8) (s2 rele s9) (s10 efbs s7)
(s10 errs s3) (s7 srsl s3) (s3 efbs s4) (s11 efbs s8) (s11 errs s4) (s3 ells s8)
(s3 rele s11) (s8 srsl s4) (s12 efbs s9) (s4 ells s9) (s4 rele s12) (s10 srsl s5)
(s5 efbs s6) (s5 ells s11) (s11 srsl s6) (s6 ells s12) (s7 rrllP s8) (s8 rrllP s9)
(s10 rrllP s11) (s11 rrllP s12) (s3 llrr- s1) (s4 llrr- s2) (s3 rrll- s5) (s4 rrll- s6) )"

}

For this CSP, only three slightly different path consistent scenarios exist:

{\tt
((S5 (\relnsp{EFBS}) S6)(S12 (\relnsp{LSEL}) S6)(S12 (\relnsp{LLFL}) S5)(S11 (\relnsp{SRSL}) S6)(S11 (\relnsp{LSEL}) S5)(S11 (\relnsp{RRLLP}) S12)
(S4 (\relnsp{RRLL-}) S6)(S4 (\relnsp{RRLL-}) S5)(S4 (\relnsp{RELE}) S12)(S4 (\relnsp{RSER}) S11)(S3 (\relnsp{RRLL-}) S6)(S3 (\relnsp{RRLL-}) S5)
(S3 (\relnsp{RFLL}) S12)(S3 (\relnsp{RELE}) S11)(S3 (\relnsp{EFBS}) S4)(S10 (\relnsp{RRBL}) S6)(S10 (\relnsp{SRSL}) S5)(S10 (\relnsp{RRLLP}) S12)
(S10 (\relnsp{RRLLP}) S11)(S10 (\relnsp{RRRB}) S4)(S10 (\relnsp{ERRS}) S3)(S9 (\relnsp{LBLL}) S6)(S9 (\relnsp{LLLL-}) S5)(S9 (\relnsp{BSEF}) S12)
(S9 (\relnsp{LLRRP}) S11)(S9 (\relnsp{LSEL}) S4)(S9 (\relnsp{LLFL}) S3)(S9 (\relnsp{LLRRP}) S10)(S8 (\relnsp{BRLL}) S6)(S8 (\relnsp{LBLL}) S5)
(S8 (\relnsp{RRLLP}) S12)(S8 (\relnsp{BSEF}) S11)(S8 (\relnsp{SRSL}) S4)(S8 (\relnsp{LSEL}) S3)(S8 (\relnsp{LLRRP}) S10)(S8 (\relnsp{RRLLP}) S9)
(S2 (\relnsp{RRLLP}) S6)(S2 (\relnsp{RRLLP}) S5)(S2 (\relnsp{RRLF}) S12)(S2 (\relnsp{RRFR}) S11)(S2 (\relnsp{RRLL+}) S4)(S2 (\relnsp{RRLL+}) S3)
(S2 (\relnsp{RRRR+}) S10)(S2 (\relnsp{RELE}) S9)(S2 (\relnsp{RSER}) S8)(S1 (\relnsp{RRLLP}) S6)(S1 (\relnsp{RRLLP}) S5)(S1 (\relnsp{RRLL+}) S12)
(S1 (\relnsp{RRLF}) S11)(S1 (\relnsp{RRLL+}) S4)(S1 (\relnsp{RRLL+}) S3)(S1 (\relnsp{RRFR}) S10)(S1 (\relnsp{RFLL}) S9)(S1 (\relnsp{RELE}) S8)
(S1 (\relnsp{EFBS}) S2)(S7 (\relnsp{RRLL-}) S6)(S7 (\relnsp{BRLL}) S5)(S7 (\relnsp{RRLLP}) S12)(S7 (\relnsp{RRLLP}) S11)(S7 (\relnsp{RRBL}) S4)
(S7 (\relnsp{SRSL}) S3)(S7 (\relnsp{BSEF}) S10)(S7 (\relnsp{RRLLP}) S9)(S7 (\relnsp{RRLLP}) S8)(S7 (\relnsp{RRRB}) S2)(S7 (\relnsp{ERRS}) S1))\\\\
((S5 (\relnsp{EFBS}) S6)(S12 (\relnsp{LSEL}) S6)(S12 (\relnsp{LLFL}) S5)(S11 (\relnsp{SRSL}) S6)(S11 (\relnsp{LSEL}) S5)(S11 (\relnsp{RRLLP}) S12)
(S4 (\relnsp{RRLL-}) S6)(S4 (\relnsp{RRLL-}) S5)(S4 (\relnsp{RELE}) S12)(S4 (\relnsp{RSER}) S11)(S3 (\relnsp{RRLL-}) S6)(S3 (\relnsp{RRLL-}) S5)
(S3 (\relnsp{RFLL}) S12)(S3 (\relnsp{RELE}) S11)(S3 (\relnsp{EFBS}) S4)(S10 (\relnsp{RRBL}) S6)(S10 (\relnsp{SRSL}) S5)(S10 (\relnsp{RRLLP}) S12)
(S10 (\relnsp{RRLLP}) S11)(S10 (\relnsp{RRRB}) S4)(S10 (\relnsp{ERRS}) S3)(S9 (\relnsp{LBLL}) S6)(S9 (\relnsp{LLLL-}) S5)(S9 (\relnsp{BSEF}) S12)
(S9 (\relnsp{LLRRP}) S11)(S9 (\relnsp{LSEL}) S4)(S9 (\relnsp{LLFL}) S3)(S9 (\relnsp{LLRRP}) S10)(S8 (\relnsp{BRLL}) S6)(S8 (\relnsp{LBLL}) S5)
(S8 (\relnsp{RRLLP}) S12)(S8 (\relnsp{BSEF}) S11)(S8 (\relnsp{SRSL}) S4)(S8 (\relnsp{LSEL}) S3)(S8 (\relnsp{LLRRP}) S10)(S8 (\relnsp{RRLLP}) S9)
(S2 (\relnsp{RRLL-}) S6)(S2 (\relnsp{RRLL-}) S5)(S2 (\relnsp{RRLF}) S12)(S2 (\relnsp{RRFR}) S11)(S2 (\relnsp{RRLL+}) S4)(S2 (\relnsp{RRLL+}) S3)
(S2 (\relnsp{RRRR+}) S10)(S2 (\relnsp{RELE}) S9)(S2 (\relnsp{RSER}) S8)(S1 (\relnsp{RRLL-}) S6)(S1 (\relnsp{RRLL-}) S5)(S1 (\relnsp{RRLL+}) S12)
(S1 (\relnsp{RRLF}) S11)(S1 (\relnsp{RRLL+}) S4)(S1 (\relnsp{RRLL+}) S3)(S1 (\relnsp{RRFR}) S10)(S1 (\relnsp{RFLL}) S9)(S1 (\relnsp{RELE}) S8)
(S1 (\relnsp{EFBS}) S2)(S7 (\relnsp{RRLL-}) S6)(S7 (\relnsp{BRLL}) S5)(S7 (\relnsp{RRLLP}) S12)(S7 (\relnsp{RRLLP}) S11)(S7 (\relnsp{RRBL}) S4)
(S7 (\relnsp{SRSL}) S3)(S7 (\relnsp{BSEF}) S10)(S7 (\relnsp{RRLLP}) S9)(S7 (\relnsp{RRLLP}) S8)(S7 (\relnsp{RRRB}) S2)(S7 (\relnsp{ERRS}) S1))\\\\
((S5 (\relnsp{EFBS}) S6)(S12 (\relnsp{LSEL}) S6)(S12 (\relnsp{LLFL}) S5)(S11 (\relnsp{SRSL}) S6)(S11 (\relnsp{LSEL}) S5)(S11 (\relnsp{RRLLP}) S12)
(S4 (\relnsp{RRLL-}) S6)(S4 (\relnsp{RRLL-}) S5)(S4 (\relnsp{RELE}) S12)(S4 (\relnsp{RSER}) S11)(S3 (\relnsp{RRLL-}) S6)(S3 (\relnsp{RRLL-}) S5)
(S3 (\relnsp{RFLL}) S12)(S3 (\relnsp{RELE}) S11)(S3 (\relnsp{EFBS}) S4)(S10 (\relnsp{RRBL}) S6)(S10 (\relnsp{SRSL}) S5)(S10 (\relnsp{RRLLP}) S12)
(S10 (\relnsp{RRLLP}) S11)(S10 (\relnsp{RRRB}) S4)(S10 (\relnsp{ERRS}) S3)(S9 (\relnsp{LBLL}) S6)(S9 (\relnsp{LLLL-}) S5)(S9 (\relnsp{BSEF}) S12)
(S9 (\relnsp{LLRRP}) S11)(S9 (\relnsp{LSEL}) S4)(S9 (\relnsp{LLFL}) S3)(S9 (\relnsp{LLRRP}) S10)(S8 (\relnsp{BRLL}) S6)(S8 (\relnsp{LBLL}) S5)
(S8 (\relnsp{RRLLP}) S12)(S8 (\relnsp{BSEF}) S11)(S8 (\relnsp{SRSL}) S4)(S8 (\relnsp{LSEL}) S3)(S8 (\relnsp{LLRRP}) S10)(S8 (\relnsp{RRLLP}) S9)
(S2 (\relnsp{RRLL+}) S6)(S2 (\relnsp{RRLL+}) S5)(S2 (\relnsp{RRLF}) S12)(S2 (\relnsp{RRFR}) S11)(S2 (\relnsp{RRLL+}) S4)(S2 (\relnsp{RRLL+}) S3)
(S2 (\relnsp{RRRR+}) S10)(S2 (\relnsp{RELE}) S9)(S2 (\relnsp{RSER}) S8)(S1 (\relnsp{RRLL+}) S6)(S1 (\relnsp{RRLL+}) S5)(S1 (\relnsp{RRLL+}) S12)
(S1 (\relnsp{RRLF}) S11)(S1 (\relnsp{RRLL+}) S4)(S1 (\relnsp{RRLL+}) S3)(S1 (\relnsp{RRFR}) S10)(S1 (\relnsp{RFLL}) S9)(S1 (\relnsp{RELE}) S8)
(S1 (\relnsp{EFBS}) S2)(S7 (\relnsp{RRLL-}) S6)(S7 (\relnsp{BRLL}) S5)(S7 (\relnsp{RRLLP}) S12)(S7 (\relnsp{RRLLP}) S11)(S7 (\relnsp{RRBL}) S4)
(S7 (\relnsp{SRSL}) S3)(S7 (\relnsp{BSEF}) S10)(S7 (\relnsp{RRLLP}) S9)(S7 (\relnsp{RRLLP}) S8)(S7 (\relnsp{RRRB}) S2)(S7 (\relnsp{ERRS}) S1))\\
3 scenarios found, no further scenarios exist.
}

This result can be visualized with a diagram and can be interpreted with respect to the goals
of the reasoning task (see Section~\ref{application}).
\end{appendix}

\end{document}